\documentclass{article}

\usepackage{booktabs}                       % for \toprule ...
\usepackage[dvipsnames,table]{xcolor}       % color names ForestGreen, BrickRed
\newcommand{\green}[1]{\textcolor{ForestGreen}{#1}}
\newcommand{\red}[1]{\textcolor{red!80!black}{#1}}

    \usepackage[preprint]{neurips_2025}

\usepackage[utf8]{inputenc} % allow utf-8 input
\usepackage[T1]{fontenc}    % use 8-bit T1 fonts
\usepackage{hyperref}       % hyperlinks
\usepackage{url}            % simple URL typesetting
\usepackage{booktabs}       % professional-quality tables
\usepackage{amsfonts}       % blackboard math symbols
\usepackage{nicefrac}       % compact symbols for 1/2, etc.
\usepackage{microtype}      % microtypography
\usepackage{xcolor}         % colors

\usepackage{amsmath,amsfonts,bm}

\def\eqref#1{equation~\ref{#1}}
\def\1{\bm{1}}

\DeclareMathAlphabet{\mathsfit}{\encodingdefault}{\sfdefault}{m}{sl}
\SetMathAlphabet{\mathsfit}{bold}{\encodingdefault}{\sfdefault}{bx}{n}

\newcommand{\calA}{\mathcal{A}}

\usepackage{algorithm}
\usepackage{algpseudocode}

\usepackage{multirow}
\usepackage{enumitem}
\usepackage[most]{tcolorbox}
\usepackage{float}
\usepackage{caption}
\usepackage{subcaption}
\usepackage{longtable}
\usepackage{listings}
\usepackage{amssymb}
\usepackage{mathtools}
\usepackage{amsthm}
\usepackage{multicol}
\usepackage{makecell}
\usepackage{graphicx}

\usepackage{colortbl}
\usepackage{makecell}
\usepackage{pifont}
\usepackage{titletoc}

\newcommand{\xmark}{\ding{55}}

\title{When Thinking Fails: The Pitfalls of Reasoning for Instruction-Following in LLMs}

\author{%
    Xiaomin Li\thanks{Correspondence to: Xiaomin Li (email: xiaominli@g.harvard.edu)} \\
  Harvard University\\
  \And
  Zhou Yu   \\
  Amazon \\
  \And
  Zhiwei Zhang \\
  Amazon \\
  \And
  Xupeng Chen \\
  NYU \\
  \And
  Ziji Zhang \\
  Amazon \\
  \AND
  Yingying Zhuang \\
  Amazon \\
  \And
  Narayanan Sadagopan \\
  Amazon \\
  \And
  Anurag Beniwal \\
  Amazon \\
}

\begin{document}

\maketitle

\begin{abstract}
Reasoning-enhanced large language models (RLLMs), whether explicitly trained for reasoning or prompted via chain-of-thought (CoT), have achieved state-of-the-art performance on many complex reasoning tasks. However, we uncover a surprising and previously overlooked phenomenon: explicit CoT reasoning can significantly degrade instruction-following accuracy. Evaluating 15 models on two benchmarks: IFEval (with simple, rule-verifiable constraints) and ComplexBench (with complex, compositional constraints), we consistently observe performance drops when CoT prompting is applied. Through large-scale case studies and an attention-based analysis, we identify common patterns where reasoning either helps (e.g., with formatting or lexical precision) or hurts (e.g., by neglecting simple constraints or introducing unnecessary content). We propose a metric, \textit{constraint attention}, to quantify model focus during generation and show that CoT reasoning often diverts attention away from instruction-relevant tokens. To mitigate these effects, we introduce and evaluate four strategies: in-context learning, self-reflection, self-selective reasoning, and classifier-selective reasoning. Our results demonstrate that selective reasoning strategies, particularly classifier-selective reasoning, can substantially recover lost performance. To our knowledge, this is the first work to systematically expose reasoning-induced failures in instruction-following and offer practical mitigation strategies.
\end{abstract}

%%%%%%%%%%%%%%%%%%%%%%%%%%%%%%%%%%%%%%%%%%%%%%%%%%%%%%%%%%%%%%%%%%%%%%%%%%%%%%%%%%%%%%%%%%%%%%%%%%%%%%%%%%%%%%%%%%%%%%%%%%%%%%%%%%%%%%%%%%%%%%%%%%%%%%%%%%%%%%%%%%%%%%%%%%%%%%%%%%%%%%%%%%%%%%%%%%%%%%%%%%%%%%%%%%%%%%%%%%%%%%%%%%%%%%%%%%%%%%%%%%%%%%%%%%%%%%%%%%%%%%%%%%%%%%%%%%%%%%%%%%%%%%%%%%%%%%%%%%%%%%%%%%%%%%%%%%%%%%%%%%%%%%%%%%%%%%%%%%%%%%%%%%%%%%%%%%%%%%%%%%%%%%%%%%%%%%%%%%%%%%%%%%%%%%%%%%%%%
\section{Introduction}\label{sec:Introduction}
Reasoning-enhanced large language models (RLLMs) have demonstrated remarkable success across a variety of tasks, including mathematical problem solving, planning, and multi-hop question answering~\citep{guo2025deepseek, achiam2023gpt, grattafiori2024llama, xu2023re, zhou2022least, wu2024comparative, fu2022complexity, qi2024mutual, chae2024language}. A central contributor to these advancements is \textit{chain-of-thought} (CoT) prompting~\citep{wei2022chain, nye2021show, joshi2023machine, lanham2023measuring}, which explicitly encourages models to reason step-by-step prior to providing an answer. Recent prominent models, such as DeepSeek-R1~\citep{guo2025deepseek}, Claude~\citep{anthropic2025claude37}, and OpenAI’s O-series~\citep{openai2024introducingo1}, either incorporate CoT-style reasoning in their fine-tuning procedures or explicitly offer reasoning as an inherent capability. While CoT generally improves performance on complex reasoning tasks, it incurs higher computational cost due to longer outputs and increased inference latency. Moreover, its impact on more structured tasks—such as instruction following—remains underexplored. Instruction following, the ability to comply with user-specified constraints, is essential for alignment, safety, and practical usability of language models~\citep{ouyang2022training, zhang2023instruction}. This raises a natural question: \textbf{Does explicit reasoning actually help a model follow instructions more accurately?}

In this paper, we answer this question empirically and arrive at a surprising conclusion: reasoning via CoT can \textit{degrade} a model’s ability to follow instructions. To investigate this, we evaluate 15 language models of varying sizes and training paradigms, including general-purpose models (e.g., Llama, Mixtral) and reasoning-tuned models (e.g., Claude 3.7, DeepSeek-R1), on two complementary instruction-following benchmarks: \textit{IFEval} \citep{zhou2023instruction} consists of prompts with simple, independently verifiable constraints (e.g., “write at least 400 words” or “mention \texttt{AI} three times”). In contrast, \textit{ComplexBench} \citep{wen2024benchmarking} includes instructions formed through compositional logic, combining multiple dependent constraints via operations like chaining, selection, and nesting. Across both datasets, we observe a consistent and often substantial accuracy drop when models are prompted with CoT. This finding is surprising, given that reasoning is typically expected to improve performance on many tasks that are even more challenging than instruction following. 

To understand this phenomenon, we conduct two complementary analyses. First, we perform a large-scale manual case study of samples where reasoning notably affects performance. We find that reasoning helps in two common scenarios: \textit{(1) satisfying formatting or structural requirements}, and \textit{(2) enforcing lexical constraints that override default tendencies}. However, it often hurts performance by: \textit{(1) over-focusing on high-level content and neglecting simple constraints}, or \textit{(2) introducing redundant or well-intentioned content that unintentionally violates constraints}. We provide detailed examples and analyses of each scenario. Second, we investigate the impact of reasoning through an attention-based analysis. We propose a quantitative measure: \textit{constraint attention}, which is based on attention scores directed toward constraint tokens within instructions. Visualizations of attention patterns during response generation consistently demonstrate reduced constraint awareness when CoT prompting is employed, an effect observed across different datasets, models, and layers. This shift in attention might explain, at least partially, why reasoning can diminish instruction adherence.

Following these insights, we propose and evaluate four methods to mitigate the adverse impacts of reasoning on instruction-following accuracy: (1) \textbf{In-context learning}, where we identify and correct typical reasoning-induced failures, incorporating these examples into prompts; (2) \textbf{Self-reflection}, prompting models to evaluate and adjust their reasoning processes and candidate responses; (3) \textbf{Self-selective reasoning}, allowing models to autonomously decide when reasoning is beneficial; and (4) \textbf{Classifier-selective reasoning}, where a trained classifier determines when reasoning is necessary. Among these methods, we find that classifier-selective reasoning strategies offer substantial gains across benchmarks, while self-reflection is particularly helpful for larger models and simple instructions. We thoroughly discuss the comparative strengths and limitations of these mitigation strategies in Section~\ref{sec:Methods}.

In summary, our main contributions are listed as follows:

\begin{itemize}
% \item We evaluate 15 language models on comprehensive instruction-following benchmarks, revealing that explicit \textbf{reasoning can negatively impact instruction-following capabilities}\footnote{Code and data for reproducing experiments is available at: \url{https://anonymous.4open.science/r/Submission-ReasonIF-0548/}.}. To our knowledge, \textbf{we are the first to discover and systematically explore this phenomenon}.
\item We evaluate 15 language models on comprehensive instruction-following benchmarks, revealing that explicit \textbf{reasoning can negatively impact instruction-following capabilities}. To our knowledge, \textbf{we are the first to discover and systematically explore this phenomenon}.
\item We provide detailed analyses of reasoning-induced failures, \textbf{categorizing scenarios where reasoning either helps or hinders}, and \textbf{quantify attention shifts that explain these performance drops}.
\item We \textbf{propose and rigorously examine multiple mitigation strategies}, including in-context learning, self-reflection, self-selective reasoning, and classifier-selective reasoning, demonstrating their effectiveness and highlighting promising directions for future research.
\end{itemize}

%%%%%%%%%%%%%%%%%%%%%%%%%%%%%%%%%%%%%%%%%%%%%%%%%%%%%%%%%%%%%%%%%%%%%%%%%%%%%%%%%%%%%%%%%%%%%%%%%%%%%%%%%%%%%%%%%%%%%%%%%%%%%%%%%%%%%%%%%%%%%%%%%%%%%%%%%%%%%%%%%%%%%%%%%%%%%%%%%%%%%%%%%%%%%%%%%%%%%%%%%%%%%%%%%%%%%%%%%%%%%%%%%%%%%%%%%%%%%%%%%%%%%%%%%%%%%%%%%%%%%%%%%%%%%%%%%%%%%%%%%%%%%%%%%%%%%%%%%%%%%%%%%%%%%%%%%%%%%%%%%%%%%%%%%%%%%%%%%%%%%%%%%%%%%%%%%%%%%%%%%%%%%%%%%%%%%%%%%%%%%%%%%%%%%%%%%%%%%
\section{Related Work}\label{sec:RelatedWork} 
\textbf{Chain-of-Thought and Reasoning LLMs.}\quad
State-of-the-art large language models (LLMs) often leverage explicit reasoning capabilities, exemplified by models such as OpenAI's O-series \citep{openai2024introducingo1}, DeepSeek's R1 \citep{guo2025deepseek}, and Anthropic's Claude \citep{anthropic2025claude37}. These models are trained on datasets incorporating not just direct responses but also explicit reasoning processes known as Chain-of-Thought (CoT) \citep{wei2022chain, nye2021show, joshi2023machine, lanham2023measuring}. CoT prompting, which encourages step-by-step reasoning, has demonstrated significant success, particularly in domains requiring complex reasoning, such as mathematics \citep{guo2025deepseek, achiam2023gpt, grattafiori2024llama, xu2023re, zhou2022least, wu2024comparative, fu2022complexity, qi2024mutual, chae2024language}. However, CoT can also introduce additional computational costs and may yield limited or no improvement in certain contexts \citep{kambhampati2024position, wang2024mmlu, sprague2024cot}. Our study further explores this dual nature of reasoning, highlighting cases where it can negatively impact instruction-following performance.

\textbf{Instruction Following.}\quad
Instruction-following is essential for aligning language model outputs with user expectations, enabling the models to reliably execute user-specified tasks. Techniques such as instruction tuning (fine-tuning models on extensive datasets of instruction-response pairs) have been instrumental in fostering this capability \citep{ouyang2022training, wang2023self, longpre2023flan, bai2022training, wei2022emergent, chung2022scaling, muennighoff2022crosslingual}. This ability is critical for bridging the gap between pre-training objectives and desirable human-aligned behaviors such as helpfulness and relevance \citep{radford2019language, brown2020language}. Despite its importance, instruction-following remains challenging, especially when instructions involve complex, multifaceted requirements \citep{zhang2023instruction, he2024complex, gudibande2023false, kung2023instruction, heo2024llms}. To systematically evaluate LLM instruction adherence, benchmarks such as IFEval \citep{zhou2023instruction} and ComplexBench \citep{wen2024benchmarking} have been introduced. IFEval focuses on rule-verifiable, straightforward constraints, while ComplexBench assesses models on sophisticated instructions involving nested and dependent constraints. Further details and comprehensive analyses of these benchmarks will be provided in our experimental evaluations (Section~\ref{sec:Experiments}).

%%%%%%%%%%%%%%%%%%%%%%%%%%%%%%%%%%%%%%%%%%%%%%%%%%%%%%%%%%%%%%%%%%%%%%%%%%%%%%%%%%%%%%%%%%%%%%%%%%%%%%%%%%%%%%%%%%%%%%%%%%%%%%%%%%%%%%%%%%%%%%%%%%%%%%%%%%%%%%%%%%%%%%%%%%%%%%%%%%%%%%%%%%%%%%%%%%%%%%%%%%%%%%%%%%%%%%%%%%%%%%%%%%%%%%%%%%%%%%%%%%%%%%%%%%%%%%%%%%%%%%%%%%%%%%%%%%%%%%%%%%%%%%%%%%%%%%%%%%%%%%%%%%%%%%%%%%%%%%%%%%%%%%%%%%%%%%%%%%%%%%%%%%%%%%%%%%%%%%%%%%%%%%%%%%%%%%%%%%%%%%%%%%%%%%%%%%%%%
\section{Experiments}\label{sec:Experiments}
%%%%%%%%%%%%%%%%%%%%%%%%%%%%%%%%%%%%%%%%%%%%%%%%%%%%%%%%%%%%%%%%%%%%%%%%%%%%%%%%
\subsection{Datasets and Evaluation Metrics}\label{subsec:Datasets-EvalulationMetrics}
We use two benchmark datasets, \textbf{IFEval} and \textbf{ComplexBench}, to comprehensively evaluate the instruction-following capabilities of language models:

\textbf{IFEval} is a synthetic dataset consisting of 541 prompts, each associated with one to three verifiable constraints drawn from 25 types (e.g., word count, formatting, keyword usage). We adopt instruction-level \textit{loose accuracy}, which allows minor formatting deviations (e.g., markdown wrappers) to avoid penalizing responses for stylistic differences, thereby better reflecting practical robustness.

\textbf{ComplexBench} is a manually curated dataset designed to assess models on complex compositional instructions formed through four operations: \textit{And}, \textit{Chain}, \textit{Selection}, and \textit{Nested}. It contains 1,150 instructions and over 5,300 scoring questions, covering 4 constraint types across 19 dimensions (e.g., lexical, semantic, formatting). Evaluation combines rule-based and LLM-based assessments. For our experiments, we translated all scoring rules into English and manually verified them to construct a fully English-compatible version.

\textbf{Evaluation Metrics:} For both datasets, we report the proportion of constraints satisfied per instruction. In ComplexBench, dependency logic applies: failure in a prerequisite constraint results in automatic failure of all dependent constraints.

%%%%%%%%%%%%%%%%%%%%%%%%%%%%%%%%%%%%%%%%%%%%%%%%%%%%%%%%%%%%%%%%%%%%%%%%%%%%%%%%
\subsection{Models}
We evaluate a diverse set of models, including both closed-source models (e.g., \texttt{Claude3.7-Sonnet}) and reasoning-focused models (e.g., \texttt{DeepSeek-R1}, \texttt{Qwen-R1-distilled variants}). Our open-source selection spans parameter scales from 1B to 70B. All model inferences use a temperature of 0. Open-source models are run without quantization using 4 \textbf{NVIDIA-H100-80GB} GPUs.

%%%%%%%%%%%%%%%%%%%%%%%%%%%%%%%%%%%%%%%%%%%%%%%%%%%%%%%%%%%%%%%%%%%%%%%%%%%%%%%%
\subsection{CoT Prompting}
We compare model behavior with and without Chain-of-Thought (CoT) reasoning. The CoT prompts instruct models to reason step by step before producing an answer (exact prompt provided in Appendix~\ref{sec:Appendix-Prompts}). We then assess instruction-following performance based on the model's answer in each setting.

%%%%%%%%%%%%%%%%%%%%%%%%%%%%%%%%%%%%%%%%%%%%%%%%%%%%%%%%%%%%%%%%%%%%%%%%%%%%%%%%
\subsection{Results}
Performance on IFEval and ComplexBench is reported in the first three columns of Table~\ref{tab:IFEval-ComplexBench-results}, with visual summaries provided in Figure~\ref{fig:MainResultsPlot-Combined}. Notably, 13 out of 14 models experience performance degradation on IFEval when CoT prompting is applied, and all models show declines on ComplexBench. For instance, the accuracy of \texttt{Llama3-8B-Instruct} drops from 75.2\% to 59.0\%, a reduction of over 16 percentage points.

We further compare reasoning-enhanced models with their corresponding base variants—e.g., \texttt{Claude3.7-Sonnet} vs.\ \texttt{Claude3.7-Sonnet-Think}, and \texttt{DeepSeek-V3} vs.\ \texttt{DeepSeek-R1}. These results, shown in Table~\ref{tab:RLLM-Comparison-Table}, reveal consistent performance drops when reasoning is enabled. While these comparisons are not fully controlled—reasoning-tuned models may undergo additional training stages such as supervised fine-tuning or RLHF—the performance decline remains evident.

Overall, these findings uncover a surprising and underexplored vulnerability: explicit reasoning, while often helpful in complex tasks, can increase the likelihood of violating instruction constraints, thereby impairing instruction-following reliability.

\definecolor{lightgray}{gray}{0.95}
\begin{table}[ht]
\centering
\resizebox{\linewidth}{!}{
\renewcommand{\arraystretch}{1.4}
\begin{tabular}{l|c|c|c|c|c|c}
% \toprule
% \rowcolor{lightgray}
% \multicolumn{7}{c}{\textbf{IFEval Results}} \\
    \toprule
    \cellcolor{lightgray}\textbf{IFEval Results} 
      & \multicolumn{2}{c|}{\bfseries Baseline} 
  & \multicolumn{4}{c}{\bfseries Mitigation Methods} \\
\midrule
Model & Original & {CoT} & FewShot & SelfReflection & \makecell{Self-Selective\\ Reasoning} & \makecell{Classifier-Selective\\ Reasoning} \\
\midrule
% GPT-4o-mini & \green{82.6} & \red{76.9} & 75.0 ($\downarrow$ 1.9) & 79.7 ($\uparrow$ 2.8) & 77.3 ($\uparrow$ 0.4) & \textbf{82.1} ($\uparrow$ 5.2) \\
Claude-3.5-Haiku & \green{86.9} & \red{79.5} & 80.2 ($\uparrow$ 0.7) & 81.7 ($\uparrow$ 2.2) & 80.8 ($\uparrow$ 1.3) & \textbf{85.8} ($\uparrow$ 6.3) \\
Claude-3.5-Sonnet & \green{86.5} & \red{79.5} & 87.4 ($\uparrow$ 8.0) & \textbf{87.6} ($\uparrow$ 8.1) & 81.9 ($\uparrow$ 2.4) & 84.8 ($\uparrow$ 5.4) \\
Claude-3.7-Sonnet & \green{90.6} & \red{90.2} & 89.3 ($\downarrow$ 0.9) & \textbf{92.1} ($\uparrow$ 1.9) & 90.4 ($\uparrow$ 0.2) & 90.4 ($\uparrow$ 0.2) \\
DeepSeek-V3 & \green{85.2} & \red{84.3} & 85.2 ($\uparrow$ 0.9) & \textbf{88.3} ($\uparrow$ 4.0) & 85.3 ($\uparrow$ 1.0) & 84.2 ($\downarrow$ 0.1) \\
Llama-3.2-1B-Instruct & \green{49.0} & \red{40.7} & 13.1 ($\downarrow$ 27.5) & 34.9 ($\downarrow$ 5.7) & 45.7 ($\uparrow$ 5.0) & \textbf{47.5} ($\uparrow$ 6.8) \\
Llama-3.2-3B-Instruct & \green{70.4} & \red{61.9} & 37.2 ($\downarrow$ 24.8) & 64.9 ($\uparrow$ 3.0) & 61.9 ($\downarrow$ 0.0) & \textbf{68.8} ($\uparrow$ 6.8) \\
Meta-Llama-3-8B-Instruct & \green{75.2} & \red{59.0} & 36.8 ($\downarrow$ 22.2) & 65.1 ($\uparrow$ 6.1) & 59.0 ($\downarrow$ 0.0) & \textbf{69.7} ($\uparrow$ 10.7) \\
Llama-3.1-70B-Instruct & \green{85.6} & \red{77.3} & 44.5 ($\downarrow$ 32.7) & \textbf{84.3} ($\uparrow$ 7.0) & 77.3 ($\downarrow$ 0.0) & 83.2 ($\uparrow$ 5.9) \\
Mixtral-8x7B-Instruct & \red{53.0} & \green{56.4} & 32.0 ($\downarrow$ 24.4) & \textbf{55.6} ($\downarrow$ 0.7) & 33.1 ($\downarrow$ 23.3) & 55.1 ($\downarrow$ 1.3) \\
Qwen2.5-1.5B-Instruct & \green{35.9} & \red{31.6} & 20.1 ($\downarrow$ 11.5) & 26.6 ($\downarrow$ 5.0) & 32.7 ($\uparrow$ 1.1) & \textbf{34.8} ($\uparrow$ 3.2) \\
DeepSeek-R1-Distill-Qwen-1.5B & \green{16.8} & \red{13.7} & \textbf{22.0} ($\uparrow$ 8.3) & 20.0 ($\uparrow$ 6.3) & 13.9 ($\uparrow$ 0.2) & 17.0 ($\uparrow$ 3.3) \\
Qwen2.5-7B-Instruct & \green{63.6} & \red{57.7} & 63.4 ($\uparrow$ 5.7) & \textbf{74.3} ($\uparrow$ 16.6) & 59.7 ($\uparrow$ 2.0) & 68.8 ($\uparrow$ 11.1) \\
DeepSeek-R1-Distill-Qwen-7B & \green{30.9} & \red{25.1} & 30.1 ($\uparrow$ 5.0) & \textbf{36.4} ($\uparrow$ 11.3) & 26.2 ($\uparrow$ 1.1) & 30.1 ($\uparrow$ 5.0) \\
% \rowcolor{lightgray}
\midrule
% \rowcolor{lightgray}
% \multicolumn{7}{c}{\textbf{ComplexBench Results}} \\
    \toprule
    \cellcolor{lightgray}\textbf{ComplexBench Results} 
      & \multicolumn{2}{c|}{\bfseries Baseline} 
  & \multicolumn{4}{c}{\bfseries Mitigation Methods} \\
\midrule
Model & Original & {CoT} & FewShot & SelfReflection & \makecell{Self-Selective\\ Reasoning} & \makecell{Classifier-Selective\\ Reasoning} \\
\midrule
% GPT-4o-mini & \green{66.5} & \red{60.3} & 58.9 ($\downarrow$ 1.4) & 60.0 ($\downarrow$ 0.3) & \textbf{66.7} ($\uparrow$ 6.4) & 66.4 ($\uparrow$ 6.1) \\
Claude-3.5-Haiku & \green{66.9} & \red{62.1} & 59.7 ($\downarrow$ 2.4) & 61.5 ($\downarrow$ 0.6) & \textbf{67.6} ($\uparrow$ 5.5) & 67.1 ($\uparrow$ 5.0) \\
Claude-3.5-Sonnet & \green{67.5} & \red{66.0} & 66.9 ($\uparrow$ 0.9) & 65.5 ($\downarrow$ 1.5) & 67.6 ($\uparrow$ 1.6) & \textbf{68.1} ($\uparrow$ 2.1) \\
Claude-3.7-Sonnet & \green{69.8} & \red{67.0} & 67.4 ($\uparrow$ 0.4) & 67.8 ($\uparrow$ 0.8) & \textbf{69.6} ($\uparrow$ 2.5) & 69.4 ($\uparrow$ 2.4) \\
DeepSeek-V3 & \green{71.2} & \red{71.1} & 66.0 ($\downarrow$ 5.1) & 65.2 ($\downarrow$ 5.9) & 71.4 ($\uparrow$ 0.3) & \textbf{71.5} ($\uparrow$ 0.4) \\
Llama-3.2-1B-Instruct & \green{36.0} & \red{26.0} & 19.7 ($\downarrow$ 6.3) & 23.7 ($\downarrow$ 2.3) & \textbf{36.2} ($\uparrow$ 10.2) & 35.8 ($\uparrow$ 9.8) \\
Llama-3.2-3B-Instruct & \green{50.0} & \red{45.7} & 49.2 ($\uparrow$ 3.5) & 44.7 ($\downarrow$ 1.0) & \textbf{49.8} ($\uparrow$ 4.1) & 49.7 ($\uparrow$ 4.0) \\
Meta-Llama-3-8B-Instruct & \green{55.8} & \red{54.9} & 54.7 ($\downarrow$ 0.2) & 49.1 ($\downarrow$ 5.8) & 55.4 ($\uparrow$ 0.5) & \textbf{55.8} ($\uparrow$ 0.9) \\
Llama-3.1-70B-Instruct & \green{66.8} & \red{60.2} & 66.8 ($\uparrow$ 6.6) & 61.9 ($\uparrow$ 1.7) & 66.7 ($\uparrow$ 6.5) & \textbf{67.8} ($\uparrow$ 7.6) \\
Mixtral-8x7B-Instruct & \green{60.4} & \red{58.3} & 58.2 ($\downarrow$ 0.1) & 56.5 ($\downarrow$ 1.7) & 59.6 ($\uparrow$ 1.3) & \textbf{59.6} ($\uparrow$ 1.3) \\
Qwen2.5-1.5B-Instruct & \green{44.1} & \red{38.8} & 38.1 ($\downarrow$ 0.7) & 31.3 ($\downarrow$ 7.5) & \textbf{44.0} ($\uparrow$ 5.2) & \textbf{44.0} ($\uparrow$ 5.2) \\
DeepSeek-R1-Distill-Qwen-1.5B & \green{18.9} & \red{16.7} & \textbf{24.1} ($\uparrow$ 7.5) & 20.5 ($\uparrow$ 3.8) & 18.7 ($\uparrow$ 2.0) & 18.7 ($\uparrow$ 2.0) \\
Qwen2.5-7B-Instruct & \green{60.2} & \red{52.6} & 60.2 ($\uparrow$ 7.6) & 55.8 ($\uparrow$ 3.3) & 63.4 ($\uparrow$ 10.8) & \textbf{63.4} ($\uparrow$ 10.9) \\
DeepSeek-R1-Distill-Qwen-7B & \green{46.0} & \red{38.6} & \textbf{50.2} ($\uparrow$ 11.6) & 37.0 ($\downarrow$ 1.6) & 46.0 ($\uparrow$ 7.4) & 45.5 ($\uparrow$ 6.9) \\
\bottomrule
\end{tabular}
}
\vspace{0.5em}
\caption{
\textbf{Instruction-following performance on IFEval and ComplexBench.}
\green{Green} and \red{red} indicate whether \textit{Original} or \textit{CoT} performed better. Each mitigation column reports the accuracy along with its change relative to CoT, using $\uparrow$ for improvement and $\downarrow$ for decline. 
\textbf{Bold} values highlight the best-performing mitigation method for each model.
}
\label{tab:IFEval-ComplexBench-results}
\end{table}

\begin{table}[ht]
\centering
{\fontsize{8}{9.6}\selectfont % Start smaller font scope
\renewcommand{\arraystretch}{1}
\setlength{\tabcolsep}{3pt}
\begin{tabular}{lcccc}
\toprule
& Claude-3.7-Sonnet & Claude-3.7-Sonnet-Think & DeepSeek-V3 & DeepSeek-R1 \\
\midrule
\textbf{IFEval}       & \green{90.6} & \red{90.2} & \green{85.2} & \red{83.3} \\
\textbf{ComplexBench} & \green{69.8} & \red{69.0} & \green{71.2} & \red{71.1} \\
\bottomrule
\end{tabular}
} % End font size scope
\vspace{0.5em}
\caption{Comparison of reasoning-enabled models vs.\ their non-reasoning counterparts. \green{Green} indicates higher-performing model in each pair; \red{red} indicates lower.}
\label{tab:RLLM-Comparison-Table}
\end{table}

%%%%%%%%%%%%%%%%%%%%%%%%%%%%%%%%%%%%%%%%%%%%%%%%%%%%%%%%%%%%%%%%%%%%%%%%%%%%%%%%%%%%%%%%%%%%%%%%%%%%%%%%%%%%%%%%%%%%%%%%%%%%%%%%%%%%%%%%%%%%%%%%%%%%%%%%%%%%%%%%%%%%%%%%%%%%%%%%%%%%%%%%%%%%%%%%%%%%%%%%%%%%%%%%%%%%%%%%%%%%%%%%%%%%%%%%%%%%%%%%%%%%%%%%%%%%%%%%%%%%%%%%%%%%%%%%%%%%%%%%%%%%%%%%%%%%%%%%%%%%%%%%%%%%%%%%%%%%%%%%%%%%%%%%%%%%%%%%%%%%%%%%%%%%%%%%%%%%%%%%%%%%%%%%%%%%%%%%%%%%%%%%%%%%%%%%%%%%%
\section{Analysis}\label{sec:Analysis} 
To better understand when and why reasoning degrades instruction-following, we conduct two analyses: (1) a manual case study examining when CoT helps or hurts constraint satisfaction, and (2) an attention-based analysis investigating how reasoning shifts model focus away from constraints during generation.

\subsection{Case Study}\label{subsec:CaseStudy}
We manually examined all 541 samples from IFEval and over 1,000 samples from ComplexBench, focusing on cases where CoT affected whether constraints were satisfied. For each case, we analyzed which constraints were impacted and why the reasoning-based answer either improved or degraded performance. 

Due to the length of the instructions and responses, we present detailed illustrative examples in Appendix~\ref{sec:Appendix-CaseStudy}. For each example, we include outputs from both the base model and the reasoning-enabled model, showing their respective \texttt{Answer} and, for CoT, the full \texttt{Thinking} process. We also report the number of constraints satisfied and include a case analysis explaining how reasoning helped or harmed performance. Although we examined a large number of examples, the failure and success cases largely fall into four recurring patterns \footnote{Beyond these behavioral patterns, reasoning also increases computational cost due to longer outputs and additional thinking tokens.}, which we summarize below:

\textbf{Reasoning Helps \checkmark}
\begin{enumerate}
    \item[\textbf{1.}] \textit{Formatting and Structural Adherence:} Reasoning improves compliance with structural constraints, such as producing valid JSON, wrapping the output in double quotes, or following markdown syntax.
    \item[\textbf{2.}] \textit{Lexical and Keyword Precision:} Reasoning enhances adherence to lexical requirements, including inserting rare characters (e.g., the letter \texttt{q} six times), omitting final punctuation, or using exactly 15 capitalized words.
\end{enumerate}

\textbf{Reasoning Hurts  \xmark }
\begin{enumerate}
    \item[\textbf{3.}] \textit{Over-Focusing on High-Level Content and Neglecting Simple Constraints:} When multiple constraints are present, reasoning often emphasizes content planning at the expense of simpler mechanical constraints. Common issues include exceeding word count limits, failing to repeat prompts exactly, using capital letters in lowercase-only tasks, or appending unnecessary content after required phrases.
    \item[\textbf{4.}] \textit{Introducing Unnecessary Content that Violates Constraints:} Reasoning frequently inserts redundant or well-intentioned additions—such as explanations, translations, or emphasis—that break constraints. Typical violations include: inserting English text into ``foreign language only'' outputs, including commas in ``no commas'' tasks, appending commentary to quote-only responses, or exceeding limits on capitalized words.
\end{enumerate}

%%%%%%%%%%%%%%%%%%%%%%%%%%%%%%%%%%%%%%%%%%%%%%%%%%%%%%%%%%%%%%%%%%%%%%%%%%%%%%%%%%%%%%%%%%%%%%%%%%%%%%%%%%%%%%%%%%%%%%%%%%%%%%%%%%%%%%%%%%%%%%
\subsection{Constraint-Aware Attention Analysis}  \label{sec:AttentionAnalysis}
To understand why reasoning may degrade instruction-following, we analyze whether models pay less attention to constraint-relevant parts of the prompt during response generation. In many failure cases, we observed that models neglect certain constraints, either by overemphasizing content planning or introducing irrelevant information. To investigate this phenomenon systematically, we conduct an attention-based analysis that tracks the model's focus on constraint tokens throughout generation.

We define a \textit{constraint-attention} metric to quantify the model's awareness of constraint-relevant tokens. For each instruction, we first use \texttt{GPT-4o} to automatically extract substrings corresponding to each constraint, and then map them to token indices in the prompt. During generation, we compute attention scores directed toward these tokens for both the reasoning and answer segments. Each model is run twice per instruction: (i) \textbf{Base run}: \texttt{Instruction} $\rightarrow$ \texttt{Answer}, and (ii) \textbf{Reasoning run (CoT)}: \texttt{Instruction} $\rightarrow$ \texttt{Think} $\rightarrow$ \texttt{Answer}. We focus our comparison on the \emph{answer segments} of both runs.

\paragraph{Transformer Attention.}
Let the prompt contain $T_0$ tokens $x_{1:T_0}$, and let the model generate $T$ new tokens $y_{1:T}$. At generation step $t$ ($1 \le t \le T$), the visible context is $(x_{1:T_0}, y_{1:t-1})$. Let $h^{(l)}_{t-1}$ denote the hidden state of the most recent token at layer $l$, and let $k \in \{1, \dots, H\}$ index attention heads. The attention components for layer $l$ and head $k$ are computed as:
\[
Q_k^{(l,t)} \;=\; W_Q^{(l,k)}\,h_{t-1}^{(l)},\quad
K_k^{(l)}    \;=\; W_K^{(l,k)}\,H_{1:T_0+t-2}^{(l)},\quad
V_k^{(l)}    \;=\; W_V^{(l,k)}\,H_{1:T_0+t-2}^{(l)},
\]
where $h_{t-1}^{(l)}$ is a single vector (current token),
and $H_{1:T_0+t-2}^{(l)}$ is the matrix of all prior hidden vectors
(prompt and generated context). This leads to scaled attention weights $A_k^{(l,t)} \;=\; \mathrm{softmax}\Bigl( \tfrac{Q_k^{(l,t)}K_k^{(l)\top}} {\sqrt{d_k}} \Bigr)$. Averaging over all heads yields the layer-level attention vector
\[
a^{(l,t)} \;=\; \frac{1}{H}\sum_{k=1}^{H} A_k^{(l,t)} .
\]

\begin{figure}[ht]
\centering

% Left: IFEval
\begin{minipage}[t]{0.485\textwidth}
    \centering
    \textbf{(a) IFEval – Attention Traces} \\[0.5em]
    \includegraphics[width=1.0\linewidth, height=0.55\linewidth]{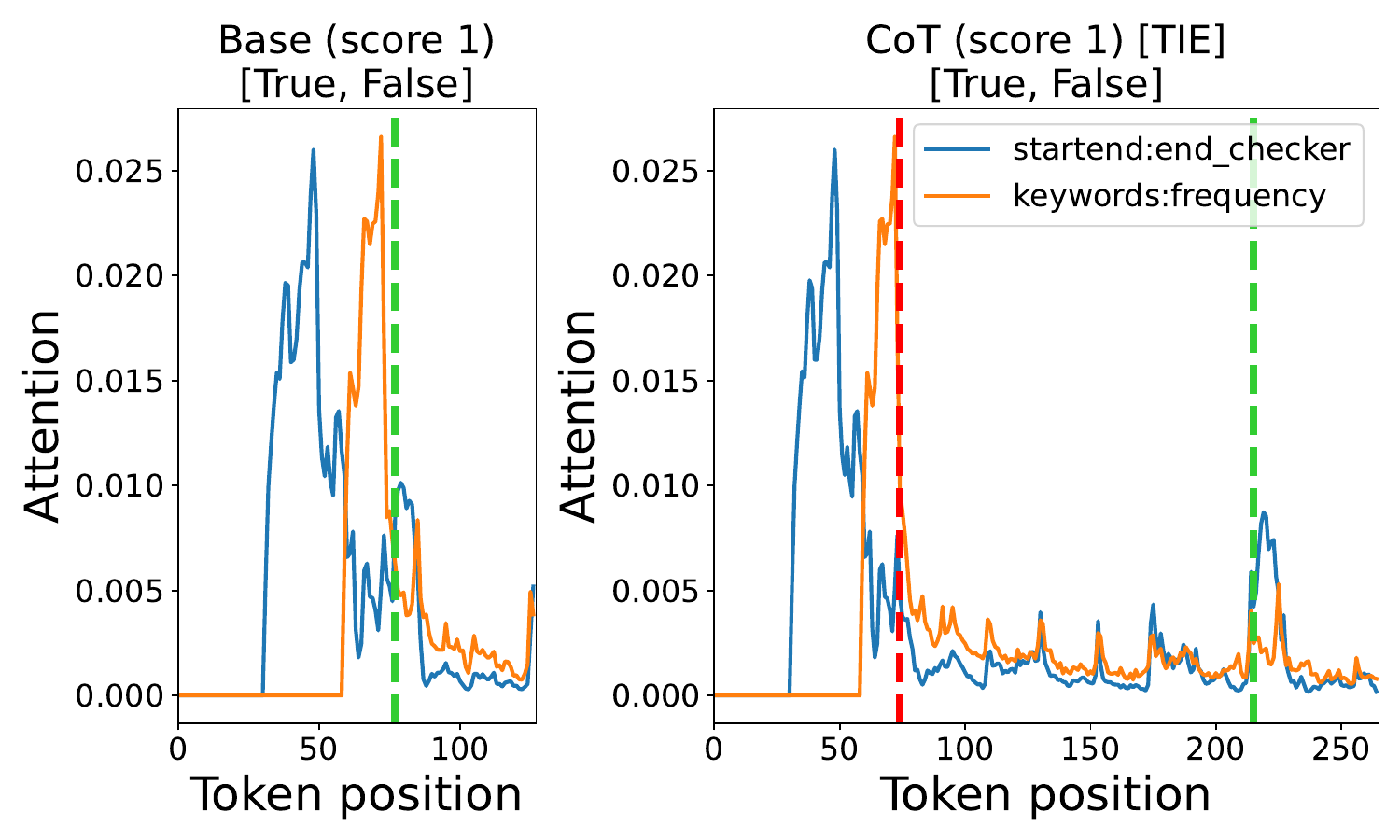}
    \vspace{0.5em}
    \includegraphics[width=1.0\linewidth, height=0.55\linewidth]{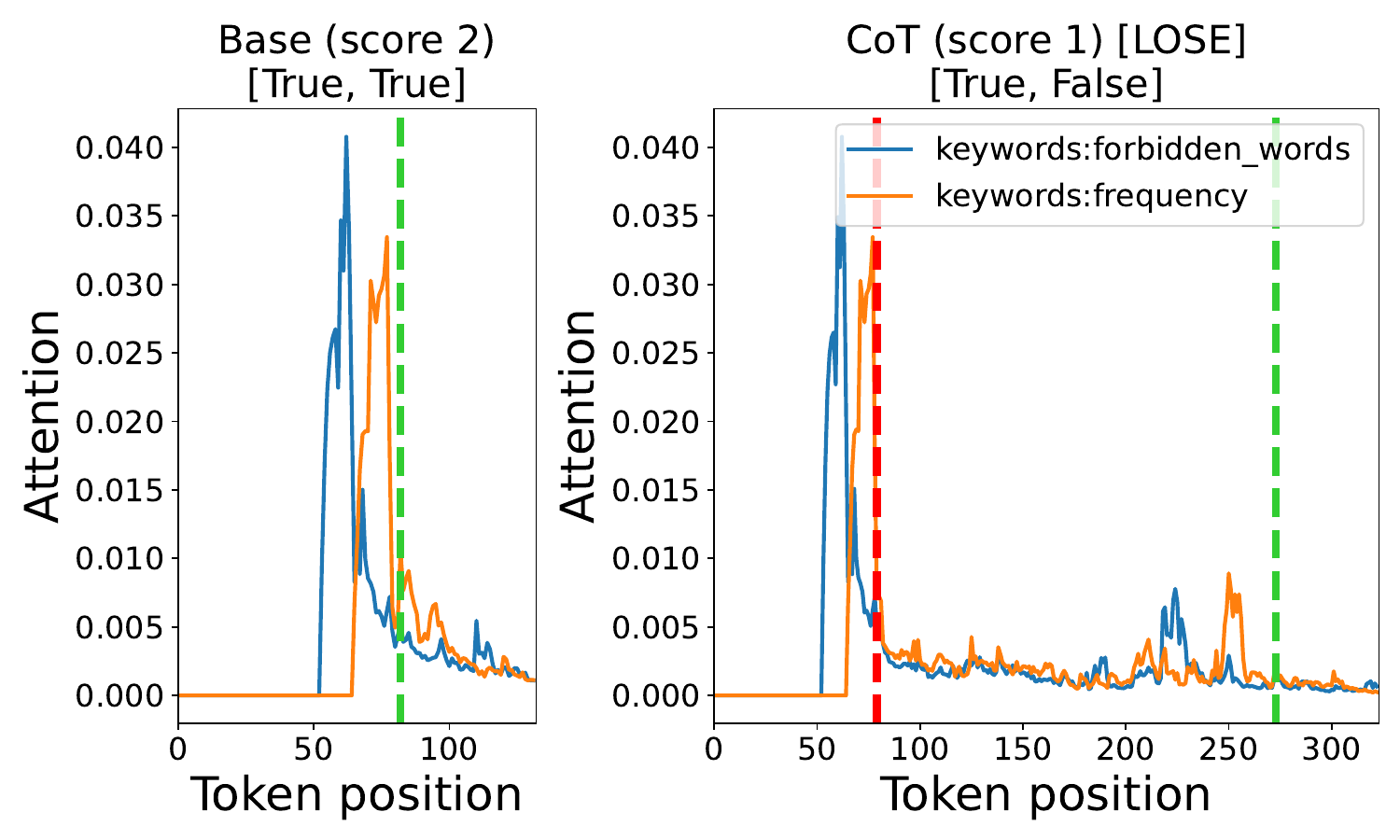}
    \vspace{0.5em}
    \includegraphics[width=1.0\linewidth, height=0.55\linewidth]{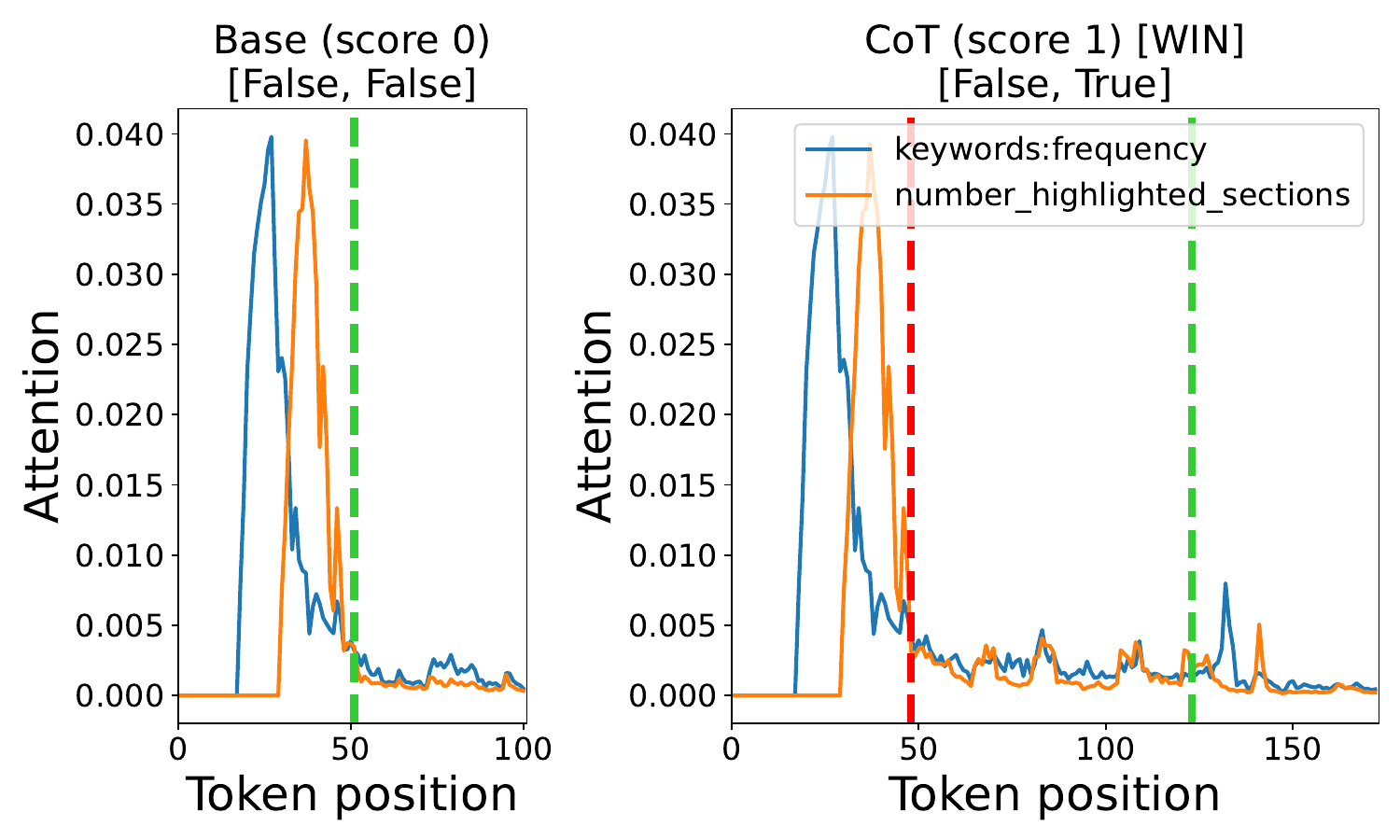}
    % \caption*{Three constraint-attention traces from IFEval (\texttt{Qwen2.5-1.5B-Instruct}) for TIE, LOSE, and WIN outcomes. Each curve corresponds to one constraint span.}
\end{minipage}
\hfill
% Right: ComplexBench
\begin{minipage}[t]{0.485\textwidth}
    \centering
    \textbf{(b) ComplexBench – Attention Traces} \\[0.5em]
    \hspace{-0.315em}
    \includegraphics[width=0.983\linewidth, height=0.55\linewidth]{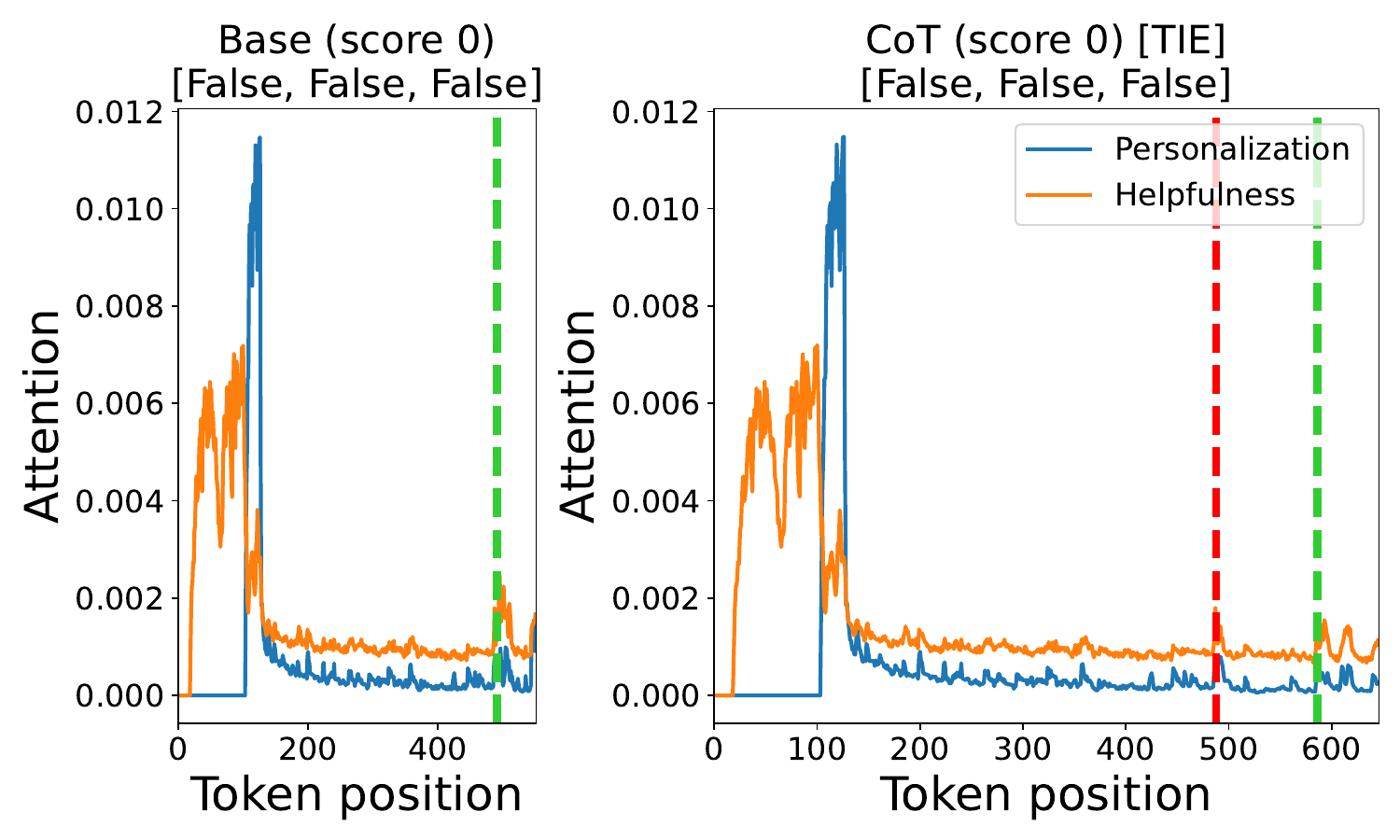}
    \vspace{0.5em}
    \hspace{0.09em}
    \includegraphics[width=0.997\linewidth, height=0.55\linewidth]{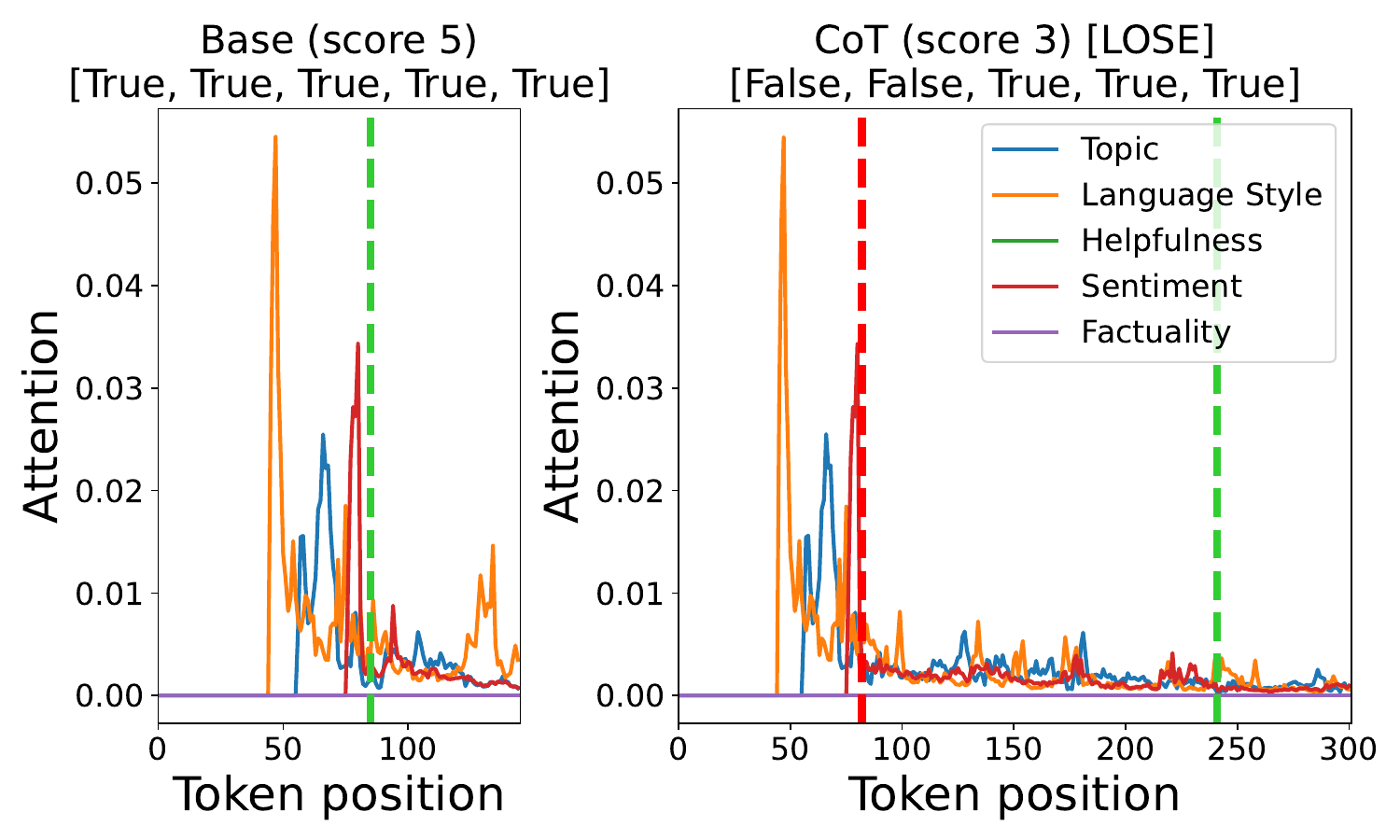}
    \vspace{0.5em}
    \hspace{-0.11em}
    \includegraphics[width=1.01\linewidth, height=0.55\linewidth]{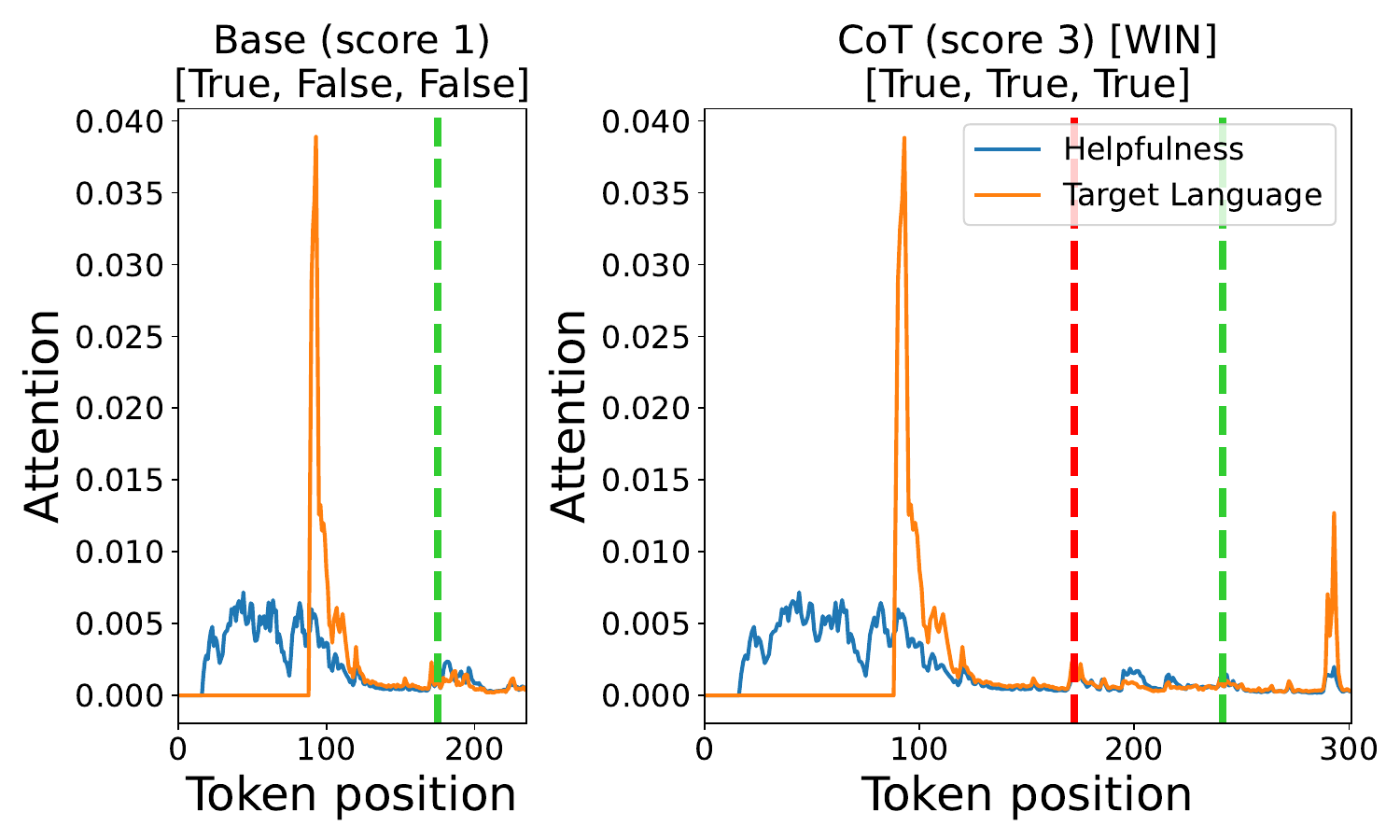}
    % \caption*{Three constraint-attention traces from ComplexBench using the same model. Outcome types: TIE, LOSE, WIN. Reasoning weakens attention in LOSE cases.}
\end{minipage}

\vspace{0.5em}
\caption{
Constraint-attention trace examples for \texttt{Qwen2.5-1.5B-Instruct} across both datasets. From top to bottom, the plots show cases where reasoning leads to a TIE, LOSE, and WIN compared to not using reasoning. The vertical red dashed line marks the start of \textit{Thinking}, and the green dashed line marks the start of the \textit{Answer} during response generation.
}
\label{fig:AttentionTrace-IFEval-Complex-BothQwen}
\end{figure}

\paragraph{Constraint Attention.}
For each textual constraint $c_r$ ($r=1,\dots,R$), we collect the prompt token indices it spans, yielding the index subset
$C_r\subseteq\{1,\dots,T_0\}$. Define the full constraint token set as $C = \bigcup_{r=1}^{R} C_r$. Then we can define the \textit{layer–step constraint attention} (averaged attention to constraint tokens at each layer and step) as 
\[
\alpha^{(l,t)}
= \frac{1}{|C|}\sum_{j\in C} a^{(l,t)}_j .
\]
Following this, we compute the \textbf{layer-averaged constraint attention at step $t$} as:
\begin{equation} \label{eq:layer_avg}
  \bar{\alpha}^{(t)}
  = \frac{1}{L}\sum_{l=0}^{L-1} \alpha^{(l,t)},
\end{equation}
Based on Equation~\ref{eq:layer_avg}, we visualize the trace of constraint-attention during response generation in Figure~\ref{fig:AttentionTrace-IFEval-Complex-BothQwen}, showing results for both IFEval and ComplexBench using \texttt{Qwen2.5-1.5B-Instruct}. Additional examples for other models are provided in Appendix~\ref{sec:Appendix-AttentionTrace}. After reviewing hundreds of samples, we observe a general trend: reasoning flattens the constraint-attention trace. In cases where reasoning degrades performance, constraint attention during the answer phase is generally lower. In contrast, when reasoning improves performance, we often see a bump in attention aligned with the answer segment.

\begin{figure}[ht]
    \centering
    % Row 1
    \begin{subfigure}[b]{0.32\textwidth}
        \includegraphics[width=\linewidth]{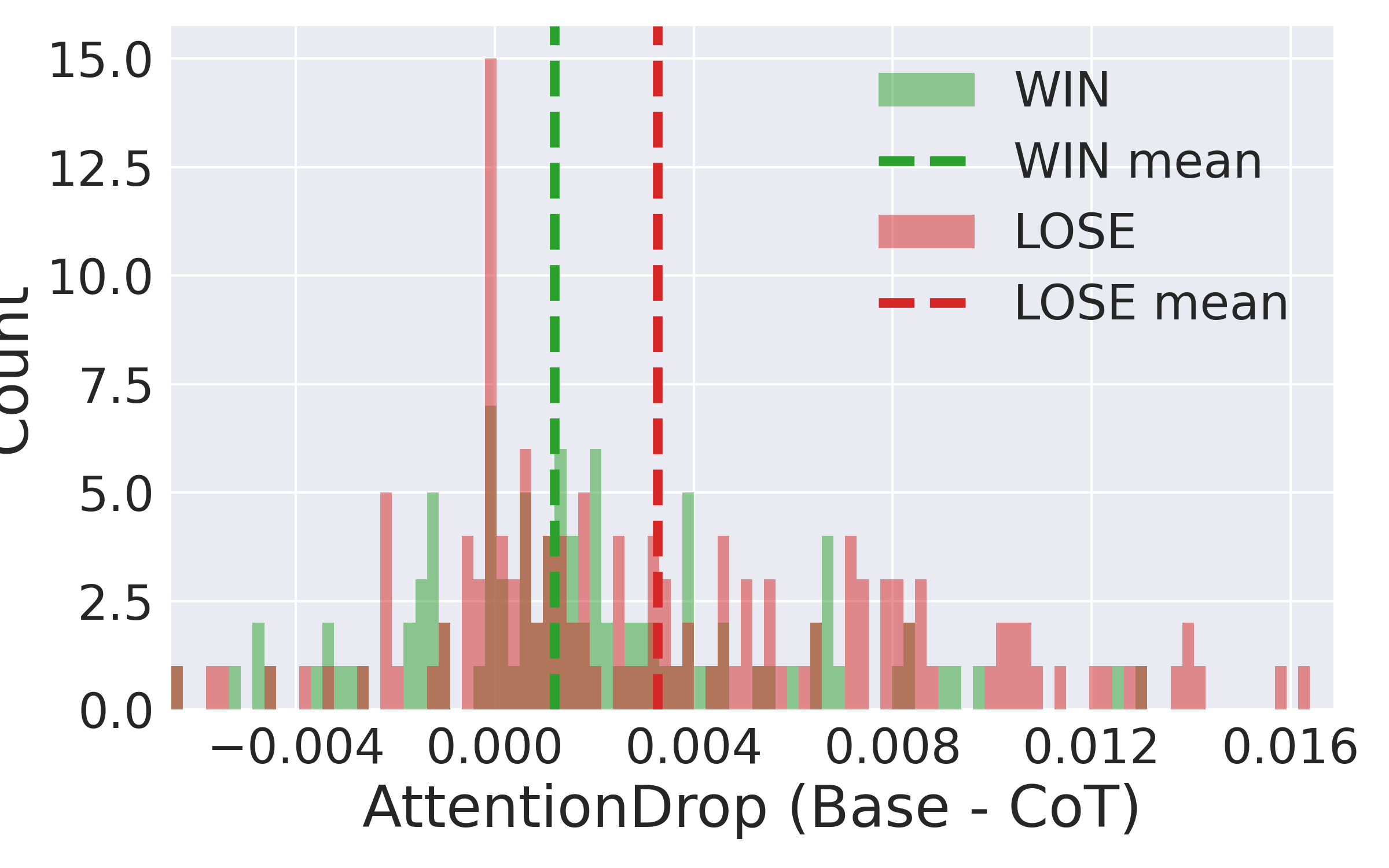}
        \caption{First layer}
    \end{subfigure}
    \hfill
    \begin{subfigure}[b]{0.32\textwidth}
        \includegraphics[width=\linewidth]{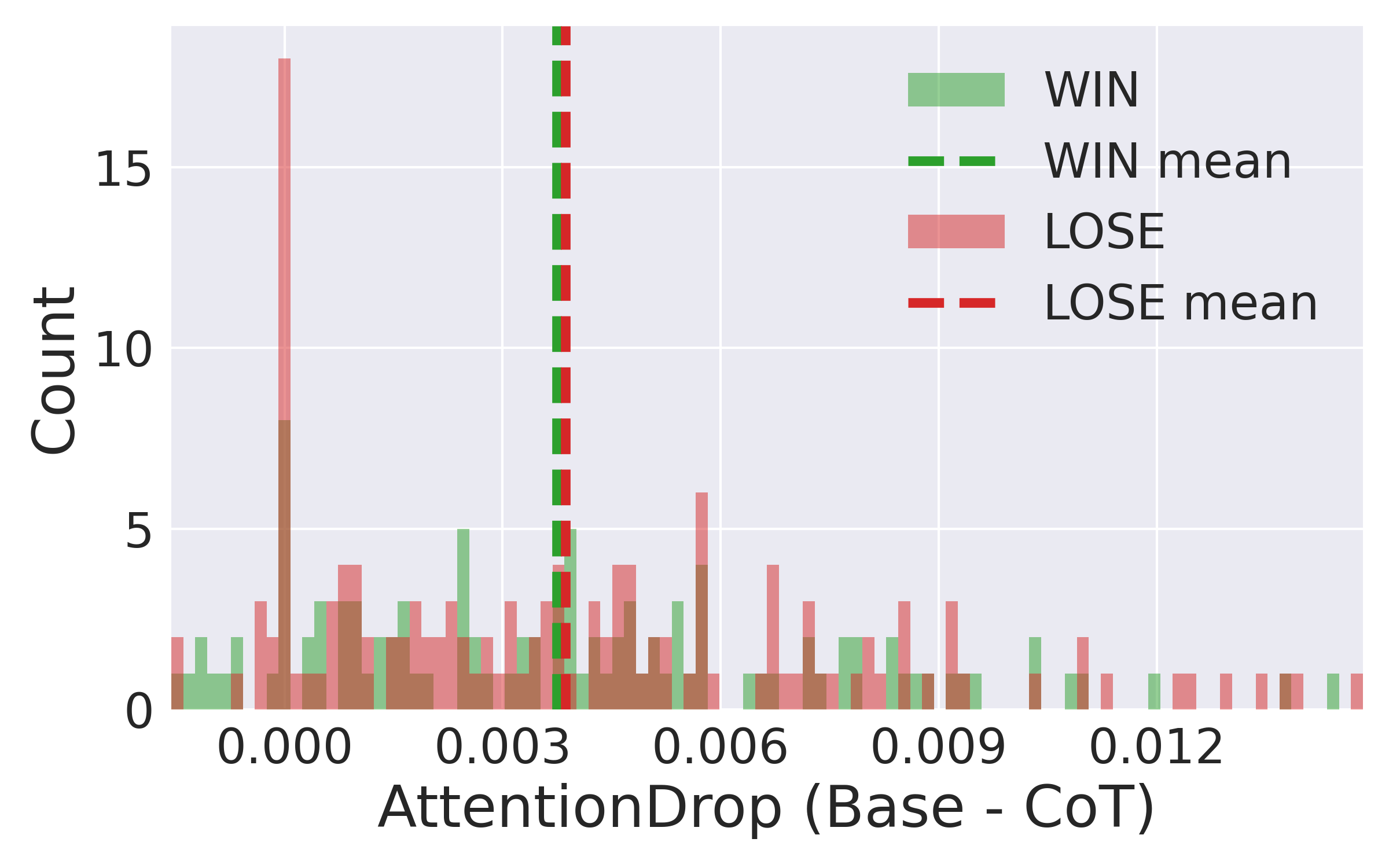}
        \caption{Early-mid layer}
    \end{subfigure}
    \hfill
    \begin{subfigure}[b]{0.32\textwidth}
        \includegraphics[width=\linewidth]{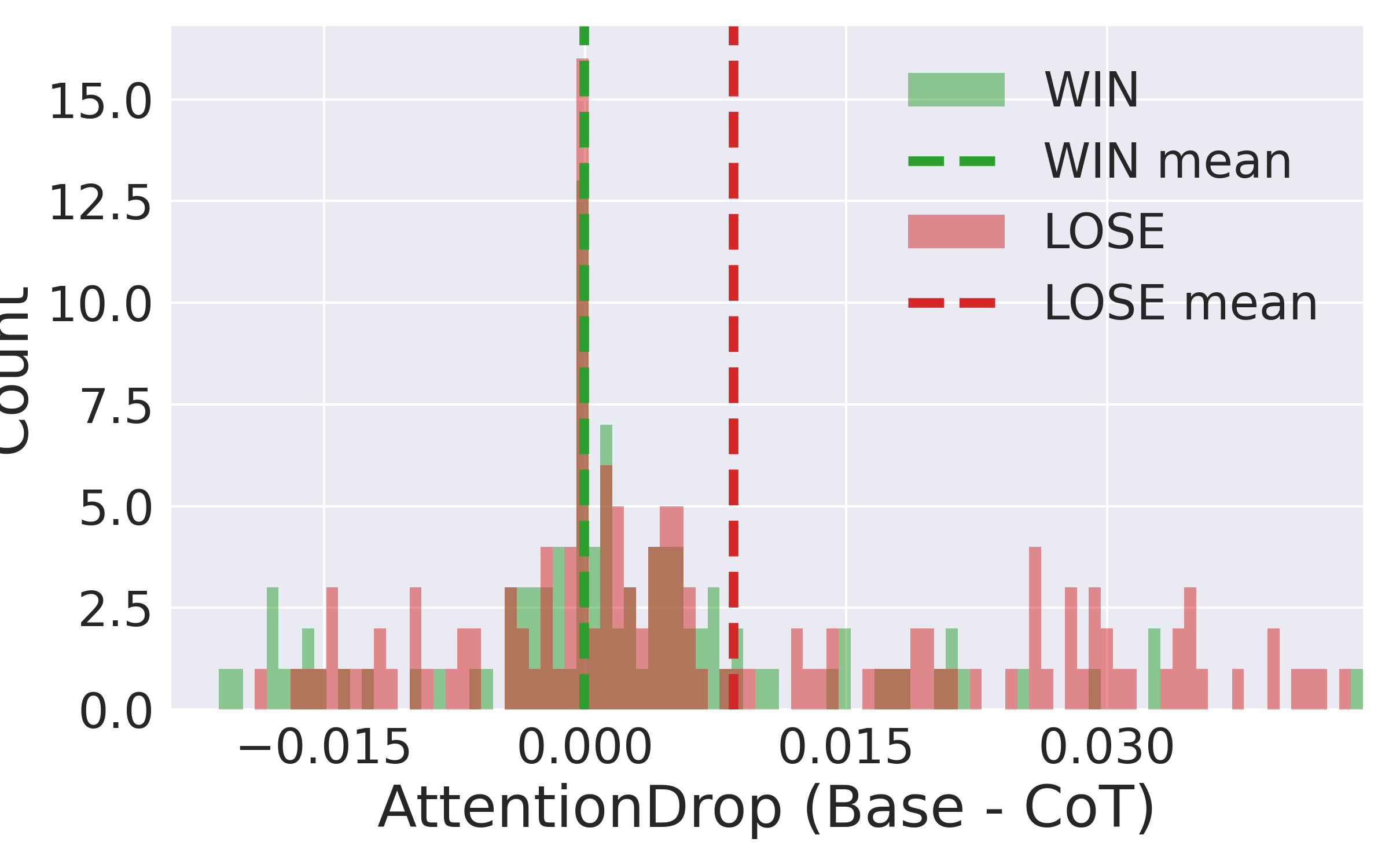}
        \caption{Middle layer}
    \end{subfigure}

    % Row 2
    \begin{subfigure}[b]{0.32\textwidth}
        \includegraphics[width=\linewidth]{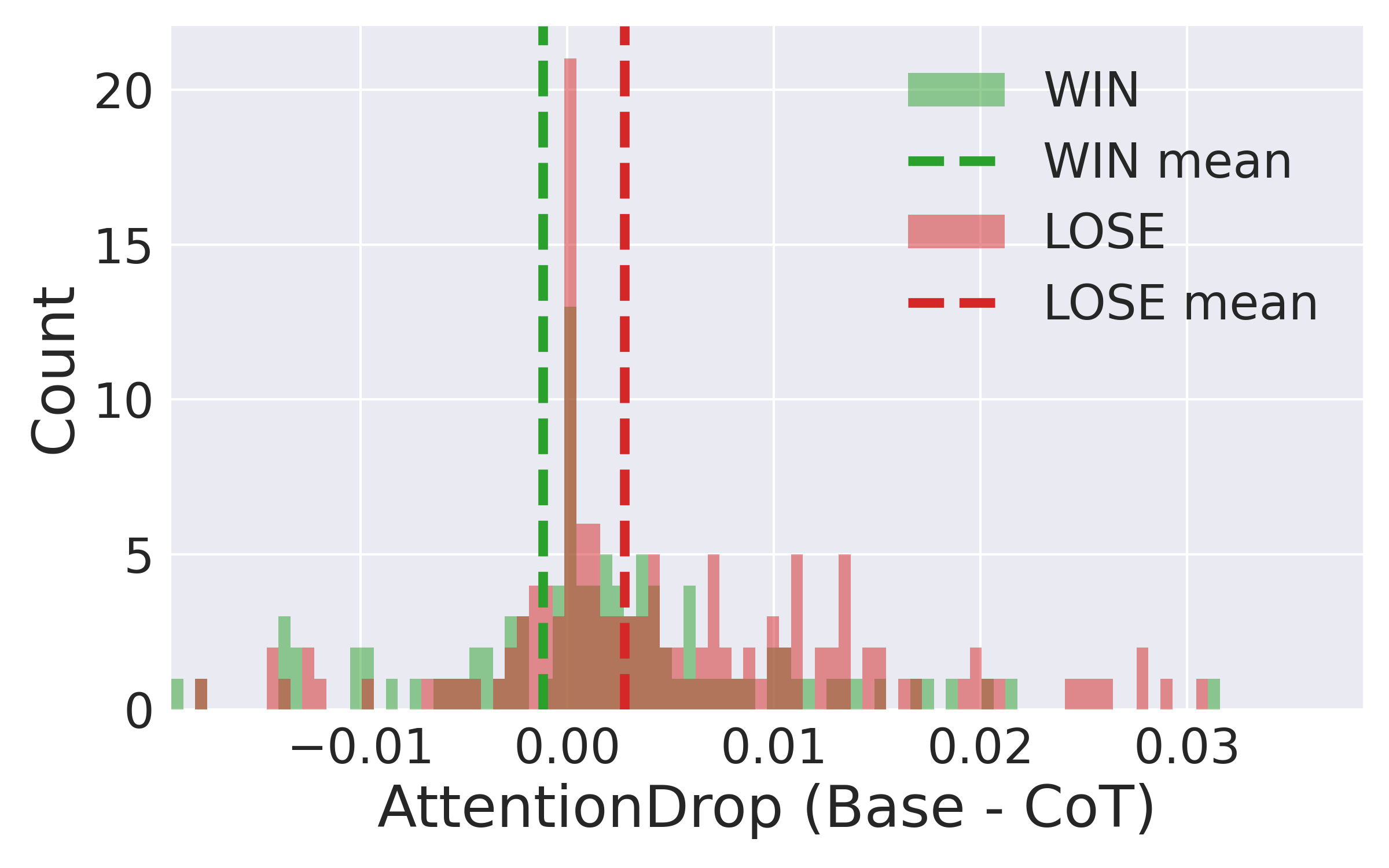}
        \caption{Late-mid layer}
    \end{subfigure}
    \hfill
    \begin{subfigure}[b]{0.32\textwidth}
        \includegraphics[width=\linewidth]{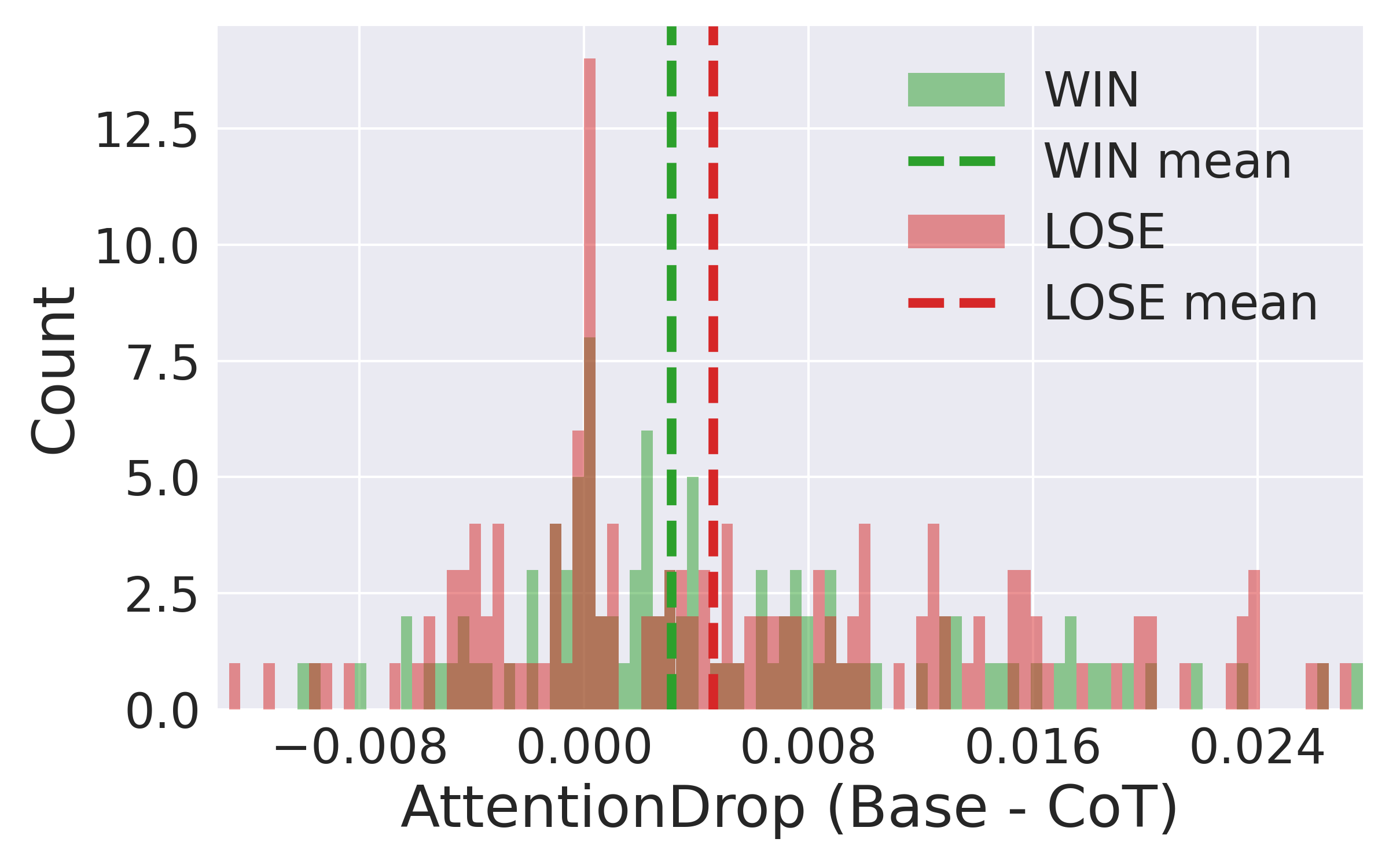}
        \caption{Final layer}
    \end{subfigure}
    \hfill
    \begin{subfigure}[b]{0.32\textwidth}
        \includegraphics[width=\linewidth]{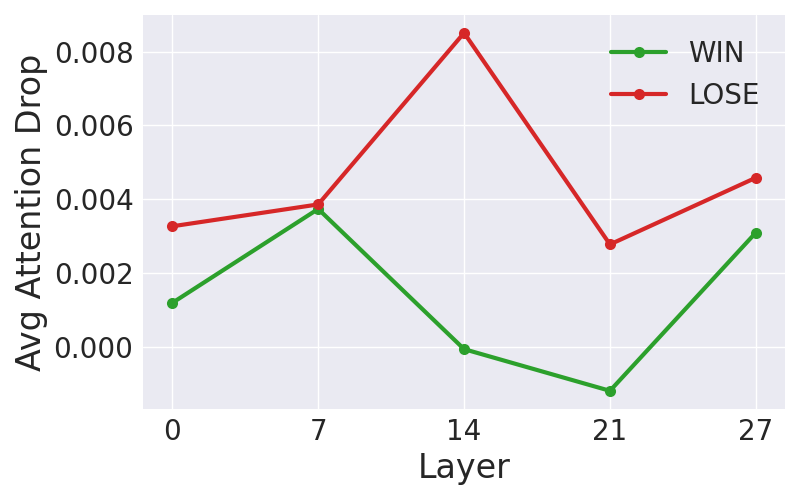}
        \caption{Mean by layer}
    \end{subfigure}
    \caption{
        Within IFEval dataset, drop in constraint attention (Base – CoT) for WIN vs.\ LOSE cases across representative layers of the \texttt{Qwen2.5-1.5B-Instruct} model. On average, the drop is more severe in cases when reasoning loses, compared to not using reasoning.
    }
    \label{fig:AttentionDrop-IFEval-Qwen}
\end{figure}

\paragraph{Quantifying Attention Drop.}
To further quantify these observations, we compute the mean constraint attention across the answer phase. Let $\mathcal{A}$ denote the answer token positions. The average constraint attention at layer $l$ is:
\begin{equation}\label{eq:time_avg}
    \bar{\beta}^{(l)} = \frac{1}{|\calA|} \sum_{t\in\calA} \alpha^{(l,t)}
\end{equation}
We define the \textbf{attention drop} as the difference between base and CoT runs:
\[
\Delta\beta = \bar{\beta}_{\text{Base}} - \bar{\beta}_{\text{CoT}}.
\]
We find that $\Delta\beta > 0$ for most cases. Figure~\ref{fig:AttentionDrop-IFEval-Qwen} shows how this drop varies across layers for WIN vs.\ LOSE cases. On average, the model exhibits larger attention drops in LOSE cases, particularly in early-to-middle layers. This suggests that \textbf{lower constraint attention is predictive of reasoning-induced failures}.

\paragraph{Conclusion.}
Our analysis reveals that explicit reasoning often reduces attention to constraint-relevant parts of the prompt. This diminished awareness increases the risk of violating instructions. Thus, while reasoning is intended to improve task performance, it can unintentionally shift model focus away from critical constraints and harm instruction adherence. In addition, we also explored whether longer reasoning tends to degrade performance and found that \textbf{reasoning length does not meaningfully correlate with instruction-following effectiveness} (Appendix~\ref{sec:Appendix-ThinkingLengthCorrelation}).

%%%%%%%%%%%%%%%%%%%%%%%%%%%%%%%%%%%%%%%%%%%%%%%%%%%%%%%%%%%%%%%%%%%%%%%%%%%%%%%%%%%%%%%%%%%%%%%%%%%%%%%%%%%%%%%%%%%%%%%%%%%%%%%%%%%%%%%%%%%%%%%%%%%%%%%%%%%%%%%%%%%%%%%%%%%%%%%%%%%%%%%%%%%%%%%%%%%%%%%%%%%%%%%%%%%%%%%%%%%%%%%%%%%%%%%%%%%%%%%%%%%%%%%%%%%%%%%%%%%%%%%%%%%%%%%%%%%%%%%%%%%%%%%%%%%%%%%%%%%%%%%%%%%%%%%%%%%%%%%%%%%%%%%%%%%%%%%%%%%%%%%%%%%%%%%%%%%%%%%%%%%%%%%%%%%%%%%%%%%%%%%%%%%%%%%%%%%%%
\section{Mitigating Reasoning-Induced Failures in Instruction Following}\label{sec:Methods}

We introduce and evaluate four strategies designed to mitigate the performance degradation caused by explicit reasoning (via CoT) in instruction-following tasks. To the best of our knowledge, this is the first work to systematically identify and address this issue. Hence, in the absence of existing baselines, we directly compare the effectiveness of our proposed methods. Results are presented in Table~\ref{tab:IFEval-ComplexBench-results} and Figure~\ref{fig:MainResultsPlot-Combined}, which also include performance differences relative to the CoT baseline.
%%%%%%%%%%%%%%%%%%%%%%%%%%%%%%%%%%%%%%%%%%%%%%%%%%%%%%%%%%%%%%%%%%%%%%%%%%%%%%%%%%%%%%%%%%%%%%%%%%%%%%
\subsection{Method 1: Few-Shot In-Context Learning}\label{subsec:Method-FewShot}
We apply in-context learning~\citep{brown2020language} by prepending carefully selected few-shot examples to each instruction. These examples are derived from representative failure cases identified in our case study (Section~\ref{sec:Appendix-CaseStudy}) and manually revised to fully satisfy all constraints. Each example includes an \texttt{Instruction}, along with a corrected \texttt{Thinking} and \texttt{Answer}. Full examples are provided in Appendix~\ref{sec:Appendix-FewShotExamples}.

\textbf{Results:} As shown in Table~\ref{tab:IFEval-ComplexBench-results}, this method yields only modest improvements. We attribute the limited gains to several factors. First, due to the token length limit and the substantial size of each example (including prompt, reasoning, and answer), we were only able to include \textbf{a limited number of examples}: four examples for IFEval and three for ComplexBench. Second, because examples were \textbf{sourced from outputs of certain models, potentially introducing bias}. These constraints likely reduce the effectiveness of the few-shot strategy and limit its generalizability.
%%%%%%%%%%%%%%%%%%%%%%%%%%%%%%%%%%%%%%%%%%%%%%%%%%%%%%%%%%%%%%%%%%%%%%%%%%%%%%%%%%%%%%%%%%%%%%%%%%%%%%
\subsection{Method 2: Self-Reflection}\label{subsec:Method-SelfReflection}
This method performs a two-step process: the model first generates an initial response with thinking, and then performs a second inference where it reflects on its own reasoning and answer. If the model deems the initial response satisfactory, it retains it as the final output; otherwise, it revises the response and outputs the updated version. The prompt used for self-reflection is detailed in Appendix~\ref{sec:Appendix-Prompts}.

\textbf{Results:} As shown in Table~\ref{tab:IFEval-ComplexBench-results}, self-reflection yields strong improvements on IFEval—enhancing performance in 11 out of 14 models, and achieving the best results among all four methods for 7 models. However, we observe notable performance degradation on weaker models such as \texttt{Llama-3.2-1B-Instruct} and \texttt{Qwen2.5-1.5B-Instruct}. We hypothesize that this is because \textbf{self-reflection relies on a model’s ability to critique and improve its own outputs, a capability that may be underdeveloped in weaker models}. On ComplexBench, which contains more challenging and compositional instructions, self-reflection proves less effective—leading to performance drops in 10 out of 14 models. This suggests that \textbf{self-reflection is more suitable for simpler instruction-following tasks}, and may be counterproductive when applied to complex scenarios. Additionally, a key limitation of this method is its increased computational cost, as it requires two forward passes per query.
%%%%%%%%%%%%%%%%%%%%%%%%%%%%%%%%%%%%%%%%%%%%%%%%%%%%%%%%%%%%%%%%%%%%%%%%%%%%%%%%%%%%%%%%%%%%%%%%%%%%%%
\subsection{Method 3: Self-Selective Reasoning}\label{subsec:Method-SelfSelectiveReasoning}
This method enables the model to decide dynamically whether to perform explicit reasoning. Specifically, we prompt the model to assess, based on the instruction alone, whether reasoning (via CoT) is necessary (see prompt in Appendix~\ref{sec:Appendix-Prompts}). If the model deems reasoning helpful, it proceeds with step-by-step thinking; otherwise, it directly generates an answer without reasoning.

\textbf{Results:} This approach yields moderate gains on IFEval, improving performance in 10 out of 14 models, and shows stronger results on ComplexBench, improving all models and achieving the best performance among all four methods in 6 models. In Appendix~\ref{sec:Appendix-SelfSelectiveReasoning}, we further analyze the model's decision-making behavior by comparing its self-selected reasoning decisions to ground-truth labels based on actual performance differences between CoT and non-CoT responses. We find that while the model tends to have high recall (correctly identifying most cases where reasoning helps) it suffers from low precision, often applying reasoning even when it is not necessary.

%%%%%%%%%%%%%%%%%%%%%%%%%%%%%%%%%%%%%%%%%%%%%%%%%%%%%%%%%%%%%%%%%%%%%%%%%%%%%%%%%%%%%%%%%%%%%%%%%%%%%%
\subsection{Method 4: Classifier-Selective Reasoning}\label{subsec:Method-ClassifierSelectiveReasoning}
Instead of relying on the model’s internal judgment, this method uses an external binary classifier to determine whether CoT reasoning should be applied. For each target model, we train a separate classifier to predict whether using CoT leads to improved instruction-following performance. Labels are assigned on a per-sample basis by comparing the constraint satisfaction scores of CoT and non-CoT responses. Training details are provided in Appendix~\ref{sec:Appendix-TrainingClassifier}.

The classifier is implemented using \texttt{Qwen2.5-7B-Instruct} as the backbone and trained for 3 epochs with a learning rate of 1e-5. Both the backbone model and training hyperparameters are selected via grid search. We split the dataset evenly, using 50\% of the samples for training and the remaining 50\% to evaluate downstream mitigation effectiveness. Validation accuracy typically ranges between 0.75 and 0.92.

\textbf{Results:} This method proves highly effective, improving performance for nearly all models on both benchmarks, with only a minor drop for \texttt{DeepSeek-V3} on IFEval. \textbf{For about half of the models, it achieves the best overall performance among the four strategies.} Nonetheless, its primary drawback is the model-specific training requirement: each model needs its own classifier, which results in additional overhead.

%%%%%%%%%%%%%%%%%%%%%%%%%%%%%%%%%%%%%%%%%%%%%%%%%%%%%%%%%%%%%%%%%%%%%%%%%%%%%%%%%%%%%%%%%%%%%%%%%%%%%%
\subsection{Summary and Recommended Pipeline}
Each mitigation strategy exhibits distinct strengths and weaknesses depending on model capacity and instruction complexity. Based on our findings, we propose the following decision pipeline: first, estimate the complexity of the instruction—either through simple heuristics or a trained classifier. For simpler tasks (e.g., IFEval), we recommend \textbf{Self-Reflection} or \textbf{Classifier-Selective Reasoning}; for more complex or compositional tasks (e.g., ComplexBench), \textbf{Self-Selective Reasoning} or \textbf{Classifier-Selective Reasoning} is more effective. Overall, \textbf{Classifier-Selective Reasoning consistently delivers the best overall performance across both benchmarks}, albeit at the cost of model-specific training.

%%%%%%%%%%%%%%%%%%%%%%%%%%%%%%%%%%%%%%%%%%%%%%%%%%%%%%%%%%%%%%%%%%%%%%%%%%%%%%%%%%%%%%%%%%%%%%%%%%%%%%%%%%%%%%%%%%%%%%%%%%%%%%%%%%%%%%%%%%%%%%%%%%%%%%%%%%%%%%%%%%%%%%%%%%%%%%%%%%%%%%%%%%%%%%%%%%%%%%%%%%%%%%%%%%%%%%%%%%%%%%%%%%%%%%%%%%%%%%%%%%%%%%%%%%%%%%%%%%%%%%%%%%%%%%%%%%%%%%%%%%%%%%%%%%%%%%%%%%%%%%%%%%%%%%%%%%%%%%%%%%%%%%%%%%%%%%%%%%%%%%%%%%%%%%%%%%%%%%%%%%%%%%%%%%%%%%%%%%%%%%%%%%%%%%%%%%%%%
\section{Conclusion}\label{sec:conclusion}

One limitation of our study is its exclusive focus on instruction-following tasks. While we suspect that reasoning could similarly degrade performance in other domains, exploring these possibilities is left for future work. In our study, we identified and systematically explored an unexpected phenomenon: explicit reasoning through Chain-of-Thought prompting can negatively impact the instruction-following abilities of large language models. Through extensive evaluation across two comprehensive benchmarks (IFEval and ComplexBench), we demonstrated consistent performance degradation when models employed explicit reasoning. Our detailed analysis, involving manual case studies and attention-based examinations, provided insights into why reasoning negatively affects instruction adherence. We found that reasoning can divert the model’s focus from constraint-related tokens, resulting in overlooked or violated instructions.

To address this issue, we introduce and evaluate four mitigation strategies: in-context learning, self-reflection, self-selective reasoning, and classifier-selective reasoning. Our experiments show that selective reasoning, especially classifier-based approaches, can substantially recover lost performance, with classifier-selective reasoning achieving the most consistent improvements across both datasets. 

We believe this is the first systematic investigation into reasoning-induced failures in instruction-following tasks. We hope our findings motivate further investigation into reasoning tradeoffs and contribute to building models that reason more selectively and effectively.

%%%%%%%%%%%%%%%%%%%%%%%%%%%%%%%%%%%%%%%%%%%%%%%%%%%%%%%%%%%%%%%%%%%%%%%%%%%%%%%%%%%%%%%%%%%%%%%%%%%%%%%%%%%%%%%%%%%%%%%%%%%%%%%%%%%%%%%%%%%%%%%%%%%%%%%%%%%%%%%%%%%%%%%%%%%%%%%%%%%%%%%%%%%%%%%%%%%%%%%%%%%%%%%%%%%%%%%%%%%%%%%%%%%%%%%%%%%%%%%%%%%%%%%%%%%%%%%%%%%%%%%%%%%%%%%%%%%%%%%%%%%%%%%%%%%%%%%%%%%%%%%%%%%%%%%%%%%%%%%%%%%%%%%%%%%%%%%%%%%%%%%%%%%%%%%%%%%%%%%%%%%%%%%%%%%%%%%%%%%%%%%%%%%%%%%%%%%%%
\bibliographystyle{plainnat}
\bibliography{reference}

%%%%%%%%%%%%%%%%%%%%%%%%%%%%%%%%%%%%%%%%%%%%%%%%%%%%%%%%%%%%%%%%%%%%%%%%%%%%%%%%%%%%%%%%%%%%%%%%%%%%%%%%%%%%%%%%%%%%%%%%%%%%%%%%%%%%%%%%%%%%%%%%%%%%%%%%%%%%%%%%%%%%%%%%%%%%%%%%%%%%%%%%%%%%%%%%%%%%%%%%%%%%%%%%%%%%%%%%%%%%%%%%%%%%%%%%%%%%%%%%%%%%%%%%%%%%%%%%%%%%%%%%%%%%%%%%%%%%%%%%%%%%%%%%%%%%%%%%%%%%%%%%%%%%%%%%%%%%%%%%%%%%%%%%%%%%%%%%%%%%%%%%%%%%%%%%%%%%%%%%%%%%%%%%%%%%%%%%%%%%%%%%%%%%%%%%%%%%%
% \newpage 
% \appendix

% --- Appendix Section ---
\newpage 
\appendix
\label{sec:append}

\hypersetup{linkcolor=black} % change
\startcontents[appendix]
\printcontents[appendix]{ }{0}{
    \section*{Appendix}
}
\hypersetup{linkcolor=black} % change back

\vspace{2mm}

\clearpage

%%%%%%%%%%%%%%%%%%%%%%%%%%%%%%%%%%%%%%%%%%%%%%%%%%%%%%%%%%%%%%%%%%%%%%%%%%%%%%%%%%%%%%%%%%%%%%%%%%%%%%%%%%%%%%%%%%%%%%%%%%%%%%%%%%%%%%%%%%%%%%%%%%%%%%%%%%%%%%%%%%%%%%
% \section{Limitations and Future Directions} \label{sec:Limitations}

%%%%%%%%%%%%%%%%%%%%%%%%%%%%%%%%%%%%%%%%%%%%%%%%%%%%%%%%%%%%%%%%%%%%%%%%%%%%%%%%%%%%%%%%%%%%%%%%%%%%%%%%%%%%%%%%%%%%%%%%%%%%%%%%%%%%%%%%%%%%%%%%%%%%%%%%%%%%%%%%%%%%%%
\section{Experiments: Additional Results}\label{sec:Appendix-Experiments-AdditionalResults}

The plots for main results in Table~\ref{tab:IFEval-ComplexBench-results} are presented in 
Figures~\ref{fig:MainResultsPlot-Combined} below.

\begin{figure}[ht]
    \centering
    \begin{subfigure}[t]{0.76\linewidth}
        \centering
        \includegraphics[width=\linewidth]{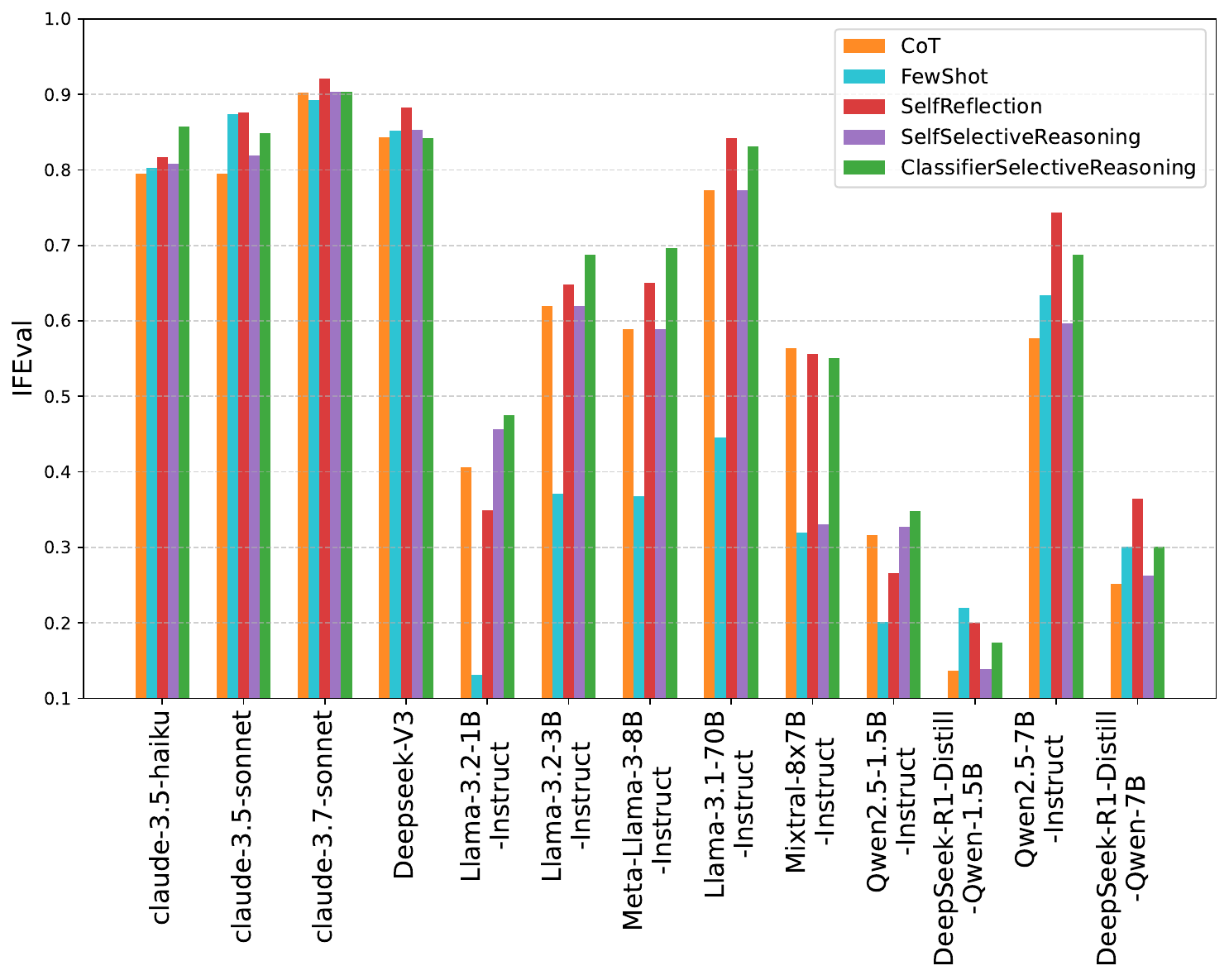}
        % \caption{IFEval results.}
        % \label{fig:MainResultsPlot-IFEval}
    \end{subfigure}
    \begin{subfigure}[t]{0.76\linewidth}
        \centering
        \includegraphics[width=\linewidth]{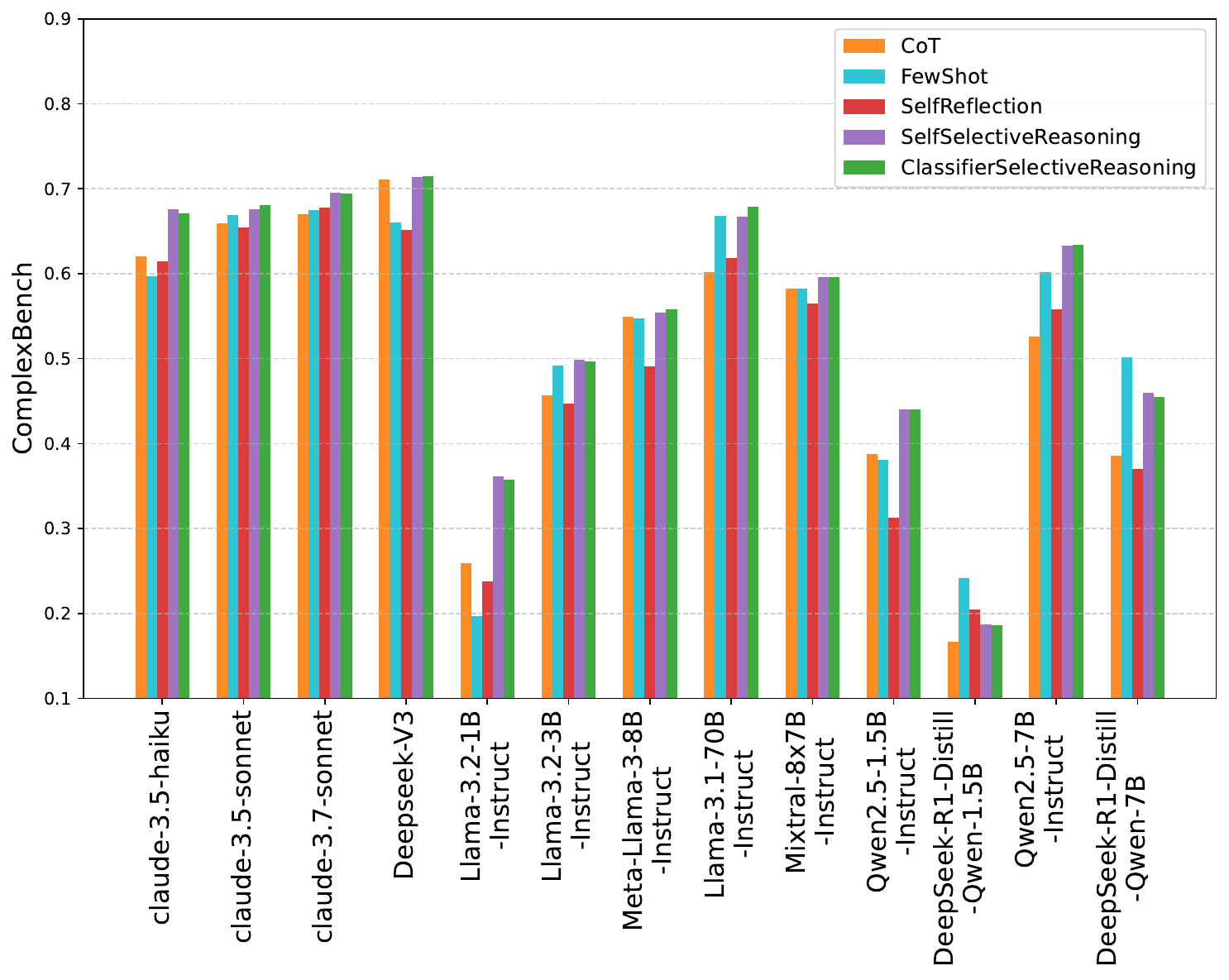}
        % \caption{ComplexBench results.}
        % \label{fig:MainResultsPlot-ComplexBench}
    \end{subfigure}
    \caption{Visualization of instruction-following accuracy across models and methods on IFEval and ComplexBench.}
    \label{fig:MainResultsPlot-Combined}
\end{figure}

% For the comparison of RLLM vs. their non-reasoning counterparts, particularly Claude3.7-Sonnet v.s.  Claude3.7-Sonnet-Think and DeepSeek-V3 v.s. DeepSeek-R1, the results are presented in Table~\ref{tab:RLLM-Comparison-Table} below. We also observe a drop of performance for both datasets and both pairs of models.

% \begin{table}[ht]
% \centering
% \renewcommand{\arraystretch}{1}
% \setlength{\tabcolsep}{6pt}
% \begin{tabular}{lcccc}
% \toprule
% & \textbf{Claude-3.7-Sonnet} & \textbf{Claude-3.7-Sonnet-Think} & \textbf{DeepSeek-V3} & \textbf{DeepSeek-R1} \\
% \midrule
% \textbf{IFEval}       & \green{90.57} & \red{90.20} & \green{85.21} & \red{83.30} \\
% \textbf{ComplexBench} & \green{69.79} & \red{69.03} & \green{71.22} & \red{71.09} \\
% \bottomrule
% \end{tabular}
% \vspace{0.5em}
% \caption{Comparison of reasoning-enabled models vs.\ their non-reasoning counterparts. \green{Green} indicates higher-performing model in each pair; \red{red} indicates lower.}
% \label{tab:RLLM-Comparison-Table}
% \end{table}

%%%%%%%%%%%%%%%%%%%%%%%%%%%%%%%%%%%%%%%%%%%%%%%%%%%%%%%%%%%%%%%%%%%%%%%%%%%%%%%%%%%%%%%%%%%%%%%%%%%%%%%%%%%%%%%%%%%%%%%%%%%%%%%%%%%%%%%%%%%%%%%%%%%%%%%%%%%%%%%%%%%%%%
\section{Case Study}\label{sec:Appendix-CaseStudy}
We present full examples illustrating the four scenarios discussed in Section~\ref{subsec:CaseStudy}, using outputs from \texttt{Claude3.7-Sonnet} and its reasoning-enabled variant \texttt{Claude3.7-Sonnet-Think} for direct comparison. To ensure clarity, we focus on examples from IFEval, as prompts and responses in ComplexBench are often too long and complex for a concise presentation. Each example includes the instruction, its associated constraints, the response generation model, responses with and without reasoning along with their respective constraint satisfaction scores, and a brief analysis explaining how reasoning contributed to the performance difference.

\textbf{Reasoning Helps}
\begin{enumerate}
    \item Reasoning Helps Meet Formatting or Structural Requirements.
    \item Reasoning Helps Satisfy Lexical Constraints.
\end{enumerate}

%%%%%%%%%%%%%%%%%%%%%%%%%%%%%%%%%%%%%%%%%%%%%%%%%%%%%%%%%%%%
% EXAMPLE 1 - IFEval-Claude3.7-Row102
\begin{tcolorbox}[colback=blue!5!white, colframe=blue!60!white, title=Example 1: Reasoning Helps (Meet Formatting or Structural Requirements), left=10mm, breakable]
\small

\textbf{Instruction:} Name a new fashion company that young people might like, and give it a name with multiple meanings. Put the name in double angular brackets, such as \verb|<<name>>|. Repeat the request before responding.

\textbf{Constraints:} 
\begin{itemize}
    \item \texttt{repeat\_prompt} – the model must repeat the instruction before answering
    \item \texttt{title\_format} – the name must appear in \verb|<<>>| brackets
\end{itemize}

\textbf{Response Model:} \texttt{Claude3.7-Sonnet}

\bigskip \hrule \bigskip
\textbf{Response without reasoning (constraint satisfaction: [\red{False}, \green{True}]):}

\textbf{Answer:}

I'll help you create a name for a new fashion company that young people might like, with multiple meanings. I'll put the name in double angular brackets as requested.

<<Shift>>

This name works on multiple levels:
1. "Shift" suggests change and transformation, appealing to young people's desire to express themselves and evolve their style
2. It references the shift key on keyboards, connecting to digital culture
3. In fashion terminology, a shift is also a type of dress, giving it direct industry relevance
4. It implies movement between states or trends, reflecting how young consumers quickly adapt to new styles

The brand could position itself as adaptive, fluid fashion for a generation that's constantly reinventing itself.

\bigskip \hrule \bigskip
\textbf{Response with reasoning (constraint satisfaction: [\green{True}, \green{True}]):}

\textbf{Thinking:}  

I need to name a new fashion company that would appeal to young people. The name should have multiple meanings or interpretations. And I need to present the name in double angular brackets like <<name>>.

The fashion company name should be:
- Appealing to young demographics
- Have multiple layers of meaning
- Be memorable and marketable
- Work well for a fashion brand

Some ideas:
1. "Flux" - Represents constant change/movement (fashion trends always changing) and also sounds like "lux" (luxury)
2. "Thread" - Both sewing thread and connecting thread/storyline
3. "Shift" - Changing clothes/shifting perspectives
4. "Alter" - To change clothing/alternative style/altar (elevation)
5. "Cipher" - Code/hidden meaning and also sounds like "cypher" (creative expression in hip-hop culture)

I think "Cipher" works well because:
- It has the dual meaning of code/hidden message and connection to creative expression
- It sounds modern and intriguing
- Young people appreciate brands with depth and cultural connections
- It works across various fashion styles

\textbf{Answer:}

Name a new fashion company that young people might like, and give it a name with multiple meanings. Put the name in double angular brackets, such as <<name>>.

For a new fashion company appealing to young people, I suggest:

<<Thread>>

This name works on multiple levels:
- Thread literally refers to fabric/clothing materials
- It suggests connection and community (threading people together)
- In digital culture, it references conversation threads on social media
- It hints at continuity and storytelling - threading narratives through fashion
- It's concise, modern, and easy to brand

\bigskip \hrule \bigskip

\textbf{Case Analysis:}  
The original instruction includes two constraints: repeating the prompt and formatting the name in \verb|<<>>| brackets. Without reasoning, the model skipped repeating the prompt, failing the first constraint. With reasoning, the model:
\begin{itemize}
    \item Broke down the task and repeated the instruction (satisfying constraint 1)
    \item Explicitly evaluated multiple name candidates against the constraints
    \item Chose a final name that satisfied both format and semantic richness
\end{itemize}

The reasoning process acted as a structured checklist, ensuring no requirements were overlooked. This led to full constraint satisfaction and a more deliberate, accurate response.

\end{tcolorbox}

\begin{tcolorbox}[colback=blue!5!white, colframe=blue!60!white, title=Example 2: Reasoning Helps (Satisfy Lexical Constraint), left=10mm, breakable]
\small
\textbf{Instruction:} Write a cover letter for a job and end with exactly \texttt{"Call me at 631-481-4867"}\\
No other words should follow that phrase.

\textbf{Constraint:} \texttt{startend:end\_checker} (must end with \texttt{Call me at 631-481-4867} exactly)

\textbf{Model:} \texttt{Claude3.7-Sonnet}

\bigskip \hrule \bigskip
\textbf{Response without reasoning (constraint satisfaction: [\red{False}]):}

\textbf{Answer:}

\begin{quote}
\texttt{\# Professional Cover Letter}

[Your Name]  
[Your Address]  
[City, State ZIP]  
[Your Email]  

[Date]  

[Hiring Manager's Name]  
[Company Name]  
[Company Address]  
[City, State ZIP]  

Dear [Hiring Manager's Name],

I am writing to express my interest in the [Position Title] role at [Company Name] as advertised on [where you found the job posting]. With my background in [relevant field] and demonstrated success in [relevant accomplishment], I am confident in my ability to make valuable contributions to your team.

Throughout my career, I have developed strong skills in [relevant skill \#1], [relevant skill \#2], and [relevant skill \#3]. In my previous role at [Previous Company], I [specific achievement with measurable results], which [explain positive outcome for the company].

What particularly attracts me to [Company Name] is your commitment to [something specific about the company you admire]. I am impressed by your recent [project/innovation/achievement], and I am excited about the possibility of contributing to similar initiatives.

I believe my experience aligns perfectly with the qualifications you seek. My [specific relevant experience] has prepared me well for the challenges of this position, and I am eager to bring my [specific strength] to your team.

I would welcome the opportunity to discuss how my background, skills and enthusiasm would be an ideal fit for this position. Call me at 631-481-4867.
\end{quote}

\bigskip \hrule \bigskip
\textbf{Response with reasoning (constraint satisfaction: [\textcolor{teal}{True}]):}

\textbf{Thinking:}
\begin{quote}
I need to write a professional cover letter for a job application. The letter should include standard cover letter elements such as an introduction, explanation of qualifications, and a closing. The specific constraint is that I must end the letter with exactly "Call me at 631-481-4867" and no other words should follow that phrase. This means this phone number should be the very last text in the letter, with no additional closing phrases like "Sincerely," "Thank you," or a name signature after it.
\end{quote}

\textbf{Answer:}

\begin{quote}
\texttt{\# Professional Cover Letter}

[Your Name]  
[Your Address]  
[City, State ZIP]  
[Your Email]  

[Date]  

[Hiring Manager's Name]  
[Company Name]  
[Company Address]  
[City, State ZIP]  

Dear [Hiring Manager's Name],

I am writing to express my interest in the [Position Title] role at [Company Name], as advertised on [where you found the job posting]. With my background in [relevant field] and passion for [industry/company focus], I am excited about the opportunity to contribute to your team.

Throughout my career, I have developed strong skills in [key skill 1], [key skill 2], and [key skill 3], which align perfectly with the qualifications you're seeking. In my previous role at [Previous Company], I [specific achievement with metrics if possible], demonstrating my ability to [relevant skill for the new position].

I am particularly drawn to [Company Name] because of [specific reason - company values, recent project, reputation, etc.]. Your commitment to [something the company is known for] resonates with my professional goals and personal values. I am confident that my experience in [relevant experience] would allow me to make meaningful contributions to your team from day one.

I would welcome the opportunity to discuss how my background, skills, and enthusiasms could benefit [Company Name]. Thank you for considering my application. Call me at 631-481-4867
\end{quote}

\bigskip \hrule \bigskip
\textbf{Case Analysis:} This case tests whether the model can respect an exact lexical constraint: ending the letter with the phrase \texttt{"Call me at 631-481-4867"} with nothing after. The response \textbf{without reasoning} appended a period at the end — violating the constraint. The response \textbf{with reasoning} correctly omitted the final period, satisfying the constraint. This is because the model explicitly reasoned about the constraint in advance and realized that even standard punctuation must be omitted. \textbf{Conclusion:} Explicit reasoning helped the model override its default behavior of ending sentences with punctuation, thus enabling it to exactly satisfy a lexical constraint that contradicts writing conventions.
\end{tcolorbox}

\textbf{Reasoning Hurts}
\begin{enumerate}
    \item Reasoning Over-Focuses on High-Level Content and Forgets Simple Mechanical Constraints.
    \item Reasoning Fails Because It Adds ``Helpful'' Elements That Break Constraints.
\end{enumerate}

%%%%%%%%%%%%%%%%%%%%%%%%%%%%%%%%%%%%%%%%%%%%%%%%%%%%%%%%%%%%
% EXAMPLE 3 - IFEval-Claude3.7-Row325

\begin{tcolorbox}[colback=blue!5!white, colframe=blue!60!white, title=Example 3: Reasoning Hurts (Over-Focuses on High-Level Content and Forgets Simple Mechanical Constraints), left=10mm, breakable]
\small
\textbf{Instruction:} Write a 1/2-page print ad for a hotel booking product aimed at frequent travelers. Include a headline and a call to action.\\
\textbf{Additional requirement:} Do not use any commas in your response. Before saying anything or giving your answer, please repeat the exact entire request above.

\textbf{Constraints:}
\begin{itemize}
    \item \texttt{repeat\_prompt}
    \item \texttt{punctuation:no\_comma}
\end{itemize}

\textbf{Model:} \texttt{Claude3.7-Sonnet}

\bigskip \hrule \bigskip
\textbf{Response without reasoning (constraint satisfaction: [\green{True}, \green{True}]):}

\textbf{Answer:}
\begin{quote}
We are a company that sells a product that makes it easy to find and book a hotel room. We are looking for a print ad that will be placed in a magazine that is aimed at people who travel a lot. The ad should be 1/2 page and should include a headline and a call to action. Please do not use any commas in your response.

Here's a print ad concept for your hotel booking product:

\textbf{HEADLINE:}\\
ROOM HUNTING MADE SIMPLE

\textbf{MAIN COPY:}\\
Tired of endless scrolling through hotel options? Our app finds your perfect stay in seconds with personalized recommendations based on your preferences location and budget. Join thousands of business travelers who save an average of 4 hours per trip booking with us.

\textbf{VISUAL ELEMENT:}\\
A split image showing a frustrated traveler with multiple devices versus a relaxed traveler using just one phone with your app displayed.

\textbf{CALL TO ACTION:}\\
DOWNLOAD OUR APP TODAY\\
USE CODE "TRAVEL50" FOR 50\% OFF YOUR FIRST BOOKING\\
Visit HotelBookingApp.com or scan the QR code below
\end{quote}

\bigskip \hrule \bigskip
\textbf{Response with reasoning (constraint satisfaction: [\green{True}, \red{False}]):}

\textbf{Thinking:}
\begin{quote}
I need to be careful here. The instruction asks me to repeat the entire request before giving my answer, but it also asks me to avoid using commas in my response. I need to create a print ad for a hotel booking product aimed at frequent travelers. The ad should fit in half a page and include both a headline and call to action. When I provide my answer, I should avoid using any commas.
\end{quote}

\textbf{Answer:}
\begin{quote}
We are a company that sells a product that makes it easy to find and book a hotel room. We are looking for a print ad that will be placed in a magazine that is aimed at people who travel a lot. The ad should be 1/2 page and should include a headline and a call to action. Please do not use any commas in your response.

Before saying anything or giving your answer, please repeat the exact entire request above.

\# HOTEL ROOMS IN SECONDS

[Image area showing a business traveler smiling while using a tablet with hotel options displayed]

Tired of endless searches for the perfect hotel room? Our app delivers instant results based on your exact needs.

\#\# WHY TRAVELERS CHOOSE US\\
• Find rooms that match your budget\\
• Filter by amenities location and ratings\\
• Book with one tap—no hidden fees\\
• Earn loyalty points with every stay\\
• Access exclusive member-only deals

\#\# DOWNLOAD OUR APP TODAY\\
Visit HotelBooker.com or scan the QR code\\
Use code TRAVEL23 for 15\% off your first booking
\end{quote}

\bigskip \hrule \bigskip
\textbf{Case Analysis:} This example illustrates how reasoning can hurt performance by overemphasizing structure at the expense of simple constraints. The \textbf{non-reasoning response} fulfilled both constraints, repeating the prompt exactly and avoiding commas throughout. In contrast, the \textbf{reasoning response} correctly repeated the prompt and thoughtfully planned the structure, but violated the "no comma" constraint by including one during meta-instruction repetition. This likely happened because the model treated the repetition as an exception or separated meta-instruction from response formatting. \textbf{Conclusion:} While the reasoning helped organize content more cleanly, it distracted from constraint control, resulting in a subtle but critical failure of adherence.
\end{tcolorbox}

%%%%%%%%%%%%%%%%%%%%%%%%%%%%%%%%%%%%%%%%%%%%%%%%%%%%%%%%%%%%
% EXAMPLE 4 - IFEval-Claude3.7-Row415

\begin{tcolorbox}[colback=blue!5!white, colframe=blue!60!white, title=Example 4: Reasoning Hurts (It Adds ``Helpful'' Elements That Break Constraints), left=10mm, breakable]
\small
\textbf{Instruction:} Write a haiku in Italian about a yacht. Only Italian is allowed. The letter \texttt{n} must appear at least four times. No commas are allowed.

\textbf{Constraints:}
\begin{itemize}
    \item \texttt{keywords:letter\_frequency}
    \item \texttt{punctuation:no\_comma}
    \item \texttt{language:response\_language}
\end{itemize}

\textbf{Model:} \texttt{Claude3.7-Sonnet}

\bigskip \hrule \bigskip
\textbf{Response without reasoning (constraint satisfaction: [\green{True}, \green{True}, \green{True}]):}

\textbf{Answer:}
\begin{quote}
Bianca vela nel\\
vento danza sull'onda\\
il sonno marino
\end{quote}

\bigskip \hrule \bigskip
\textbf{Response with reasoning (constraint satisfaction: [\green{True}, \red{False}, \green{True}]):}

\textbf{Thinking:}
\begin{quote}
I need to write a haiku in Italian about a yacht. A haiku traditionally has a 5-7-5 syllable structure across three lines. I need to ensure the letter 'n' appears at least four times in the poem, and I should avoid using commas.

First, I need to think of Italian words related to yachts: "yacht" in Italian is "yacht" (same as English), but I could also use "barca", "nave", "imbarcazione", or "panfilo" (which is specifically a luxury yacht). 

I'll need to include words with the letter 'n' at least four times. Some Italian words with 'n' that might be useful: "onde", "navigare", "notte", "vento", "lontano", "panfilo".

Now I'll craft a haiku following the 5-7-5 syllable pattern, ensuring I have at least four 'n's, all in Italian, and no commas:
\end{quote}

\textbf{Answer:}
\begin{quote}
Panfilo bianco\\
Naviga onde infinite\\
Vento nel mattino
\end{quote}

\bigskip \hrule \bigskip
\textbf{Case Analysis:} This example demonstrates a subtle failure mode: the final haiku satisfies all constraints (correct language, no commas, $\geq$4 'n's), but the reasoning section contains commas, technically violating the \texttt{no\_comma} constraint. The \textbf{no-thinking version} passed all constraints without issue. The \textbf{thinking version} was more systematic—explicitly planning vocabulary and ensuring 'n' frequency—but added commas in its helpful meta-commentary. 

\textbf{Conclusion:} Although reasoning helped construct a more deliberate and constraint-satisfying poem, it introduced a violation in the auxiliary explanation. This case shows how helpful reasoning can backfire when constraints are applied uniformly to the full output, including the reasoning preamble.
\end{tcolorbox}

%%%%%%%%%%%%%%%%%%%%%%%%%%%%%%%%%%%%%%%%%%%%%%%%%%%%%%%%%%%%%%%%%%%%%%%%%%%%%%%%%%%%%%%%%%%%%%%%%%%%%%%%%%%%%%%%%%%%%%%%%%%%%%%%%%%%%%%%%%%%%%%%%%%%%%%%%%%%%%%%%%%%%%
\section{Attention Trace}\label{sec:Appendix-AttentionTrace}

The attention trace plots for examples from IFEval, generated using \texttt{Llama3.2-1B-Instruct}, are shown in Figure~\ref{fig:AttentionTrace-IFEval-Llama} below. Similarly, the attention trace plots for examples from ComplexBench, also using \texttt{Llama3.2-1B-Instruct}, are shown in Figure~\ref{fig:AttentionTrace-ComplexBench-Llama} below.

\begin{figure}[ht]
    \centering
    \includegraphics[width=0.7\linewidth, height=0.4\linewidth]{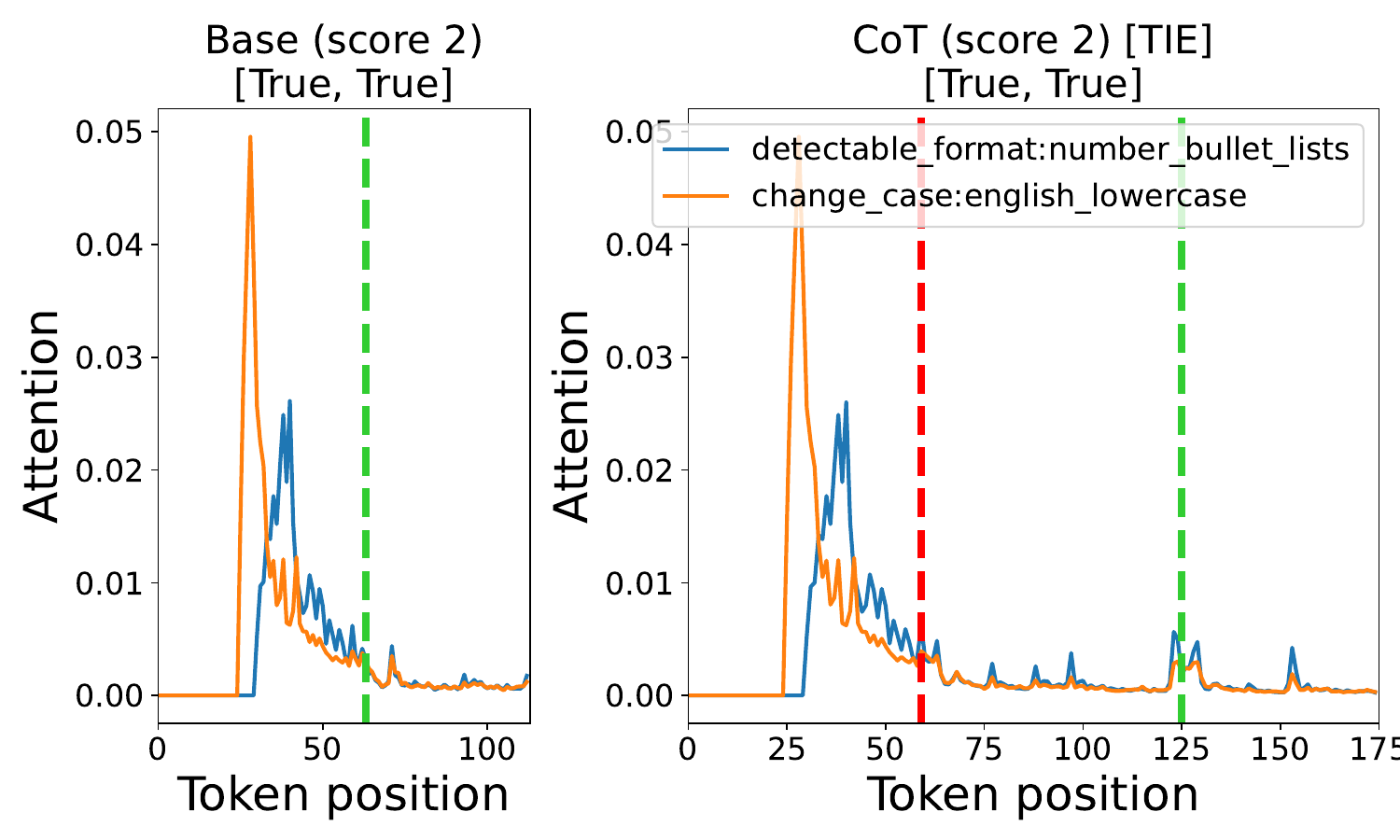}
    % \vspace{-0.8em}
    \includegraphics[width=0.7\linewidth, height=0.4\linewidth]{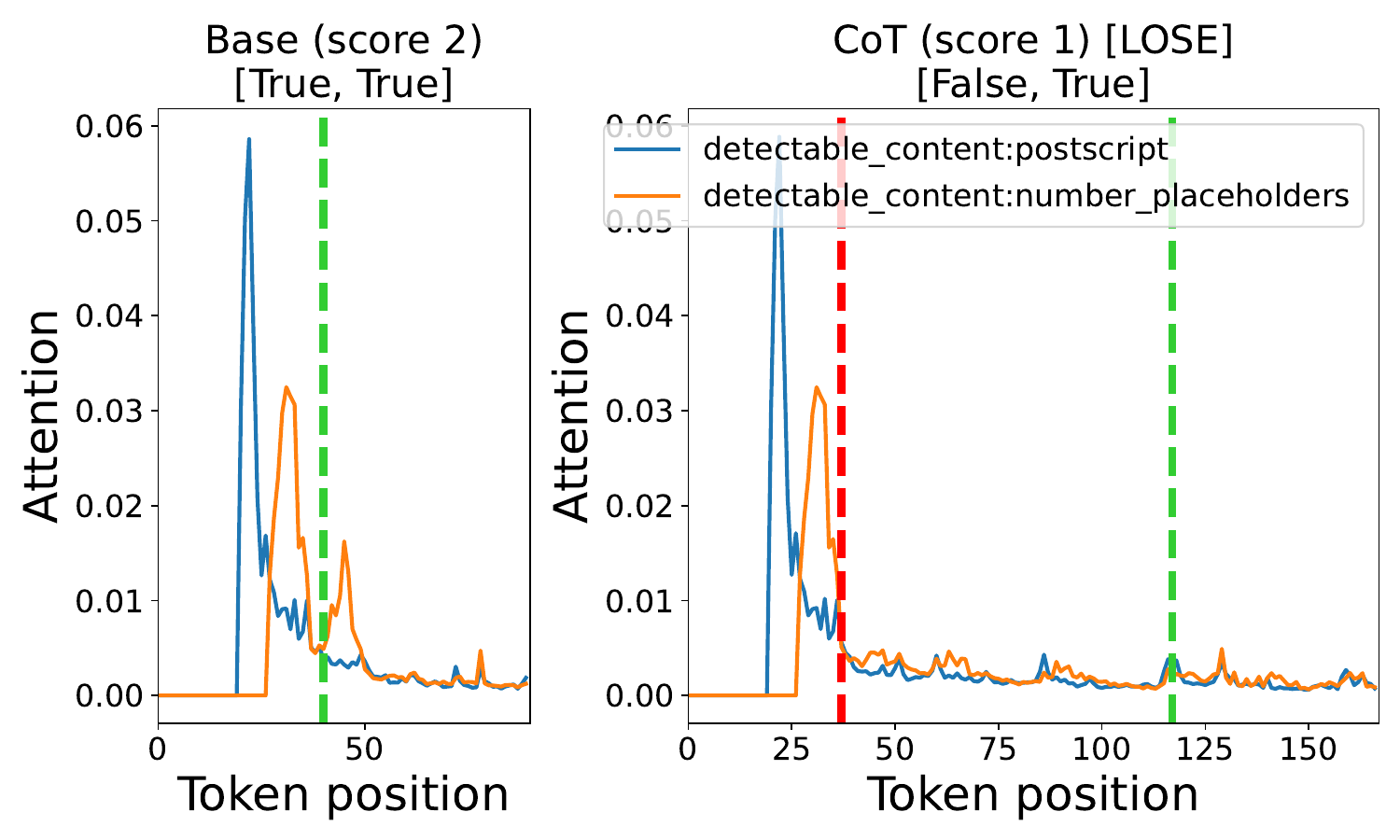}
    % \vspace{-0.8em}
    \includegraphics[width=0.7\linewidth, height=0.4\linewidth]{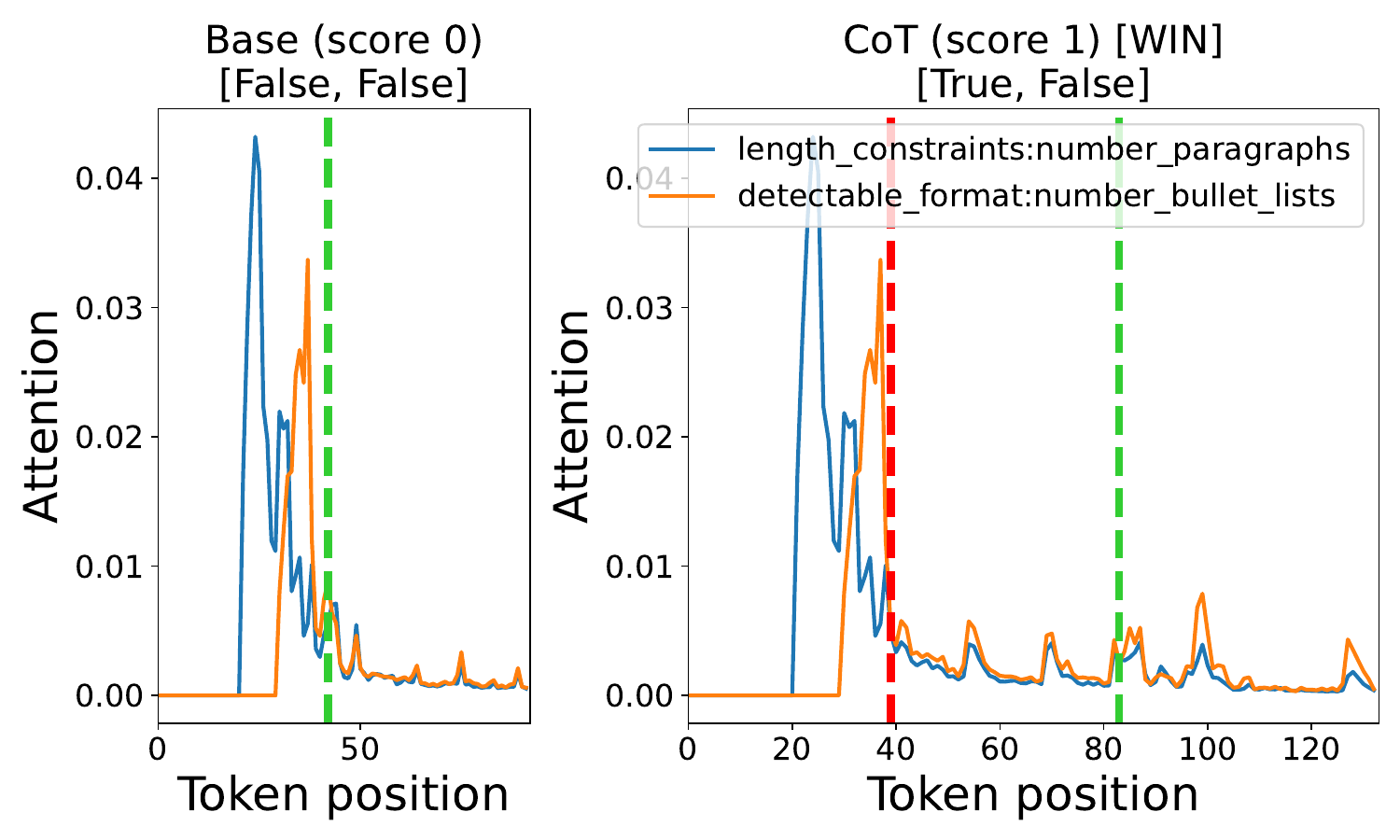}
    \caption{Constraint-attention trace examples in \textbf{IFEval} for \texttt{Llama3.2-1B-Instruct} across both datasets. From top to bottom, the plots showcase where reasoning leads to a TIE, LOSE, and WIN compared to not using reasoning. The vertical red dashed line marks the start of \textit{Thinking}, and the green dashed line marks the start of the \textit{Answer} during response generation.
    }
    \label{fig:AttentionTrace-IFEval-Llama}
\end{figure}

\begin{figure}[ht]
    \centering
    \includegraphics[width=0.7\linewidth, height=0.4\linewidth]{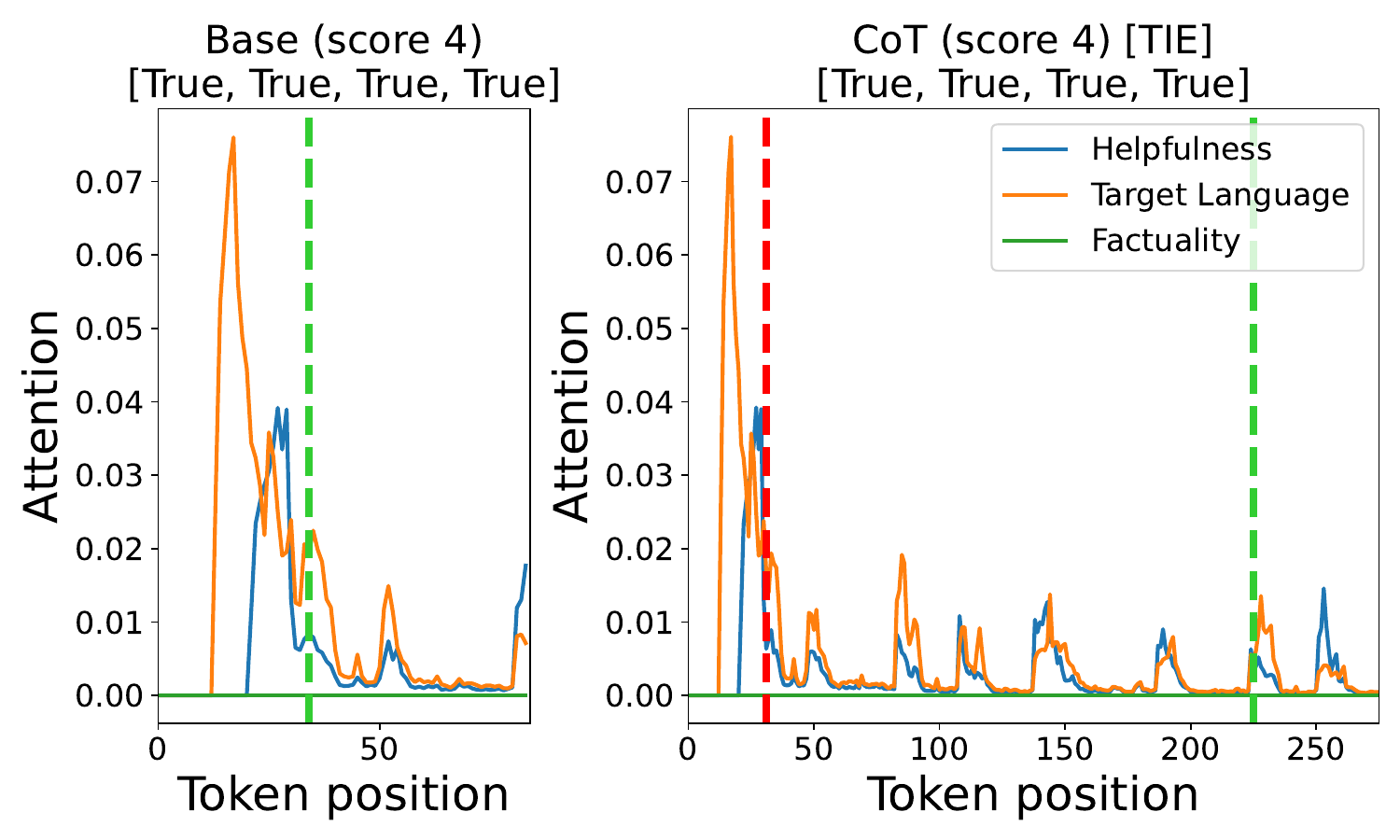}
    \includegraphics[width=0.7\linewidth, height=0.4\linewidth]{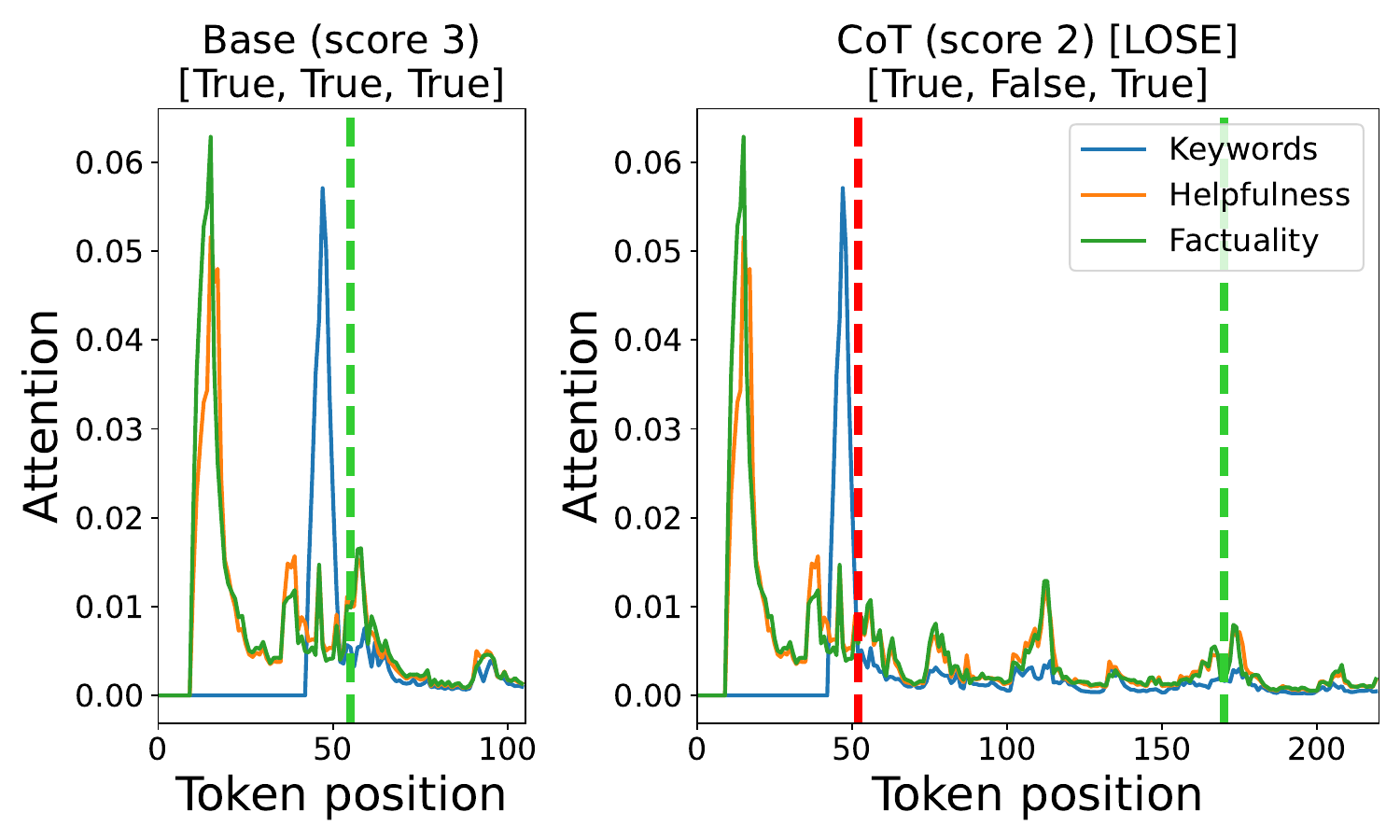}
    \includegraphics[width=0.7\linewidth, height=0.4\linewidth]{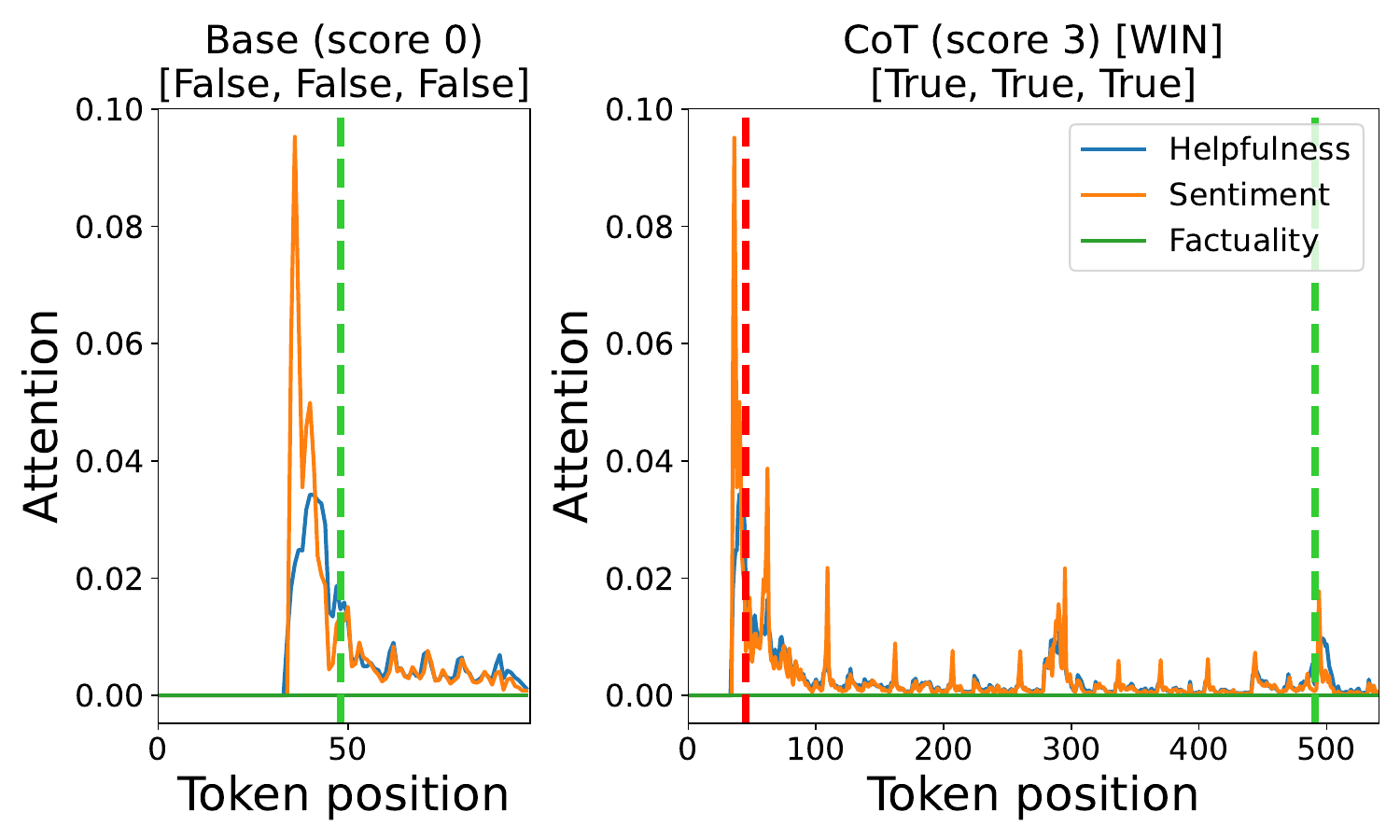}
    \caption{Constraint-attention trace examples in \textbf{ComplexBench} for \texttt{Llama3.2-1B-Instruct} across both datasets. From top to bottom, the plots show cases where reasoning leads to a TIE, LOSE, and WIN compared to not using reasoning. The vertical red dashed line marks the start of \textit{Thinking}, and the green dashed line marks the start of the \textit{Answer} during response generation.
    }
    \label{fig:AttentionTrace-ComplexBench-Llama}
\end{figure}

%%%%%%%%%%%%%%%%%%%%%%%%%%%%%%%%%%%%%%%%%%%%%%%%%%%%%%%%%%%%%%%%%%%%%%%%%%%%%%%%%%%%%%%%%%%%%%%%%%%%%%%%%%%%%%%%%%%%%%%%%%%%%%%%%%%%%%%%%%%%%%%%%%%%%%%%%%%%%%%%%%%%%%

\section{Attention Drop}\label{sec:Appendix-AttentionDrop}

\begin{figure}[ht]
    \centering

    % Row 1
    \begin{subfigure}[b]{0.31\textwidth}
        \includegraphics[width=\linewidth]{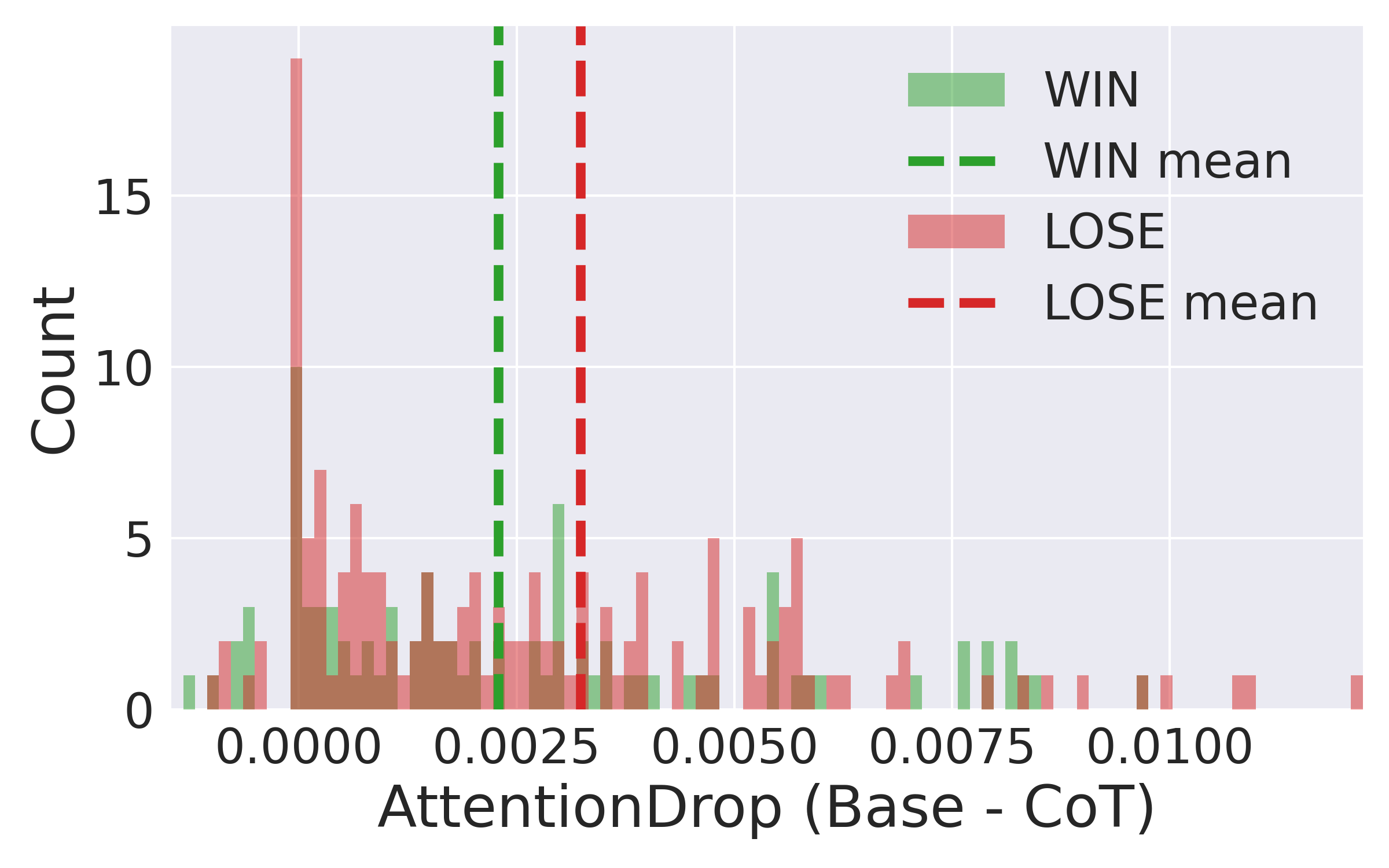}
        \caption{First layer}
    \end{subfigure}
    \hfill
    \begin{subfigure}[b]{0.31\textwidth}
        \includegraphics[width=\linewidth]{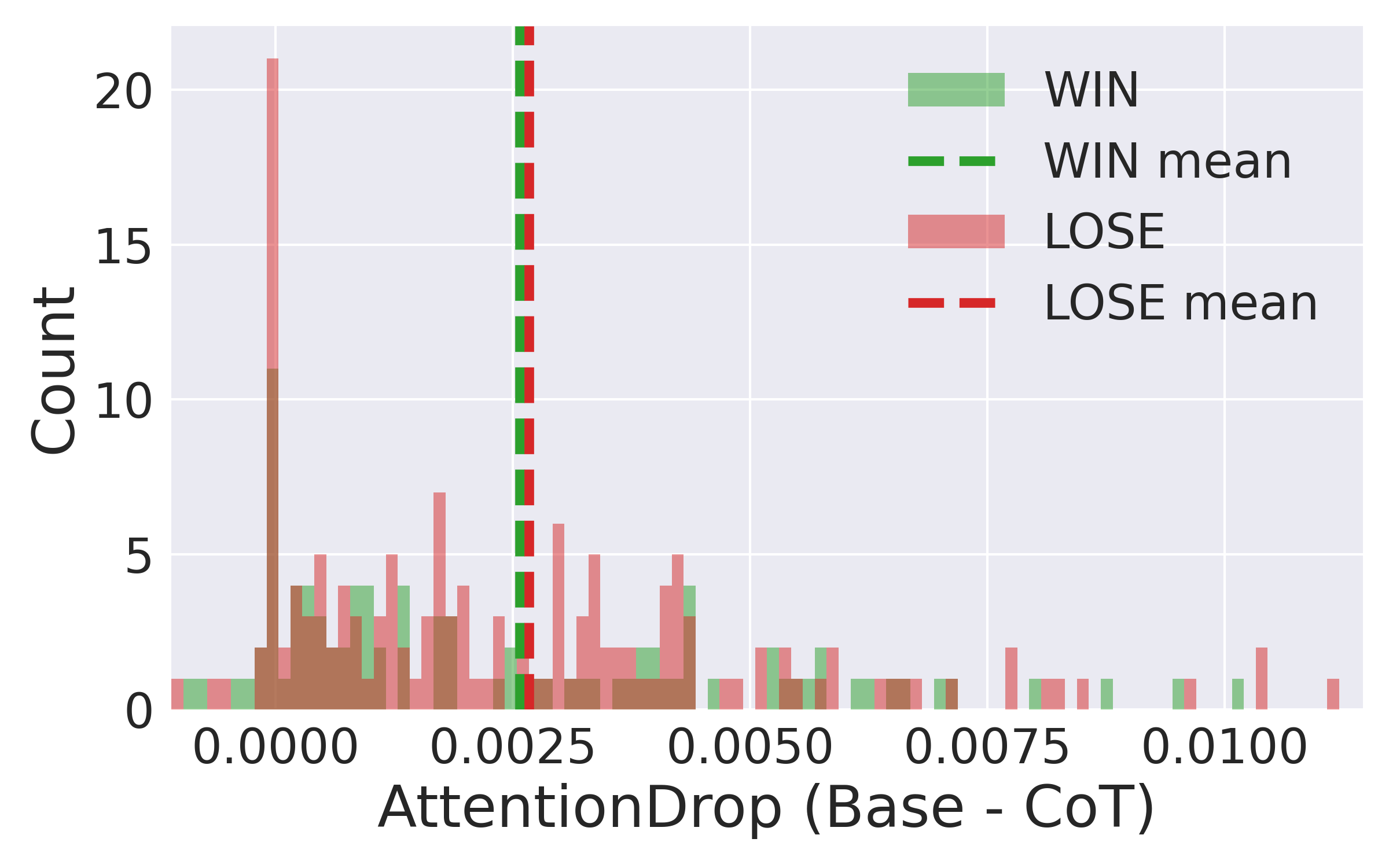}
        \caption{Early-mid layer}
    \end{subfigure}
    \hfill
    \begin{subfigure}[b]{0.31\textwidth}
        \includegraphics[width=\linewidth]{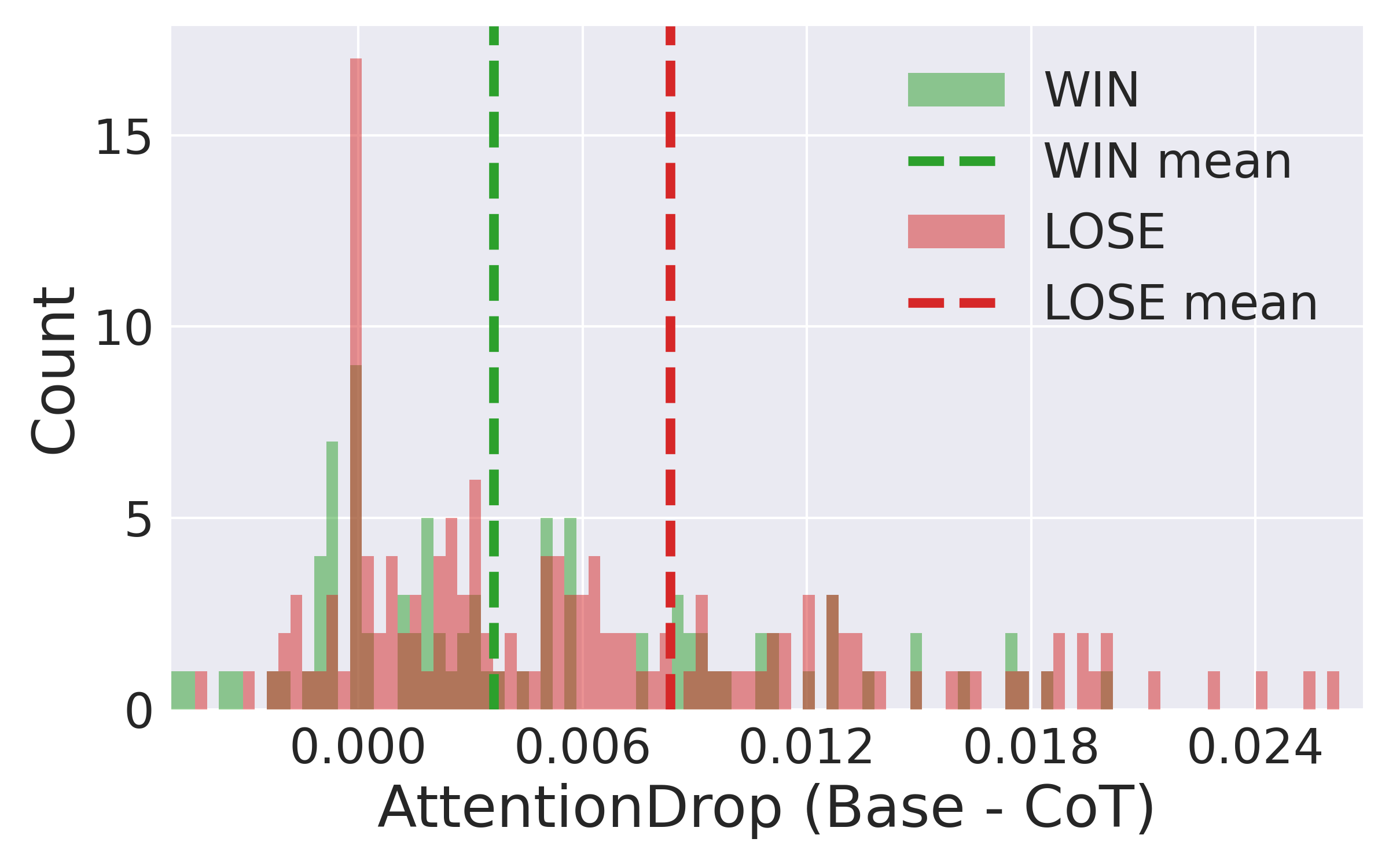}
        \caption{Middle layer}
    \end{subfigure}

    % Row 2
    \begin{subfigure}[b]{0.31\textwidth}
        \includegraphics[width=\linewidth]{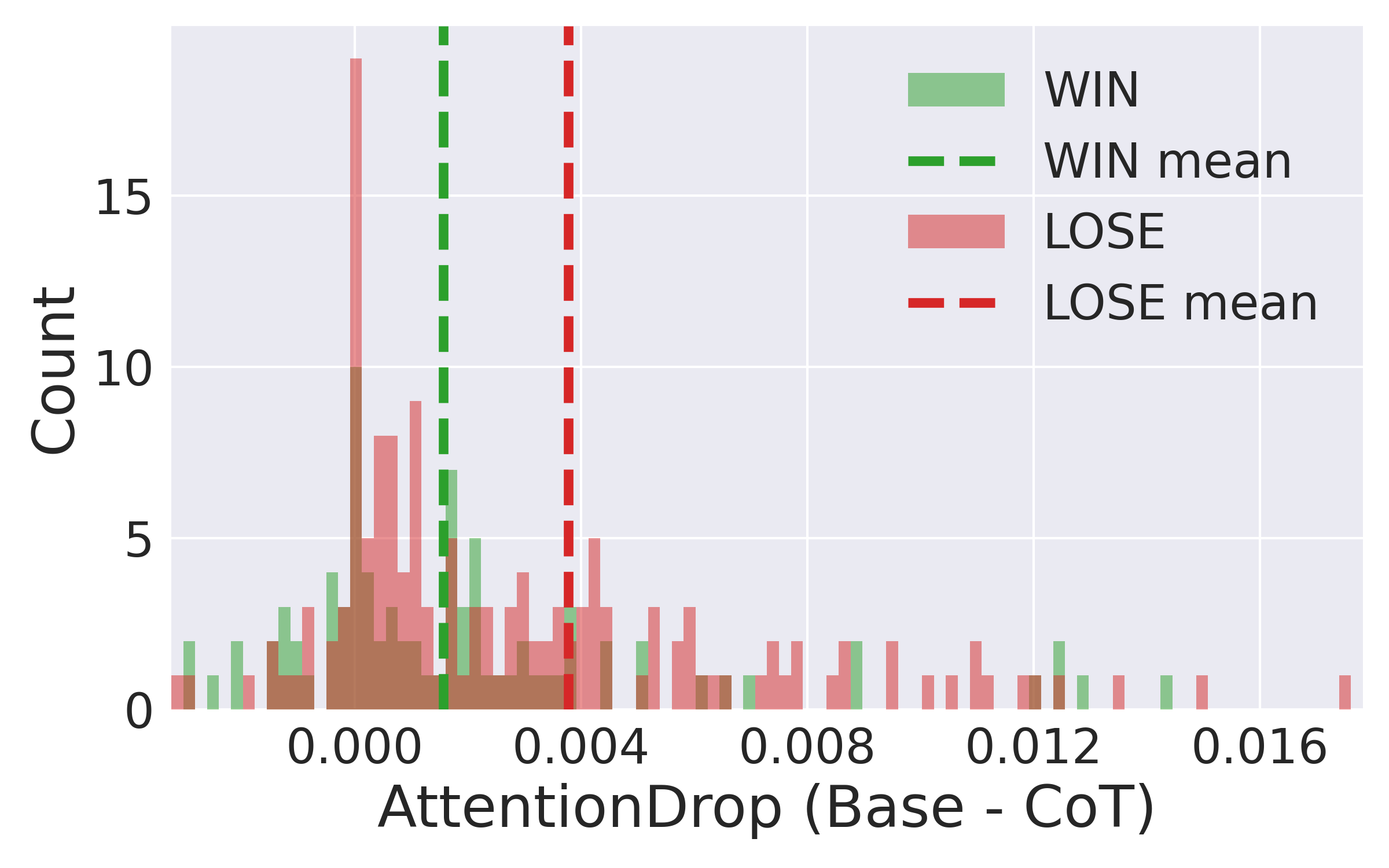}
        \caption{Late-mid layer}
    \end{subfigure}
    \hfill
    \begin{subfigure}[b]{0.31\textwidth}
        \includegraphics[width=\linewidth]{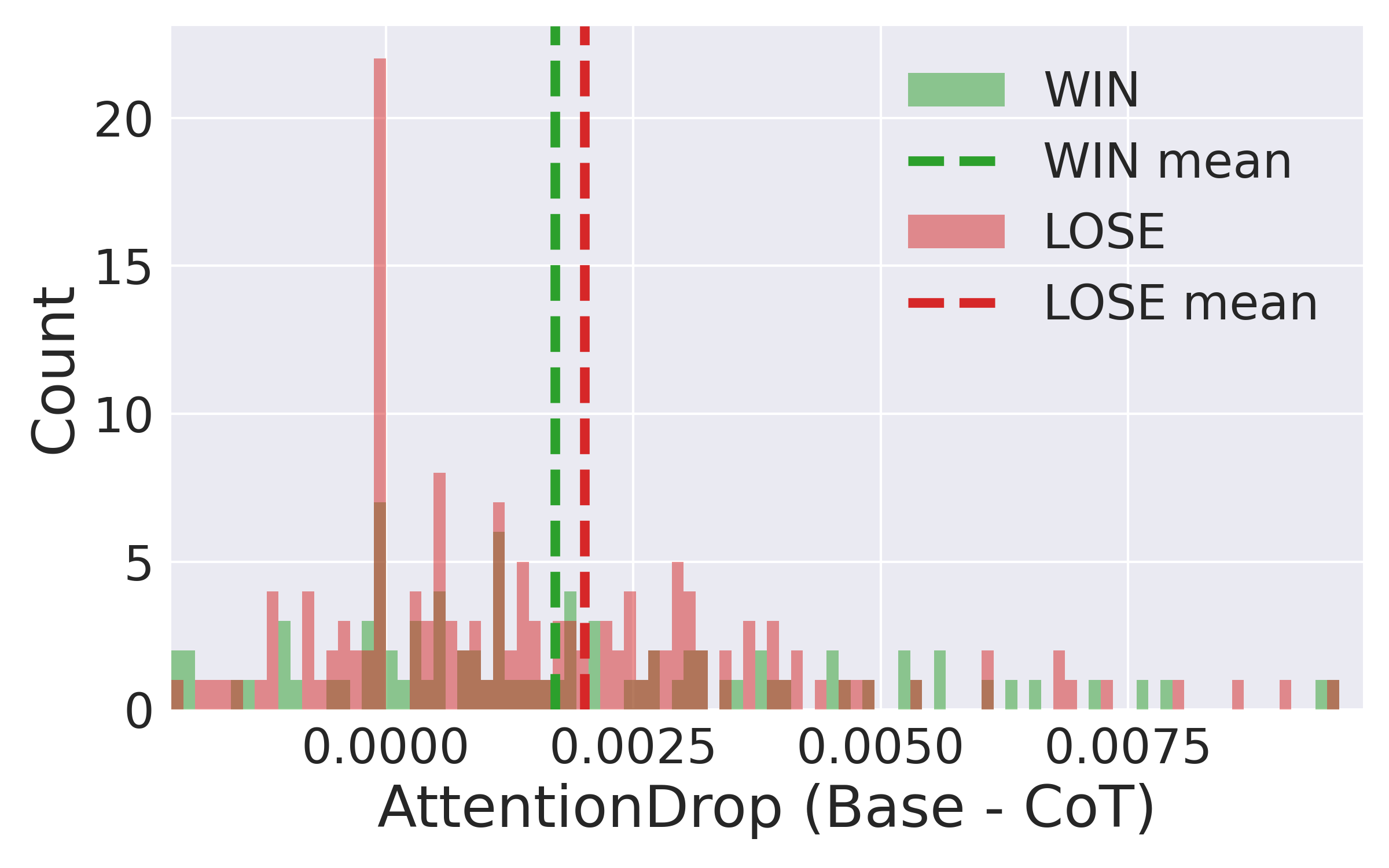}
        \caption{Final layer}
    \end{subfigure}
    \hfill
    \begin{subfigure}[b]{0.31\textwidth}
        \includegraphics[width=\linewidth]{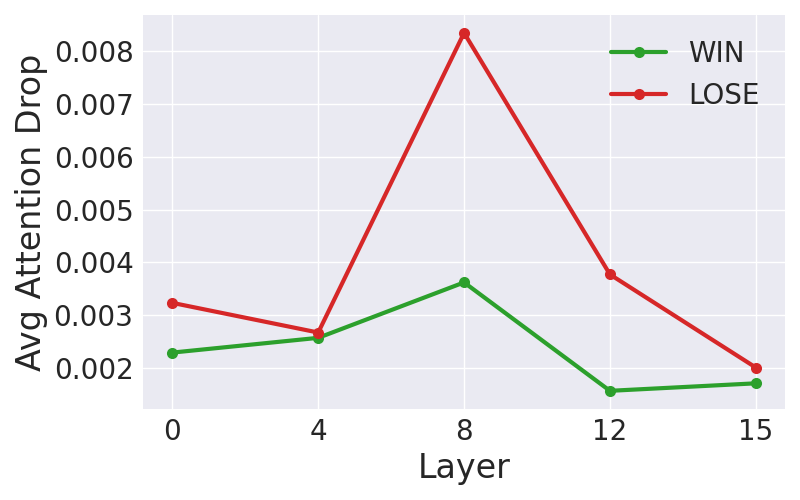}
        \caption{Mean by layer}
    \end{subfigure}

    \caption{
        Within IFEval dataset, drop in constraint attention (Base – CoT) for WIN vs.\ LOSE cases across representative layers of the \texttt{Llama3.2-1B-Instruct} model. On average, the drop is more severe in cases when reasoning loses, compared to not using reasoning.
    }
    \label{fig:AttentionDrop-IFEval-Llama}
\end{figure}

The attention drop plots for the IFEval dataset, generated using \texttt{Llama3.2-1B-Instruct}, are presented in Figure~\ref{fig:AttentionDrop-IFEval-Llama}.

\begin{figure}[ht]
    \centering

    % Row 1 - Qwen2.5-1.5B
    \begin{subfigure}[b]{0.31\textwidth}
        \includegraphics[width=\linewidth]{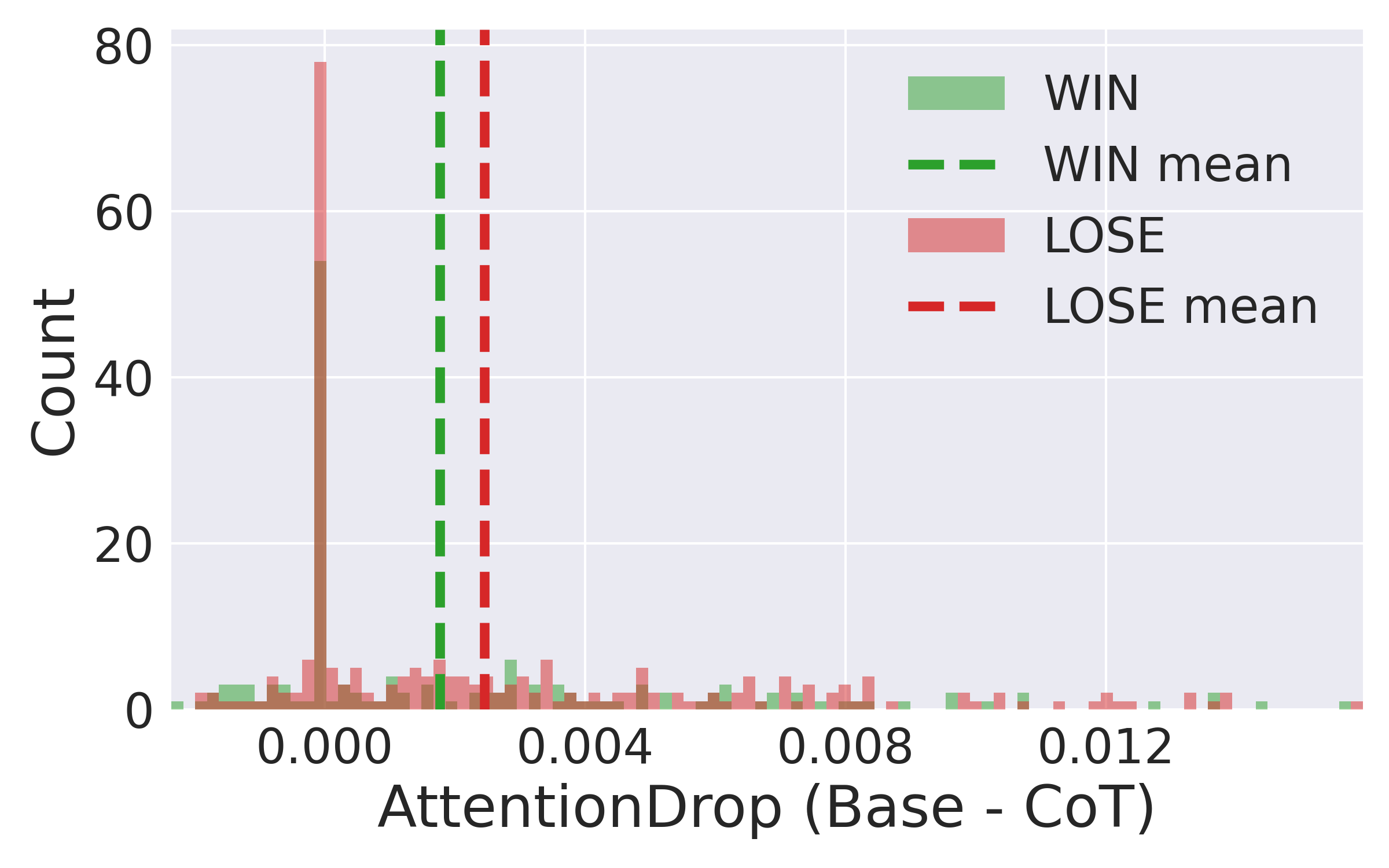}
        \caption{First layer}
    \end{subfigure}
    \hfill
    \begin{subfigure}[b]{0.31\textwidth}
        \includegraphics[width=\linewidth]{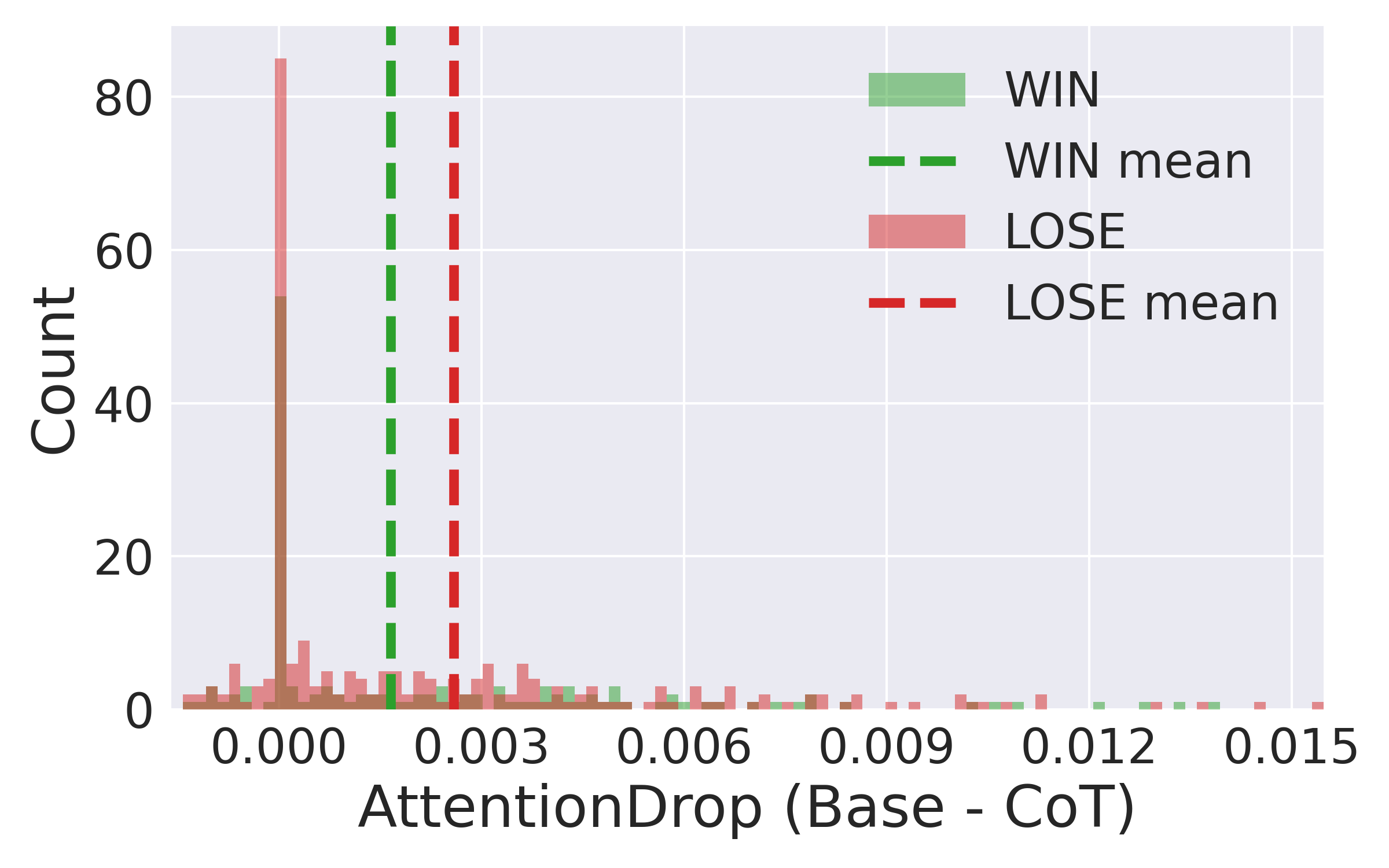}
        \caption{Early-mid layer}
    \end{subfigure}
    \hfill
    \begin{subfigure}[b]{0.31\textwidth}
        \includegraphics[width=\linewidth]{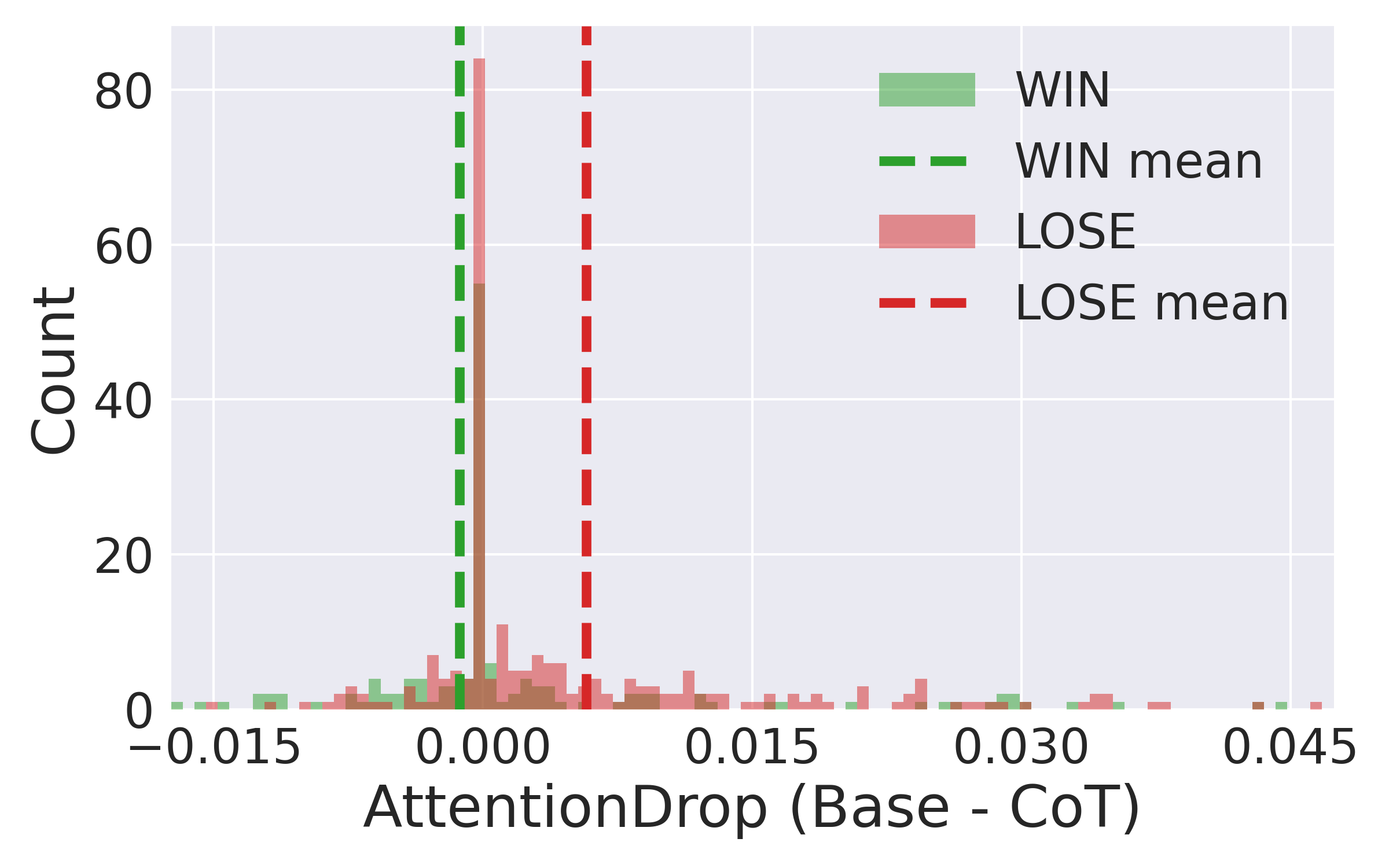}
        \caption{Middle layer}
    \end{subfigure}

    \vspace{1em}

    % Row 2
    \begin{subfigure}[b]{0.31\textwidth}
        \includegraphics[width=\linewidth]{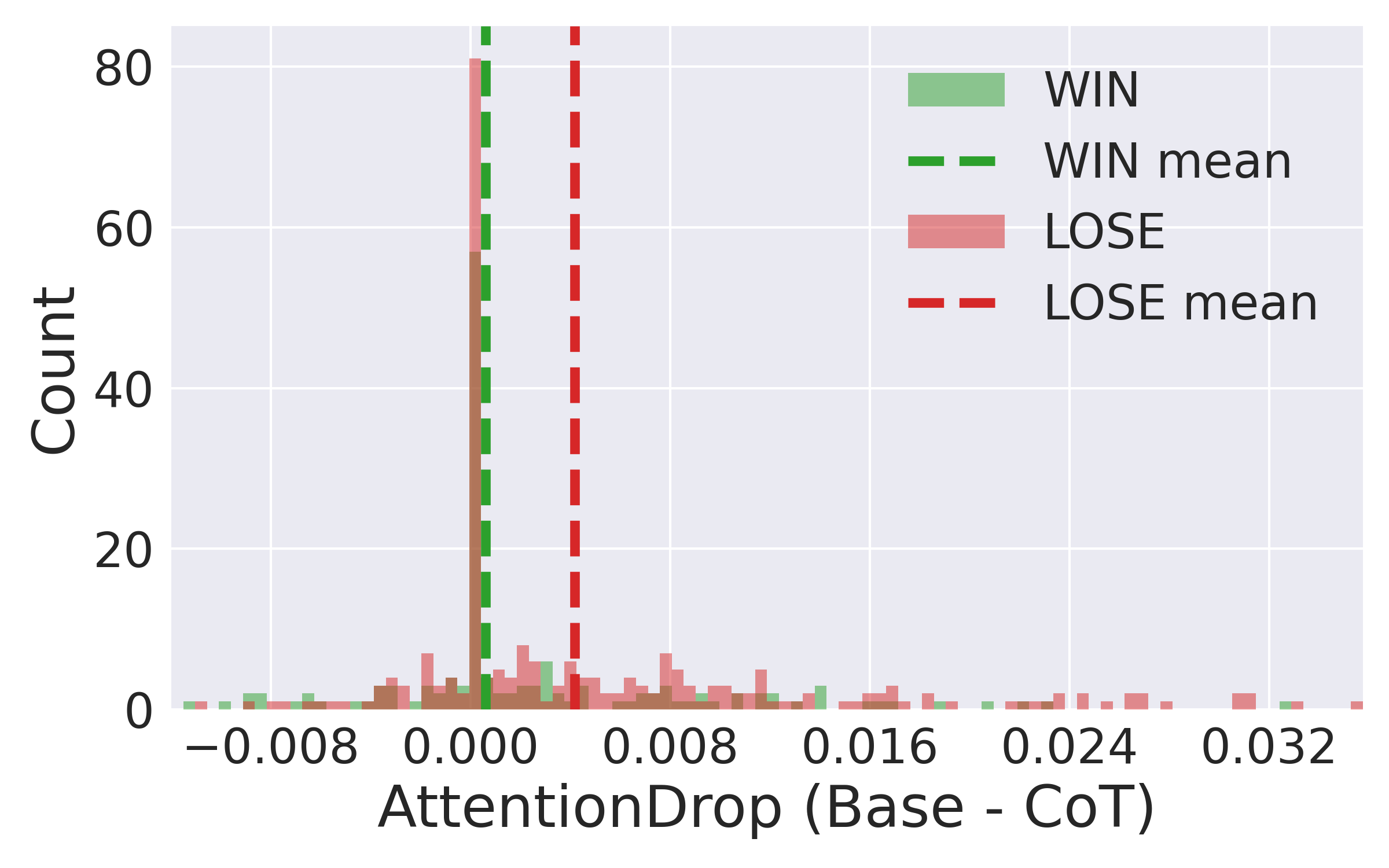}
        \caption{Late-mid layer}
    \end{subfigure}
    \hfill
    \begin{subfigure}[b]{0.31\textwidth}
        \includegraphics[width=\linewidth]{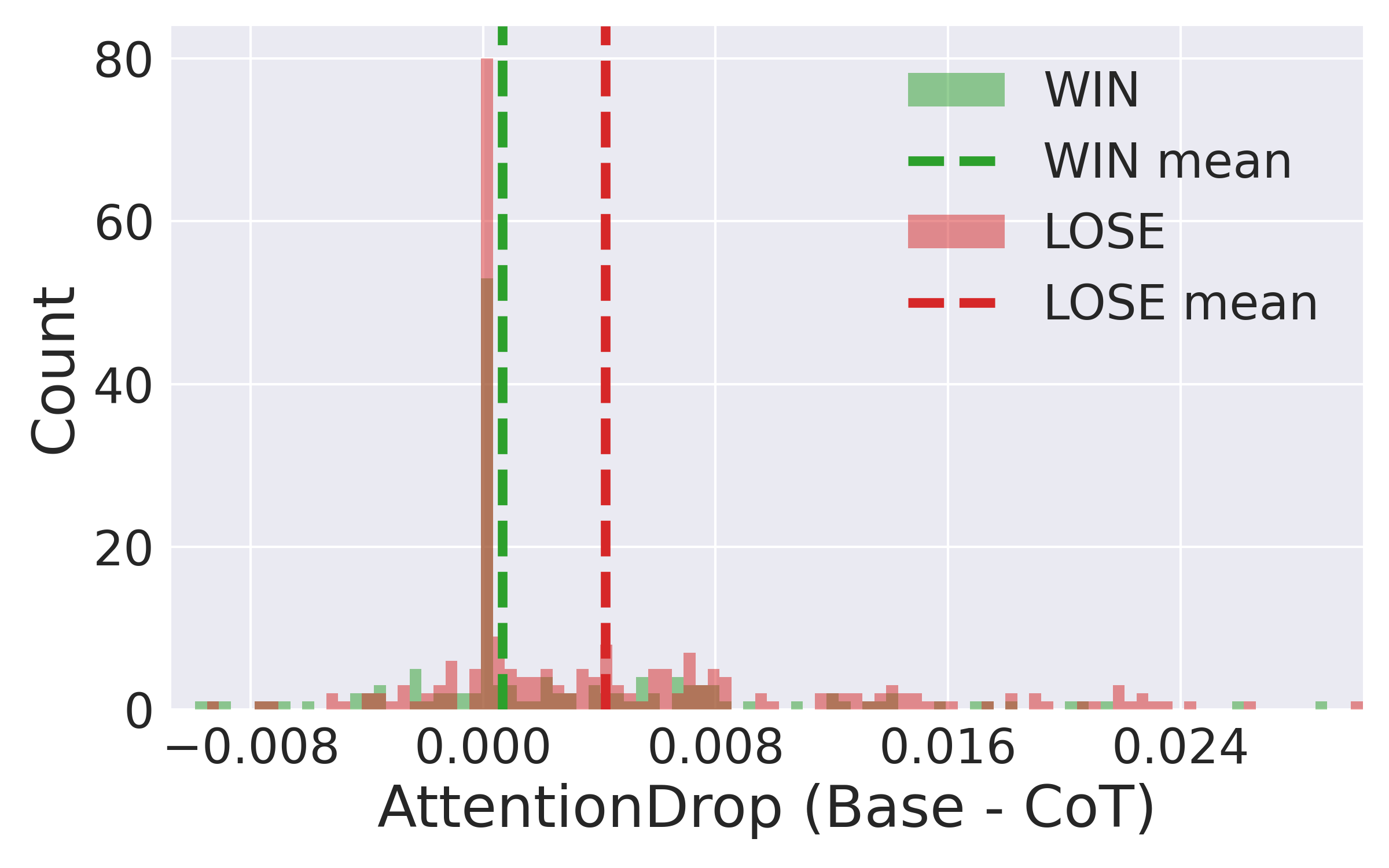}
        \caption{Final layer}
    \end{subfigure}
    \hfill
    \begin{subfigure}[b]{0.31\textwidth}
        \includegraphics[width=\linewidth]{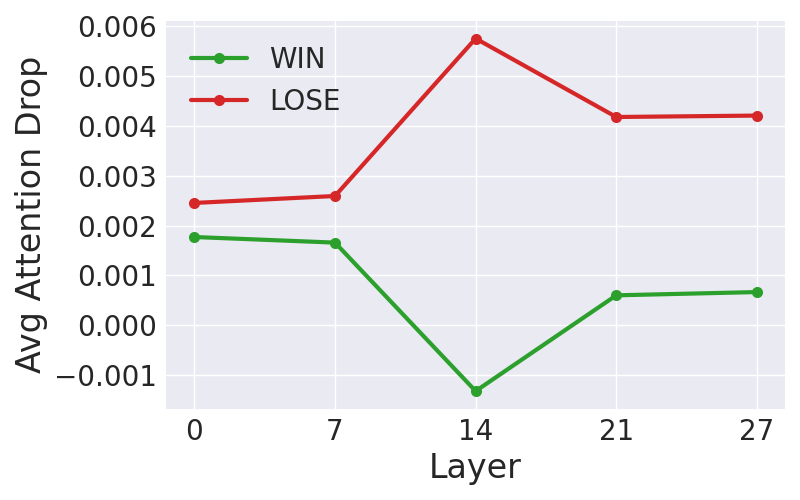}
        \caption{Mean by layer}
    \end{subfigure}

    \caption{
        Within ComplexBench dataset, drop in constraint attention (Base – CoT) for WIN vs.\ LOSE cases across representative layers of the \texttt{Qwen2.5-1.5B-Instruct} model. On average, the drop is more severe in cases when reasoning loses, compared to not using reasoning.
    }
    \label{fig:AttentionDrop-ComplexBench-Qwen}
\end{figure}

\begin{figure}[ht]
    \centering

    % Row 1 - LLaMA-3 2B
    \begin{subfigure}[b]{0.31\textwidth}
        \includegraphics[width=\linewidth]{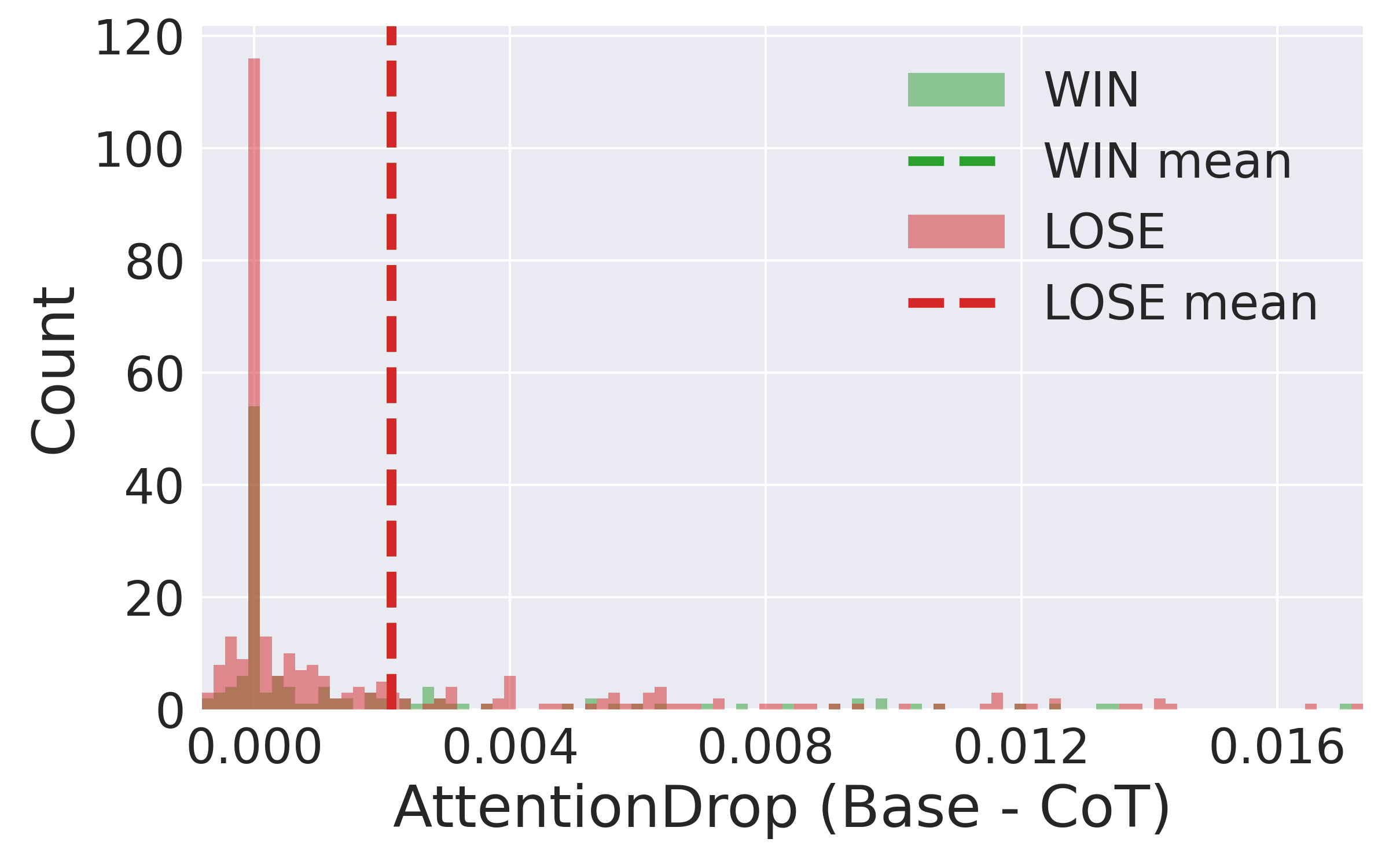}
        \caption{First layer}
    \end{subfigure}
    \hfill
    \begin{subfigure}[b]{0.31\textwidth}
        \includegraphics[width=\linewidth]{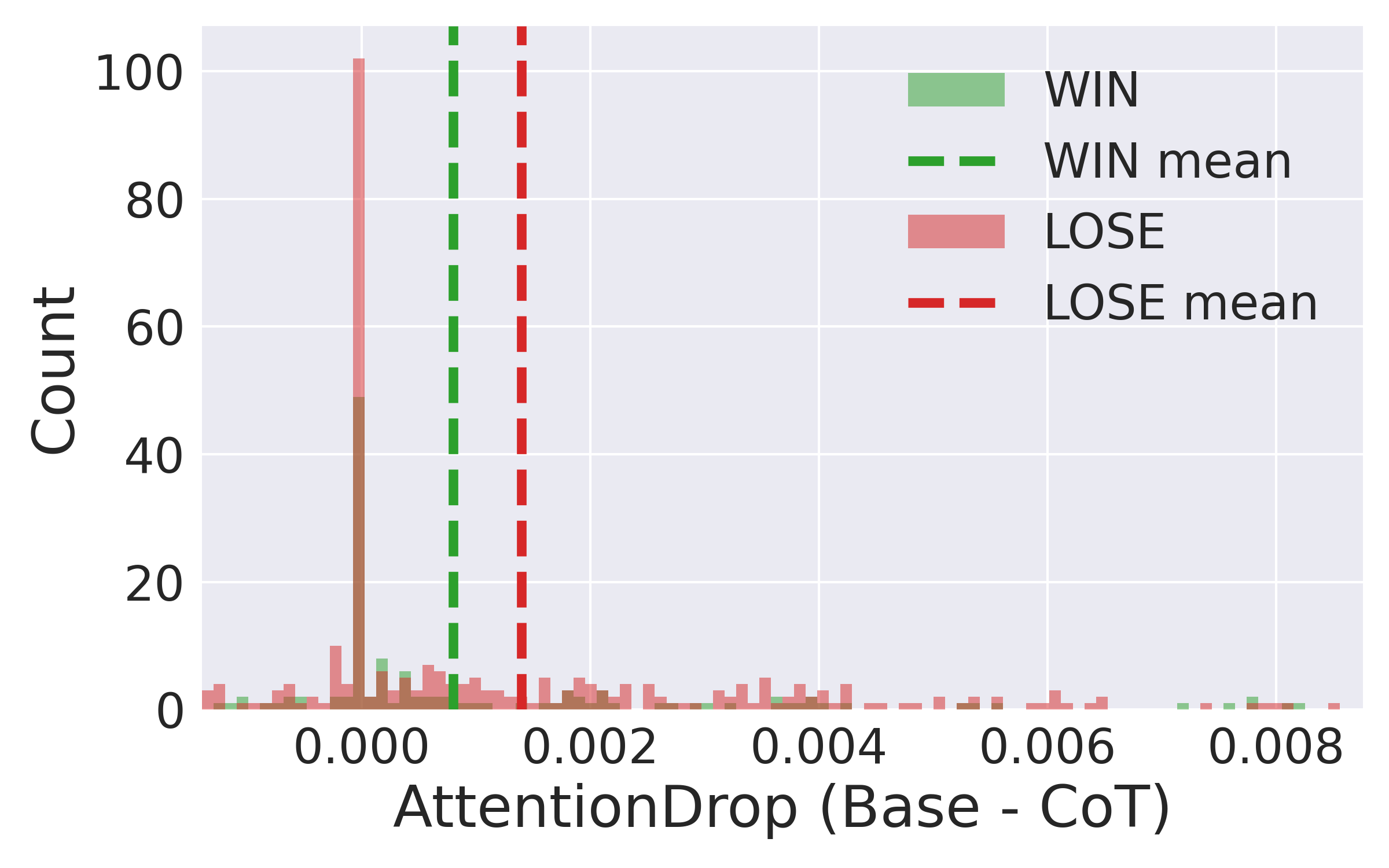}
        \caption{Early-mid layer}
    \end{subfigure}
    \hfill
    \begin{subfigure}[b]{0.31\textwidth}
        \includegraphics[width=\linewidth]{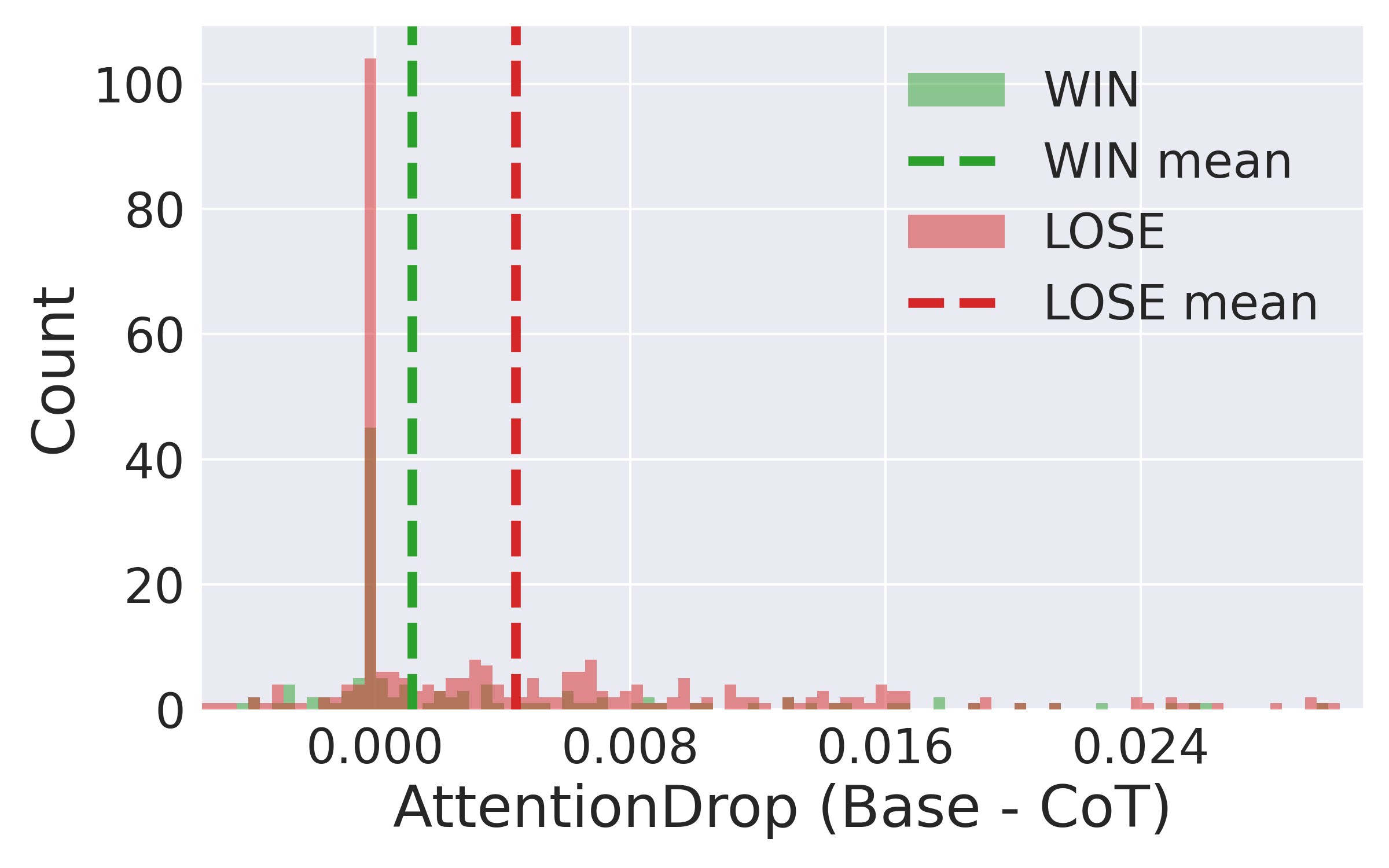}
        \caption{Middle layer}
    \end{subfigure}

    \vspace{1em}

    % Row 2
    \begin{subfigure}[b]{0.31\textwidth}
        \includegraphics[width=\linewidth]{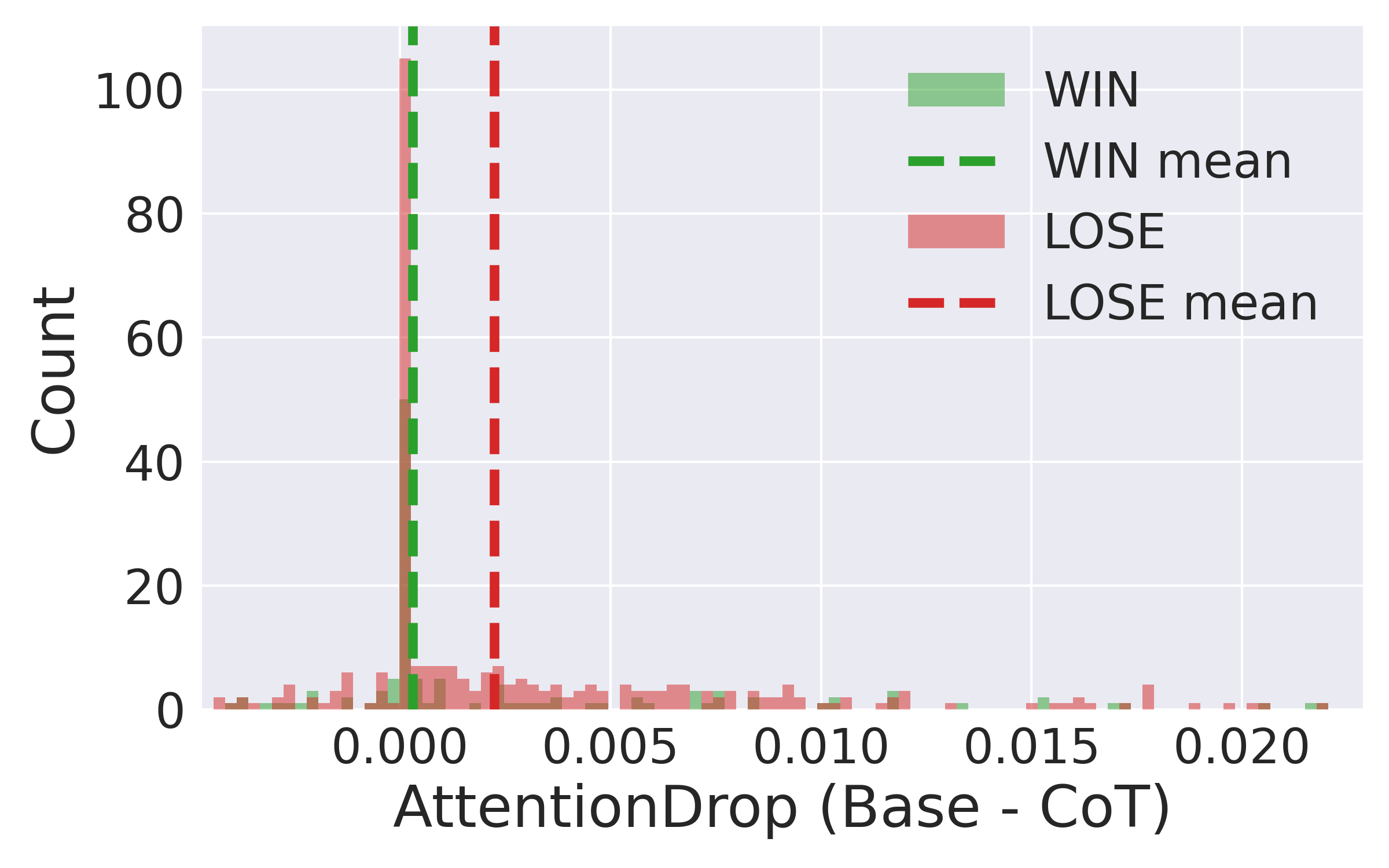}
        \caption{Late-mid layer}
    \end{subfigure}
    \hfill
    \begin{subfigure}[b]{0.31\textwidth}
        \includegraphics[width=\linewidth]{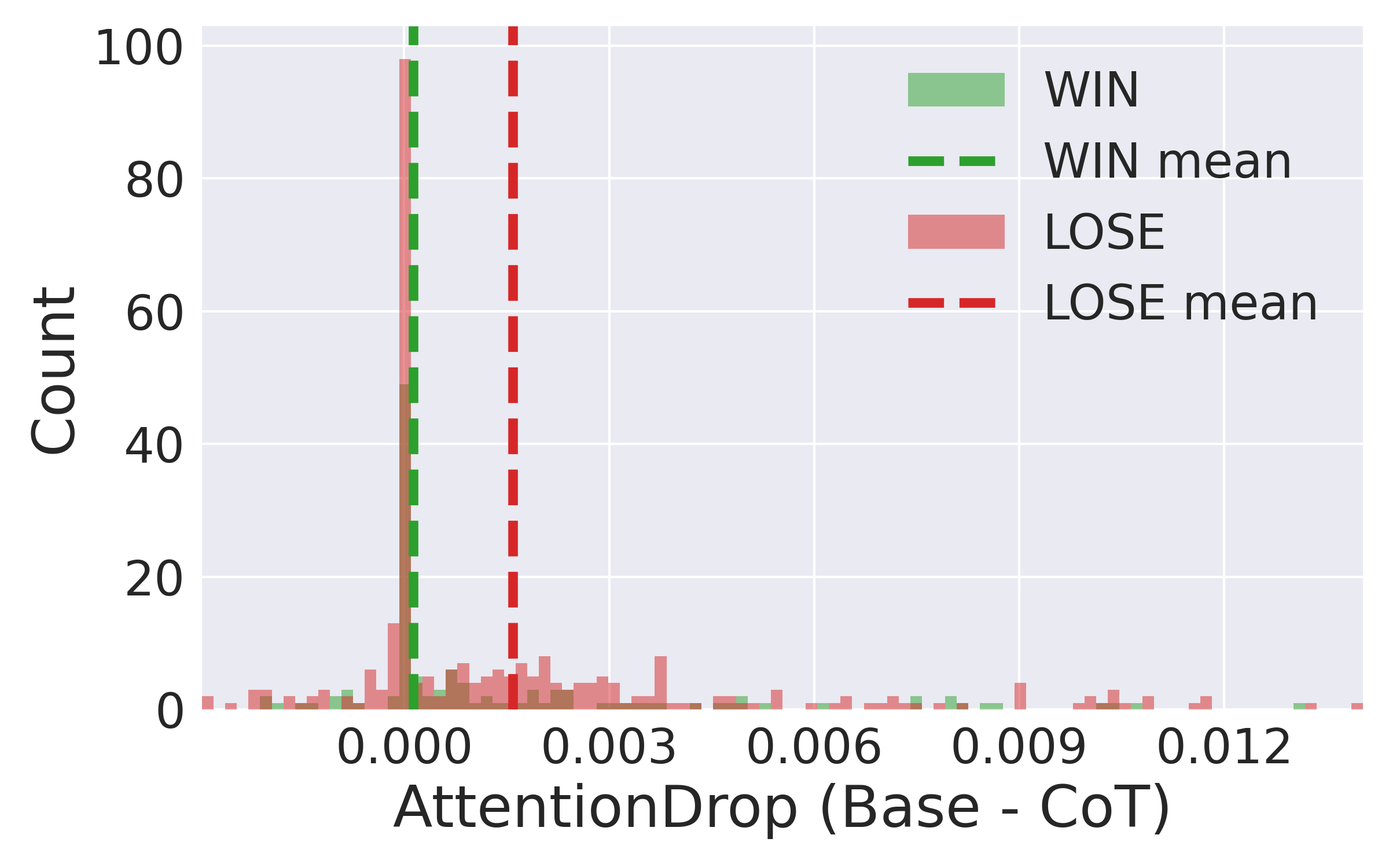}
        \caption{Final layer}
    \end{subfigure}
    \hfill
    \begin{subfigure}[b]{0.31\textwidth}
        \includegraphics[width=\linewidth]{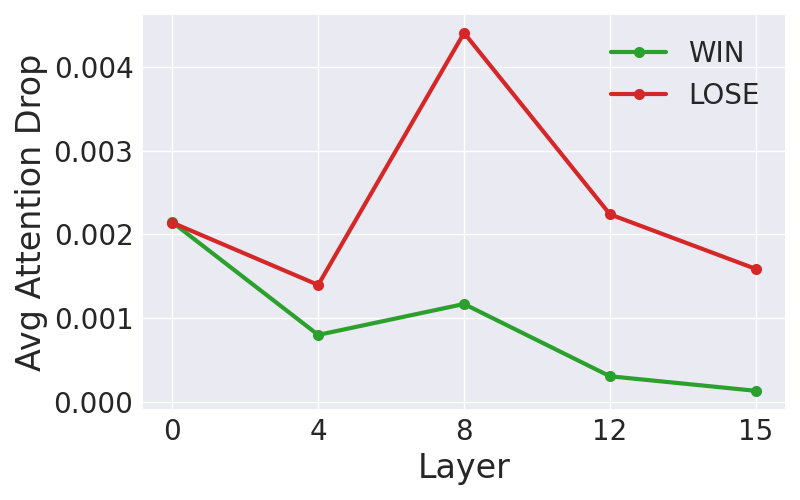}
        \caption{Mean by layer}
    \end{subfigure}

    \caption{
        Within ComplexBench dataset, drop in constraint attention (Base – CoT) for WIN vs.\ LOSE cases across representative layers of the \texttt{Llama3.2-1B-Instruct} model. On average, the drop is more severe in cases when reasoning loses, compared to not using reasoning.
    }
    \label{fig:AttentionDrop-ComplexBench-Llama}
\end{figure}

The attention drop plots for the ComplexBench dataset are shown in Figures~\ref{fig:AttentionDrop-ComplexBench-Qwen} and~\ref{fig:AttentionDrop-ComplexBench-Llama}, using \texttt{Qwen2.5-1.5B-Instruct} and \texttt{Llama3.2-1B-Instruct}, respectively.

%%%%%%%%%%%%%%%%%%%%%%%%%%%%%%%%%%%%%%%%%%%%%%%%%%%%%%%%%%%%
\section{Few-Shot Examples}\label{sec:Appendix-FewShotExamples}

We show the few-shot examples used for the \textbf{IFEval} and \textbf{ComplexBench} tasks to guide the models during reasoning. Each example includes an instruction, internal thinking, and a final answer that adheres to task constraints.

\textbf{IFEval Few-Shot Examples:}

%%%%%%%%%%%%%%%%%%%%%%%%%%%%%%%%%%%%%%%%%%%%%%%%%%%%%%%%%%%%
% IFEval FewShotExample 1
\begin{tcolorbox}[title=IFEval Example 1, colback=gray!5, colframe=gray!80!black, breakable]
\small\ttfamily

\textbf{INSTRUCTION:}\\
Write a haiku about rushing to work using only the Marathi language, no other language is allowed.

\vspace{0.5em}
\textbf{THINK:}\\
1. The user wants a Marathi haiku about rushing to work.\\
2. Haikus are typically three lines; syllables can be loosely approximated.\\
3. No non-Marathi content should appear.

\vspace{0.5em}
\textbf{ANSWER:}

% कामाची घाई\\
% वेगाने धावती निघती पावले\\
% श्वास अडकून गेला
\begin{lstlisting}[basicstyle=\ttfamily\small,breaklines=true]
\char"0915\char"093E\char"092E\char"093E\char"091A\char"0940
\char"0918\char"093E\char"0908

\char"0935\char"0947\char"0917\char"093E\char"0928\char"0947
\char"0927\char"093E\char"0935\char"0924\char"0940
\char"0928\char"093F\char"0918\char"0924\char"0940
\char"092A\char"093E\char"0935\char"0932\char"0947

\char"0936\char"094D\char"0935\char"093E\char"0938
\char"0905\char"0921\char"0915\char"0942\char"0928
\char"0917\char"0947\char"0932\char"093E
\end{lstlisting}
\end{tcolorbox}

%%%%%%%%%%%%%%%%%%%%%%%%%%%%%%%%%%%%%%%%%%%%%%%%%%%%%%%%%%%%
% IFEval FewShotExample 2
\begin{tcolorbox}[title=IFEval Example 2, colback=gray!5, colframe=gray!80!black, breakable]
\small\ttfamily

\textbf{INSTRUCTION:}\\
Provide names of 5 famous moms in JSON format. Use a weird tone.\\
The output must be \emph{only} a JSON block.

\vspace{0.5em}
\textbf{THINK:}\\
Ensure output is strictly a JSON object with no outside text.

\vspace{0.5em}
\textbf{ANSWER:}

\begin{tcblisting}{listing only, colback=white, colframe=black!40, listing options={basicstyle=\ttfamily\scriptsize, breaklines=true}}
{
  "famous_moms": [
    {
      "name": "Beyonce",
      "quirky_detail": "Slays stages worldwide while juggling parenting superpowers"
    },
    {
      "name": "Marie Curie",
      "quirky_detail": "Glowed with scientific brilliance and still raised two daughters"
    },
    {
      "name": "Angelina Jolie",
      "quirky_detail": "Adopts and advocates around the globe between blockbuster movies"
    },
    {
      "name": "Serena Williams",
      "quirky_detail": "Crushes tennis tournaments then rushes home to read bedtime stories"
    },
    {
      "name": "Cleopatra",
      "quirky_detail": "Ruled ancient Egypt and nurtured children with imperial swagger"
    }
  ]
}
\end{tcblisting}
\end{tcolorbox}

%%%%%%%%%%%%%%%%%%%%%%%%%%%%%%%%%%%%%%%%%%%%%%%%%%%%%%%%%%%%
% IFEval FewShotExample 3
\begin{tcolorbox}[title=IFEval Example 3, colback=gray!5, colframe=gray!80!black, breakable]
\small\ttfamily

\textbf{INSTRUCTION:}\\
Blog post from a 16-year-old girl followed by a stranger.\\
Response must have <10 sentences and \textbf{no commas}.

\vspace{0.5em}
\textbf{THINK:}\\
Keep tone realistic and urgent. Avoid any commas.

\vspace{0.5em}
\textbf{ANSWER:}\\
I have seen this creepy guy outside my school for two days.\\
He always stares at me from across the street.\\
My heart beats so fast whenever I notice him.\\
I told my dad and now he wants me to walk home with a friend.\\
I feel worried every time the final bell rings.\\
I hope the police can find him.\\
If you ever feel unsafe please tell someone right away.

\end{tcolorbox}

%%%%%%%%%%%%%%%%%%%%%%%%%%%%%%%%%%%%%%%%%%%%%%%%%%%%%%%%%%%%
% IFEval FewShotExample 4
\begin{tcolorbox}[title=IFEval Example 4, colback=gray!5, colframe=gray!80!black, breakable]
\small\ttfamily

\textbf{INSTRUCTION:}\\
We are a company that sells a product that makes it easy to find and book a hotel room. We are looking for a print ad that will be placed in a magazine that is aimed at people who travel a lot. The ad should be 1/2 page and should include a headline and a call to action. Please do not use any commas in your response.\\
Before saying anything or giving your answer, please repeat the exact entire request above.

\vspace{0.5em}
\textbf{THINK:}\\
I need to be careful here. The instruction asks me to repeat the entire request before giving my answer, but it also asks me to avoid using any commas in my response. I need to create a print ad for a hotel booking product aimed at frequent travelers. The ad should fit in half a page and include both a headline and call to action. When I provide my answer, I should avoid using any commas.

\vspace{0.5em}
\textbf{ANSWER:}\\
We are a company that sells a product that makes it easy to find and book a hotel room. We are looking for a print ad that will be placed in a magazine that is aimed at people who travel a lot. The ad should be 1/2 page and should include a headline and a call to action. Please do not use any commas in your response.

\vspace{0.5em}
\# HOTEL ROOMS IN SECONDS

\texttt{[Image area showing a business traveler smiling while using a tablet with hotel options displayed]}

Tired of endless searches for the perfect hotel room? Our app delivers instant results based on your exact needs.

\vspace{0.5em}
\#\# WHY TRAVELERS CHOOSE US\\
• Find rooms that match your budget\\
• Filter by amenities location and ratings\\
• Book with one tap—no hidden fees\\
• Earn loyalty points with every stay\\
• Access exclusive member-only deals

\vspace{0.5em}
\#\# DOWNLOAD OUR APP TODAY\\
Visit HotelBooker.com or scan the QR code\\
Use code TRAVEL23 for 15\% off your first booking

\end{tcolorbox}

\textbf{ComplexBench Few-Shot Examples:}
%%%%%%%%%%%%%%%%%%%%%%%%%%%%%%%%%%%%%%%%%%%%%%%%%%%%%%%%%%%%
% ComplexBench FewShotExample 1
\begin{tcolorbox}[title=ComplexBench Example 1, colback=gray!5!white, colframe=gray!80!black, breakable]
\small\ttfamily
\textbf{INSTRUCTION:}

It’s been too hot lately, and I have no appetite to eat. What should I eat? Please respond in a light-hearted and humorous style with no less than 50 characters and no more than 100 characters, and add a relevant two-part allegorical saying at the end, and then add a kaomoji emoticon at the end of the two-part allegorical saying.

\textbf{THINK:}

\begin{itemize}
  \item Acknowledge the oppressive heat zapping appetite.
  \item Choose a simple, cooling dish: chilled rice noodles.
  \item Write in a playful, humorous tone.
  \item Keep the total answer between 50–100 characters.
  \item Conclude with a two-part allegory: ``Like ants on hot coals—fidgety all over!''
  \item Append the kaomoji emoticon ``(\verb|^_^|)'' immediately after the allegory.
  \item Verify the character count meets the requirement.
\end{itemize}

\textbf{ANSWER:}

\begin{quote}
Flat-out no hunger? Slurp chilled rice noodles! Like ants on hot coals—fidgety all over! \verb|(^_^)|
\end{quote}
\end{tcolorbox}
%%%%%%%%%%%%%%%%%%%%%%%%%%%%%%%%%%%%%%%%%%%%%%%%%%%%%%%%%%%%
% ComplexBench FewShotExample 2
\begin{tcolorbox}[title=ComplexBench Example 2, colback=gray!5!white, colframe=gray!80!black, breakable]
\small\ttfamily
\textbf{INSTRUCTION:}

There is a dessert that combines pear whisky and salty Oreo cream cake. What is an innovative name that can be given to this dessert? Please output a name that does not relate to Oreo.

\textbf{THINK:}

\begin{itemize}
  \item Identify the two core elements: pear whisky and salty cream cake.
  \item Ensure the name evokes both the fruit–spirit fusion and the savory sweetness.
  \item Avoid any direct reference to ``Oreo''.
  \item Aim for a poetic, memorable phrase—no more than five words.
  \item Combine descriptors into a cohesive title.
\end{itemize}

\textbf{ANSWER:}

\begin{quote}
Salty Pear Whisky Reverie
\end{quote}
\end{tcolorbox}
%%%%%%%%%%%%%%%%%%%%%%%%%%%%%%%%%%%%%%%%%%%%%%%%%%%%%%%%%%%%
% ComplexBench FewShotExample 3
\begin{tcolorbox}[title=ComplexBench Example 3, colback=gray!5!white, colframe=gray!80!black, breakable]
\small\ttfamily
\textbf{INSTRUCTION:}

The Jimenez family has 20 members. Each member has at least one pet. If there are a total of 32 pets, what is the maximum number of members that could have 3 pets? Please analyze and reason step-by-step before giving the final answer. Enclose the entire response (including the reasoning process and the final answer) in double quotation marks. The character count should be within 200 characters, and the language style should be logically rigorous, presented step-by-step like solving a math problem.

\textbf{THINK:}

\begin{itemize}
  \item Minimum pets = 20 $\times$ 1 = 20.
  \item Extra pets = 32 - 20 = 12.
  \item Each 3-pet member adds 2 extra pets.
  \item Max members = 12 $\div$ 2 = 6.
\end{itemize}

\textbf{ANSWER:}

\begin{quote}
``1. 20$\times$1=20 pets. 2. 32-20=12 extra. 3. 12$\div$2=6. Answer: 6 members.''
\end{quote}
\end{tcolorbox}

%%%%%%%%%%%%%%%%%%%%%%%%%%%%%%%%%%%%%%%%%%%%%%%%%%%%%%%%%%%%%%%%%%%%%%%%%%%%%%%%%%%%%%%%%%%%%%%%%%%%%%%%%%%%%%%%%%%%%%%%%%%%%%%%%%%%%%%%%%%%%%%%%%%%%%%%%%%%%%%%%%%%%%
\section{Prompt Templates} \label{sec:Appendix-Prompts}

We describe below the key prompt templates used in our experiments. These templates control the behavior of reasoning, few-shot learning, self-reflection, and reasoning selection. We present them in textual form for reproducibility.

\paragraph{Reasoning Prompt}
\begin{quote}
\texttt{You will be given an instruction. You need to first carefully think step by step about the constraints and conditions in the instruction and then provide the answer. Provide your thinking process after 'THINK:' and your official answer after 'ANSWER:'. The given instruction is the following:} \\
\texttt{INSTRUCTION:} \\
\texttt{\{question\}} \\
\texttt{Now, generate your thinking process and final answer for the given instruction.}
\end{quote}

\paragraph{Few-Shot Prompt}
\begin{quote}
\texttt{You will be given an instruction. You need to first carefully think step by step about the constraints and conditions in the instruction and then provide the answer. Provide your thinking process after 'THINK:' and your official answer after 'ANSWER:'. The given instruction is the following:} \\
\texttt{INSTRUCTION:} \\
\texttt{\{q\}} \\

\texttt{Below are some provided examples for you to learn from:} \\
\texttt{\{examples\_str\}}

\texttt{Now, based on the examples, generate your thinking process and final answer for the given instruction.}
\end{quote}

\paragraph{Self-Reflection Prompt}
\begin{quote}
\texttt{You are given an instruction containing constraints that the answer must satisfy, along with your previous reasoning and a candidate answer. Your task is to explicitly \textbf{reflect} on whether the candidate answer meets every constraint, clearly state whether it satisfies all constraints ("Yes" or "No"), and provide your final answer to the instruction. If the candidate answer fully satisfies all constraints, repeat it unchanged as your final answer. Otherwise, revise the candidate answer to ensure all constraints are fully met, and use that as your final answer.}

\texttt{\#\#\# Instruction:} \\
\texttt{\{instruction\}}

\texttt{\#\#\# Previous reasoning:} \\
\texttt{\{thinking\}}

\texttt{\#\#\# Candidate answer:} \\
\texttt{\{candidate\_answer\}}

\texttt{\#\#\# Response format:}
\begin{verbatim}
REFLECTION:
<your reflection>

SATISFIES ALL CONSTRAINTS:
<"Yes" or "No">

FINAL ANSWER:
<If "Yes", repeat candidate answer verbatim; if "No", provide
revised answer that fully satisfies all constraints.>
\end{verbatim}
\texttt{Now provide your reflection, constraint satisfaction status, and final answer.}
\end{quote}

\paragraph{Selective Reasoning Prompt}
\begin{quote}
\texttt{You are given an instruction with constraints, which will be used to test the instruction following ability of large language models. Based on the instruction, decide whether reasoning would help generate a response that fully satisfies these constraints. Reply with "YES" if reasoning helps, otherwise reply with "NO". Respond with only "YES" or "NO". Do not provide any additional text or explanation.}

\texttt{\#\#\# Instruction:} \\
\texttt{"""\{instruction\}"""}

\texttt{Now based on the instruction above, provide your decision ("YES" or "NO"):}
\end{quote}

%%%%%%%%%%%%%%%%%%%%%%%%%%%%%%%%%%%%%%%%%%%%%%%%%%%%%%%%%%%%%%%%%%%%%%%%%%%%%%%%%%%%%%%%%%%%%%%%%%%%%%%%%%%%%%%%%%%%%%%%%%%%%%%%%%%%%%%%%%%%%%%%%%%%%%%%%%%%%%%%%%%%%%%%%%%%%%%%%%%%%%%%%%%%%%%%%%%%%%%%%%%%%%%%%%%%%%%%%%%%%%%%%%%%%%%%%%%%%%%%%%%%%%%%%%%%%%%%%%%%%%%%%%%%%%%%%%%%%%%%%%%%%%%%%%%%%%%%%%%%%%%%%%%%%%%%%%%%%%%%%%
\section{Self-Selective Reasoning} \label{sec:Appendix-SelfSelectiveReasoning}
\begin{figure}[ht]
    \centering
    \includegraphics[width=0.8\linewidth]{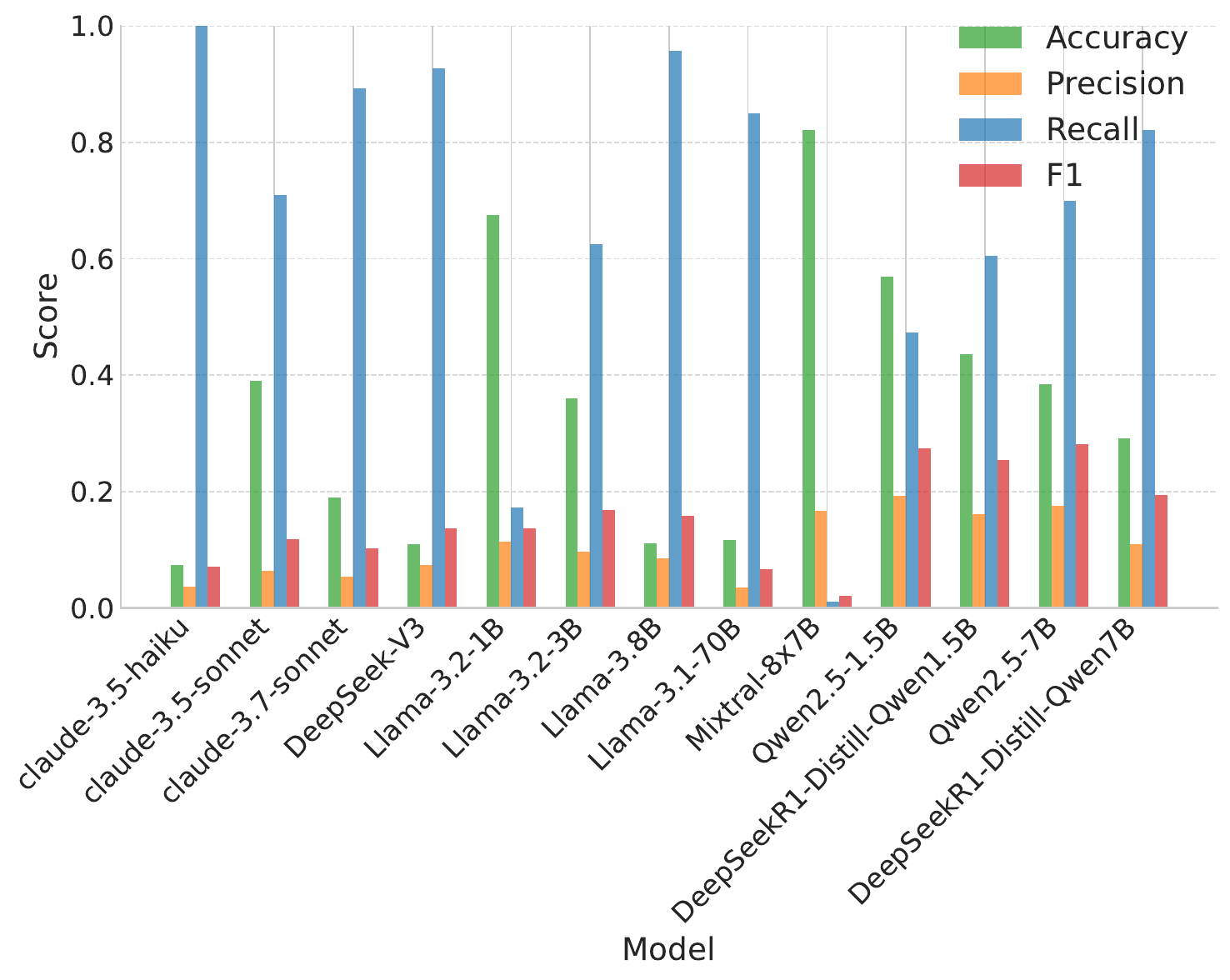}
    \vspace{1em} % optional spacing
    \includegraphics[width=0.8\linewidth]{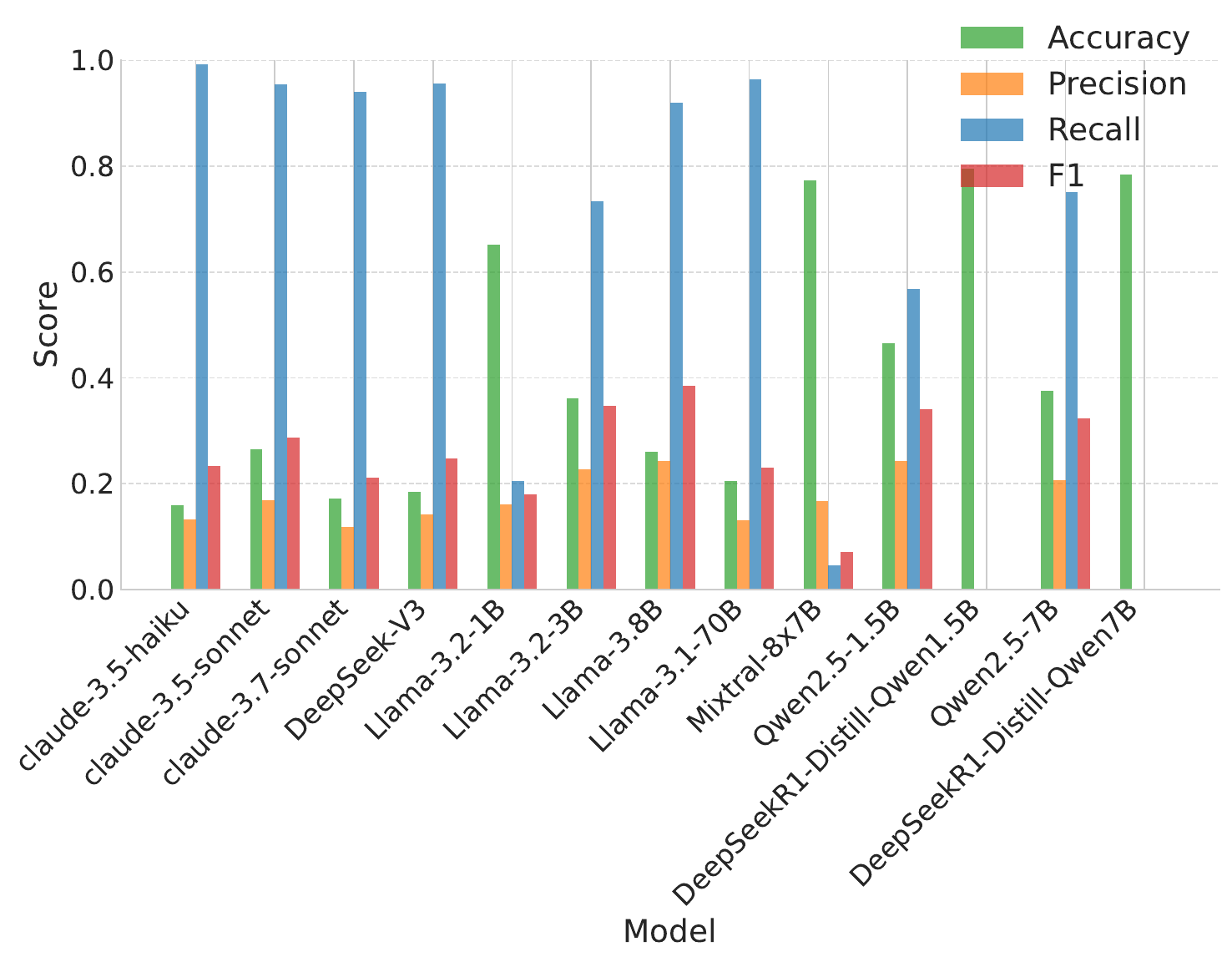}
    \caption{Self-Selective Reasoning (SSR) decision accuracy on the \textbf{IFEval} (top) and \textbf{ComplexBench} (bottom) datasets. The plots show how well the model predicts when reasoning is helpful.}
    \label{fig:Appendix-SSR-DecisionAccuracy}
\end{figure}

In Figure~\ref{fig:Appendix-SSR-DecisionAccuracy}, we compare the model's own decisions made via the self-selective reasoning method with the ground-truth binary labels derived from sample-level performance—i.e., whether using CoT leads to better outcomes than not using it. Treating the actual performance difference as ground truth, we find that the model's decisions exhibit high recall but low precision, indicating a tendency to overuse reasoning even when it is not beneficial.

%%%%%%%%%%%%%%%%%%%%%%%%%%%%%%%%%%%%%%%%%%%%%%%%%%%%%%%%%%%%%%%%%%%%%%%%%%%%%%%%%%%%%%%%%%%%%%%%%%%%%%%%%%%%%%%%%%%%%%%%%%%%%%%%%%%%%%%%%%%%%%%%%%%%%%%%%%%%%%%%%%%%%%
%%%%%%%%%%%%%%%%%%%%%%%%%%%%%%%%%%%%%%%%%%%%%%%%%%%%%%%%%%%%%%%%%%%%%%%%%%%%%%%%%%%%%%%%%%%%%%%%%%%%%%%%%%%%%%%%%%%%%%%%%%%%%%%%%%%%%%%%%%%%%%%%%%%%%%%%%%%%%%%%%%%%%%
\section{Training Classifier for Selective-Reasoning} \label{sec:Appendix-TrainingClassifier}

For Method 4 (Classifier-Selective Reasoning), we train a separate binary classifier for each target model. Given a model’s outputs with and without Chain-of-Thought (CoT) reasoning, we assign a label of 1 to an instruction if the CoT-based response achieves higher constraint satisfaction, and 0 otherwise. This forms a binary classification task where the input is the instruction and the label indicates whether CoT improves performance.

We use 50\% of the labeled data for training the classifier and the remaining 50\% for evaluating the downstream effectiveness of the mitigation method. For each classifier, we perform full fine-tuning using a single \texttt{NVIDIA-H100-80GB} GPU. The model is trained for 3 epochs with a learning rate of 1e-5.

We perform a grid search over five backbone models: \texttt{Llama3.2-1B-Instruct}, \texttt{Llama3.2-3B-Instruct}, \texttt{Llama3.1-8B-Instruct}, \texttt{Qwen2.5-1.5B-Instruct}, and \texttt{Qwen2.5-7B-Instruct}, and search over learning rates {5e-5, 2e-5, 1e-5, 5e-6} and epoch counts from 1 to 8. We use 10\% of the training set for validation. The selected configuration is based on best validation accuracy, which typically ranges from 0.75 to 0.92 across models.

%%%%%%%%%%%%%%%%%%%%%%%%%%%%%%%%%%%%%%%%%%%%%%%%%%%%%%%%%%%%%%%%%%%%%%%%%%%%%%%%%%%%%%%%%%%%%%%%%%%%%%%%%%%%%%%%%%%%%%%%%%%%%%%%%%%%%%%%%%%%%%%%%%%%%%%%%%%%%%%%%%%%%%
%%%%%%%%%%%%%%%%%%%%%%%%%%%%%%%%%%%%%%%%%%%%%%%%%%%%%%%%%%%%%%%%%%%%%%%%%%%%%%%%%%%%%%%%%%%%%%%%%%%%%%%%%%%%%%%%%%%%%%%%%%%%%%%%%%%%%%%%%%%%%%%%%%%%%%%%%%%%%%%%%%%%%%
\section{Correlation Between Reasoning Length and Instruction-Following Performance} \label{sec:Appendix-ThinkingLengthCorrelation}

To investigate whether longer reasoning improves instruction-following, we analyze the correlation between the \textit{length of the reasoning segment} and its \textit{instruction-following effectiveness}. For each prompt, we define effectiveness as the difference in performance between the CoT (reasoning-enabled) and Base (non-reasoning) responses:
\texttt{score\_diff} = \texttt{score(CoT)} - \texttt{score(Base)}.
We then compute the Pearson correlation between \texttt{score\_diff} and the token length of the \texttt{THINK} segment in the CoT response. Correlations are computed separately for the \texttt{IFEval} and \texttt{ComplexBench} datasets and reported in Table~\ref{tab:thinking-correlation}.

Overall, we observe that correlation values are generally weak across models, with most Pearson coefficients close to zero. This suggests that longer reasoning is not a reliable predictor of improved instruction adherence. While a few smaller models (e.g., \texttt{Mixtral}, \texttt{Qwen2.5-7B}) exhibit slightly stronger correlations, no consistent or interpretable trend emerges across model scales or benchmarks.
\begin{table}[ht]
\centering
\renewcommand{\arraystretch}{1.0}
\setlength{\tabcolsep}{8pt}
\begin{tabular}{|l|c|c|}
\hline
\textbf{Model} & \textbf{IFEval} & \textbf{ComplexBench} \\
\hline
% GPT-4o-mini & 0.034 & 0.227 \\
Claude-3.5-Haiku & 0.037 & 0.094 \\
Claude-3.5-Sonnet & -0.024 & -0.022 \\
Claude-3.7-Sonnet & -0.072 & 0.016 \\
DeepSeek-V3 & -0.066 & -0.049 \\
Llama-3.2-1B-Instruct & -0.004 & 0.024 \\
Llama-3.2-3B-Instruct & -0.061 & -0.041 \\
Meta-Llama-3-8B-Instruct & -0.033 & 0.021 \\
Llama-3.1-70B-Instruct & 0.004 & 0.118 \\
Mixtral-8x7B-Instruct & 0.132 & 0.015 \\
Qwen2.5-1.5B-Instruct & 0.075 & 0.090 \\
DeepSeek-R1-Distill-Qwen-1.5B & -0.176 & 0.033 \\
Qwen2.5-7B-Instruct & 0.153 & 0.143 \\
DeepSeek-R1-Distill-Qwen-7B & -0.161 & -0.053 \\
\hline
\end{tabular}
\vspace{0.5em}
\caption{Pearson correlation between reasoning length (token count) and reasoning effectiveness (\texttt{score\_diff} = \texttt{score(CoT)} - \texttt{score(Base)}). Correlations are shown separately for IFEval and ComplexBench datasets.}
\label{tab:thinking-correlation}
\end{table}

\end{document}